\documentclass[manuscript,screen]{acmart}

\renewcommand\footnotetextcopyrightpermission[1]{} 
\AtBeginDocument{%
	\providecommand\BibTeX{{%
			\normalfont B\kern-0.5em{\scshape i\kern-0.25em b}\kern-0.8em\TeX}}}

\setcopyright{acmcopyright}
\copyrightyear{2024}
\acmYear{2024}

\newcommand{\paratitle}[1]{\vspace{1.5ex}\noindent\textbf{#1}}

\newcommand{\ignore}[1]{}

\usepackage{hyperref}
\usepackage[figuresright]{rotating}
\usepackage{array}
\newcolumntype{H}{>{\setbox0=\hbox\bgroup}c<{\egroup}@{}}
\usepackage{epigraph}
\usepackage{booktabs}
\usepackage{forest}
\usepackage{xcolor}
\usepackage{mdframed}
\usepackage{tcolorbox}
\usepackage{multirow} 
\usepackage{makecell}
\begin{document}
	
	\title{Scientific Large Language Models: A Survey on Biological \& Chemical Domains}
	
	\author{Qiang Zhang}
	\authornote{Both authors contributed equally to this research. Email to: \href{qiang.zhang.cs@zju.edu.cn}{qiang.zhang.cs@zju.edu.cn}, \href{dingkeyan@zju.edu.cn}{dingkeyan@zju.edu.cn}}
	\email{qiang.zhang.cs@zju.edu.cn}
	\orcid{0000-0003-1636-5269}
	
	\author{Keyan Ding}
	\authornotemark[1]
	\email{dingkeyan@zju.edu.cn}
	\orcid{0000-0003-2900-7313}
	\affiliation{%
		\institution{College of Computer Science and Technology, Zhejiang University}
		\country{China}
	}
	\affiliation{%
		\institution{Hangzhou Innovation Center, Zhejiang University}
		\country{China}
	}
	
	\author{Tianwen Lyu}
	\orcid{0009-0009-2691-6467}
	
	\author{Xinda Wang}
	\orcid{0009-0004-5559-1714}
	
	\author{Qingyu Yin}
	\orcid{0009-0000-5612-5521}
	
	\author{Yiwen Zhang}
	\orcid{0009-0001-4551-6938}
	
	\author{Jing Yu}
	\orcid{0009-0002-1156-9326}
	
	\author{Yuhao Wang}
	\orcid{0009-0000-0013-7259}
	
	\author{Xiaotong Li}
	\orcid{0009-0003-3930-4474}
	
	\author{Zhuoyi Xiang}
	\orcid{0009-0003-4230-3026}
	
	\author{Kehua Feng}
	\orcid{0009-0006-7923-5604}
	
	\author{Xiang Zhuang}
	\orcid{0000-0002-0253-1476}
	
	\author{Zeyuan Wang}
	\orcid{0000-0002-5036-9602}
	
	\author{Ming Qin}
	\orcid{0000-0001-8607-8965}
	
	\author{Mengyao Zhang}
	\orcid{0009-0009-6545-8171}
	\affiliation{%
		\institution{College of Computer Science and Technology, Zhejiang University}
		\country{China}
	}
	
	\author{Jinlu Zhang}
	\orcid{0009-0002-3336-8074}
	\affiliation{
		\institution{College of Pharmaceutical Sciences, Zhejiang University}
		\country{China}
	}
	
	\author{Jiyu Cui}
	\orcid{0009-0005-1409-7338}
	\affiliation{
		\institution{College of Chemical and Biological Engineering, Zhejiang University}
		\country{China}
	}
	\author{Tao Huang}
	\orcid{0000-0002-7645-4290}
	\email{huangtao@him.cas.cn}
	\author{Pengju Yan}
	\orcid{0009-0007-0748-5919}
	\email{yanpengju@gmail.com}
	\affiliation{
		\institution{Hangzhou Institute of Medicine, Chinese Academy of Sciences}
		\country{China}
	}
	\author{Renjun Xu}
	\orcid{0000-0002-7566-7948}
	\affiliation{%
		\institution{Center for Data Science, Zhejiang University}
		\country{China}
	}
	
	\author{Hongyang Chen}
	\orcid{0000-0002-7626-0162}
	\affiliation{
		\institution{Zhejiang Lab}
		\country{China}
	}

	\author{Xiaolin Li}
	\orcid{0000-0002-3368-159X}
	\email{xiaolinli@ieee.org}
	\affiliation{
		\institution{Hangzhou Institute of Medicine, Chinese Academy of Sciences}
		\country{China}
	}
	
	\author{Xiaohui Fan}
	\orcid{0000-0002-6336-3007}
	\affiliation{
		\institution{College of Pharmaceutical Sciences, Zhejiang University}
		\country{China}
	}
	
	\author{Huabin Xing}
	\orcid{0000-0002-7418-0046}
	\affiliation{
		\institution{College of Chemical and Biological Engineering, Zhejiang University}
		\country{China}
	}
	\affiliation{%
		\institution{Hangzhou Innovation Center, Zhejiang University}
		\country{China}
	}
	
	\author{Huajun Chen}
	\orcid{0000-0001-5496-7442}
	\email{huajunsir@zju.edu.cn}
	\authornote{Corresponding author. Email to: \href{huajunsir@zju.edu.cn}{huajunsir@zju.edu.cn}} 
	\affiliation{%
		\institution{College of Computer Science and Technology, Zhejiang University}
		\country{China}
	}
	\affiliation{%
		\institution{Hangzhou Innovation Center, Zhejiang University}
		\country{China}
	}

	\makeatletter
	\let\@authorsaddresses\@empty
	\makeatother
	
	\renewcommand{\shortauthors}{Zhang and Ding, et al.}
	
	\begin{abstract}
		Large Language Models (LLMs) have emerged as a transformative power in enhancing natural language comprehension, representing a significant stride toward artificial general intelligence. The application of LLMs extends beyond conventional linguistic boundaries, encompassing specialized linguistic systems developed within various scientific disciplines. This growing interest has led to the advent of scientific LLMs, a novel subclass specifically engineered for facilitating scientific discovery. As a burgeoning area in the community of AI for Science, scientific LLMs warrant comprehensive exploration. However, a systematic and up-to-date survey introducing them is currently lacking. In this paper, we endeavor to methodically delineate the concept of ``scientific language'', whilst providing a thorough review of the latest advancements in scientific LLMs. Given the expansive realm of scientific disciplines, our analysis adopts a focused lens, concentrating on the biological and chemical domains. This includes an in-depth examination of LLMs for textual knowledge, small molecules, macromolecular proteins, genomic sequences, and their combinations, analyzing them in terms of model architectures, capabilities, datasets, and evaluation. Finally, we critically examine the prevailing challenges and point out promising research directions along with the advances of LLMs. By offering a comprehensive overview of technical developments in this field, this survey aspires to be an invaluable resource for researchers navigating the intricate landscape of scientific LLMs.
	\end{abstract}

	

	
	\maketitle
	\thispagestyle{empty}
	
	\newpage
	\setcounter{tocdepth}{2}
	
	\tableofcontents
	
	\newpage
	
	\section{Introduction}\label{sec:introduction}
\epigraph{\textit{``The limits of my language mean the limits of my world.''}}{--- \textit{Ludwig Wittgenstein}}

Humanity acquires knowledge of the world through perception and cognition, where \emph{natural languages} (i.e., human languages) stand as the quintessential medium for articulating this \emph{world knowledge}. Historically, this plethora of world knowledge has been expressed, chronicled, and disseminated through natural languages. Currently, \emph{Large Language Models} (LLMs) stand as cutting-edge tools in processing natural language and gathering world knowledge. Typically, LLMs refer to those based on Transformer-based architectures with hundreds of millions (or even billions) of trainable parameters, trained on extensive textual corpus~\cite{Shanahan-arxiv-2022-Talking}. Typical examples include GPT-3~\cite{Brown-NeurIPS-2020-Language}, PaLM~\cite{Chowdhery-arxiv-2022-PaLM}, Galactica~\cite{Taylor-arxiv-2022-Galactica}, LLaMA~\cite{Touvron-arxiv-2023-LLaMA}, ChatGLM~\cite{zeng2023glm-130b} and Baichuan 2~\cite{baichuan2023baichuan2}. They have exhibited strong capacities to understand natural language and address complex tasks (in the manner of text generation), and have incited substantial interest in both academic and industrial domains. The exceptional performance of LLMs sparks hope that they may evolve into Artificial General Intelligence (AGI) in our current era.

\begin{figure}[t]
    \centering
    \includegraphics[width=0.8\linewidth]{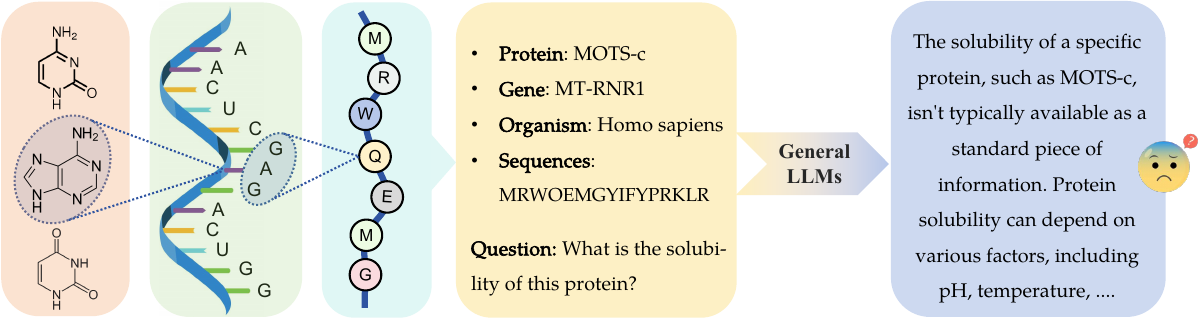}
    \caption{Illustrations that general LLMs struggle to effectively handle scientific languages, such as molecules, RNA and amino acid sequences in this example.}
    \centering
    \vspace{-0.5em}
    \label{fig:examples-sci}
\end{figure}

Besides natural languages, to encapsulate more specialized \emph{science knowledge}, an assortment of \emph{scientific languages} has been developed, as illustrated in Fig.~\ref{fig:examples-sci}. This encompasses textual expressions in the scientific research domains, mathematical languages to define mathematical formulas, chemical languages such as SMILES that represent molecular structures, and biological languages that describe proteins or genomes, and detail the complex constitution of living organisms. 
These scientific languages come with their distinct vocabularies, where each term holds a specific meaning that can be completely different from natural languages. For example, the character ``C'' in English represents the amino acid Cysteine in protein languages~\cite{gdr1984nomenclature}, while in the SMILES language system, it denotes a Carbon atom~\cite{weininger1988smiles}. Furthermore, experts in specific domains establish grammatical rules to organize these terms, enabling the construction of sentences with precise semantic functions. For instance, computational chemists create grammatical rules ensuring the accuracy of machine-generated molecules in SELFIES format~\cite{krenn2020self}.
After decades of evolution, scientific languages have become invaluable tools, significantly accelerating scientific discoveries. Due to the potential semantic and grammatical differences between scientific and natural languages, existing general LLMs (such as ChatGPT~\footnote{\url{https://chat.openai.com}} or GPT-4~\cite{OpenAI-OpenAI-2023-GPT-4}) often fail to properly deal with scientific data like molecules and proteins~\cite{ai4science2023impact}. As the well-known Austrian philosopher Ludwig Wittgenstein indicates, ``The limits of my language mean the limits of my world.''~\cite{ramsey1923tractatus} The world of general LLMs can be limited to natural languages.

 To facilitate the understanding of scientific languages, researchers have devised \emph{Scientific Large Language Models} (Sci-LLMs) customized for various scientific domains and disciplines.
 For instance, molecular language models have been developed to represent molecule structures as a string of atoms and chemical bonds~\cite{li2021mol}. These models aid in predicting molecular properties~\cite{wang2019smiles}, designing new drugs~\cite{zhang2022pushing}, and proposing retrosynthesis routes~\cite{schwaller2021mapping}. Similarly, protein language models operate based on sequences of amino acids~\cite{rives2019biological,brandes2022proteinbert}. They are used to forecast 3D protein structures and functions~\cite{lin2022language}, enhance existing proteins for improved fitness~\cite{Pascal2023ProteinNPT}, and create new proteins with specific functionalities~\cite{nijkamp2023progen2}. As a burgeoning area within the AI-for-Science research, many Sci-LLMs have been proposed with modified architectures, learning methods, training corpus, and evaluation benchmarks and criteria. Despite their notable achievements, these models have mostly been explored within their respective research domains. There is currently a lack of a comprehensive review that unifies these language modeling advancements.

In this survey, we set out to fill this gap by systematically reviewing the technical advancement of Sci-LLMs with a close reference to general LLMs. Considering the expansive scope of scientific languages, we focus our investigation on biological and chemical languages. Specifically, our examination encompassed molecular languages, protein languages, and genomic languages. In addition to these specialized scientific languages, we recognize the immense scientific knowledge embedded in textbooks, patents, and research papers, composed in natural languages. Consequently, we explore the textual LLMs with an emphasis on scientific knowledge, and more importantly, the multimodal LLMs that encompass 
various types of scientific languages. 

When delving into each language system, we first review the LLM architectures and categorize them into three classes: encoder-only, decoder-only and encoder-decoder. Then we report the model capabilities and summarize the typical downstream tasks that Sci-LLMs can conduct. In terms of model training and evaluation, we collect a bunch of commonly used training corpus and evaluation benchmarks. Finally, we present the proper criteria for discriminative and generative tasks of scientific language modeling.

\begin{figure}[t]
    \centering
    \includegraphics[width=0.7\linewidth]{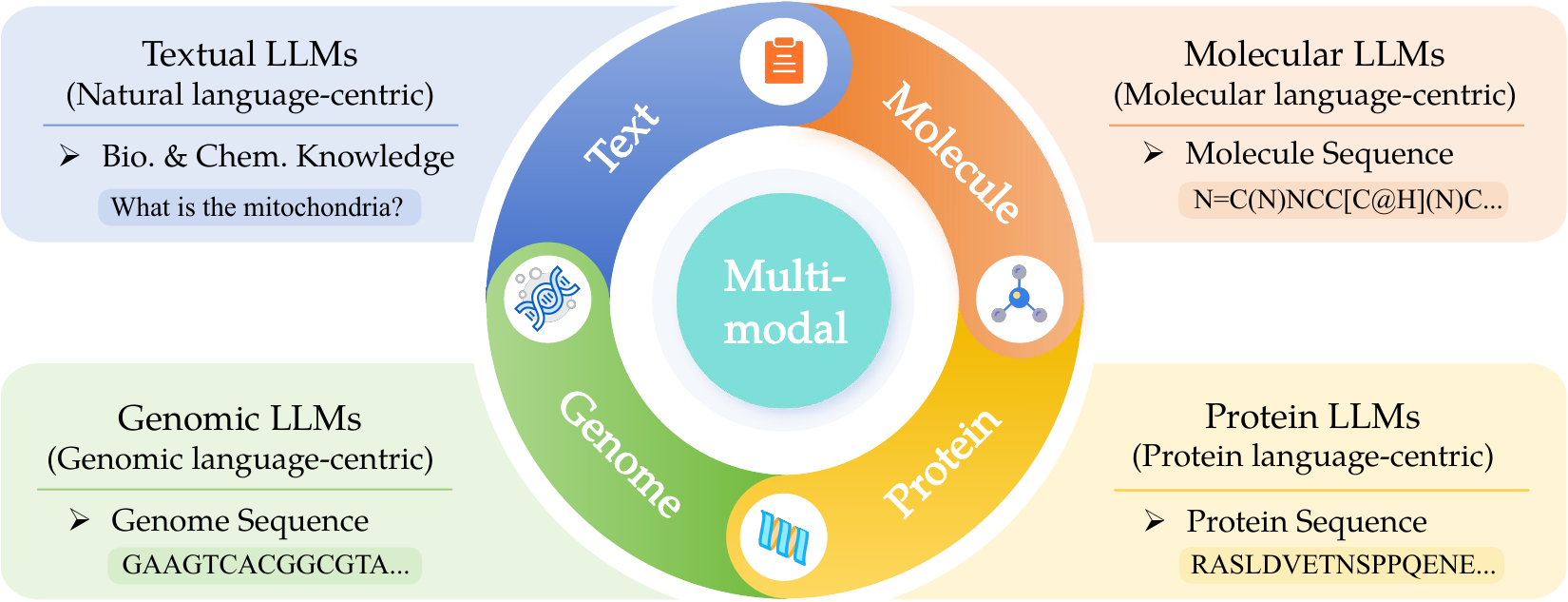}
    \caption{Research scopes of Scientific Large Language Models (Sci-LLMs) in this survey. We focus on scientific languages (i.e., textual, molecular, protein and genomic languages), as well as their combination (i.e., multimodal language),  within the realm of biochemical science.}
    \centering
    \vspace{-0.5em}
    \label{fig:scopes}
\end{figure}

This survey is confined within specific boundaries. Firstly, we focus on scientific languages, specifically chemical and biological languages. We have excluded languages lacking both universally defined vocabularies and grammatical structures, such as mathematical languages. Secondly, when discussing textual LLMs, our emphasis remains on chemical and biological domain knowledge expressed in natural languages. This choice ensures a consistent and coherent interaction with languages specific to chemistry and biology, such as molecular and protein languages. Thirdly, our technical exploration is primarily confined to Transformer-based language models. We have not included alternative neural architectures like graph neural networks and diffusion models, despite their wide applications in molecule and protein modeling. 
Fig.~\ref{fig:scopes} describes the research scopes of Sci-LLMs in this survey.

\begin{sidewaysfigure}[h]
    \vspace{50em}
    \includegraphics[width=1\linewidth]{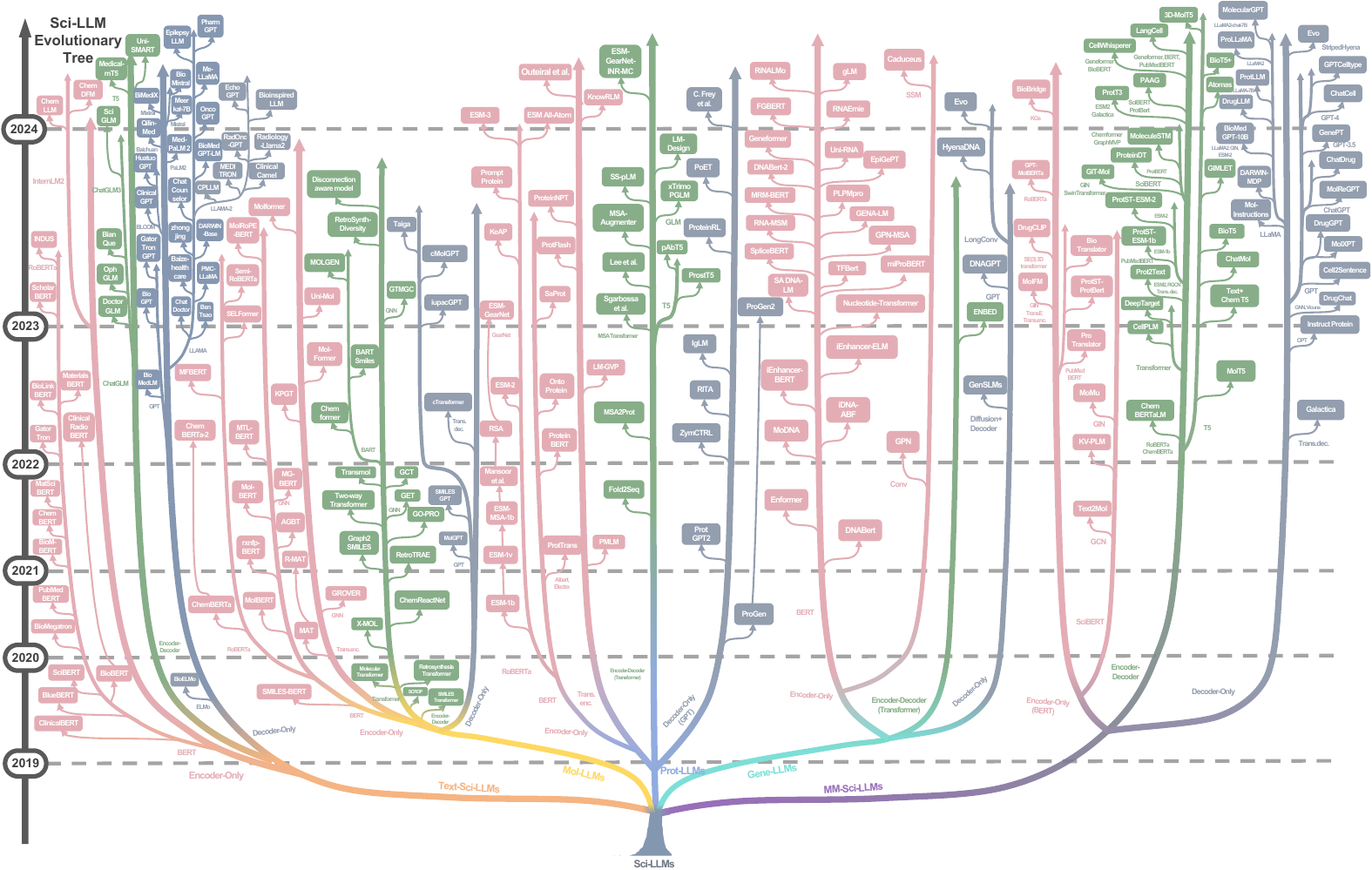} 
    \caption{ An evolutionary tree of Sci-LLMs, which consists of five main branches corresponding to the research scopes in this survey. Due to the extensive number of Sci-LLMs, it is not feasible to include all of them in this figure, despite their exceptional quality. For detailed information on the featured models, please refer to Table \ref{tab:Text-Sci-LLMs}, \ref{tab:Mol_LLMs}, \ref{tab:Prot-LLMs}, \ref{tab:Gene-LLMs}, and \ref{tab:MM-Sci-LLM}. Considering the rapid development of Sci-LLMs, we will share the source file of this figure and encourage readers to make incremental updates at \url{https://kdocs.cn/l/cbRA94QwMhmn}.}
    \label{fig:evo-tree}
\end{sidewaysfigure}

This survey's distinct boundaries set it apart from other reviews of LLMs and computational modeling for molecules, proteins, and genomes. In contrast to those primarily centered on natural languages \cite{zhao2023survey,yang2023harnessing}, our emphasis leans more towards scientific languages. Unlike surveys concentrating solely on molecule~\cite{du2022molgensurvey,xia2023systematic}, protein~\cite{bepler2021learning,hu2022protein,tran2023survey,unsal2022learning}, or genome data~\cite{consens2023transformers}, we aim to provide a comprehensive view of language models for chemical and biological research. Furthermore, we delve into multimodal LLMs, exploring the interactions between texts and molecule/protein/genome languages. To the best of our knowledge, this nuanced exploration has not been covered in previous surveys. The contributions of the survey can be summarized as follows: 
\begin{itemize}
    \item  We offer a comprehensive review of language modeling within scientific domains, encompassing textual, molecular, protein, and genomic languages, emphasizing domain-specific knowledge. 
    \item We provide a detailed summary of existing Sci-LLMs, covering model architectures, capabilities, training data, evaluation benchmarks, and assessment criteria. We also show the evolutionary tree of Sci-LLMs in Fig. \ref{fig:evo-tree}.
    \item We enumerate available resources for Sci-LLMs, open-source and maintain the related materials at \url{https://github.com/HICAI-ZJU/Scientific-LLM-Survey}, thereby facilitating accessibility for newcomers to the field.
    \item To our best knowledge, this survey represents the first comprehensive overview of multimodal Sci-LLMs designed to explore the interaction between various scientific languages.
\end{itemize}

The remainder of this survey is organized as follows: 
Sec.\ref{sec-background} introduces the background of LLMs and formulates relevant concepts. Sec.\ref{sec-text}, Sec.\ref{sec-molecule}, Sec.\ref{sec-protein}, Sec.\ref{sec-genome}  and Sec.\ref{sec-multimodal} present Sci-LLMs, including textual, molecular, protein, genomic and multimodal LLMs, respectively. Finally, we analyze the limitations of existing models, pinpoint potential research directions and conclude this survey in Sec.\ref{sec-conclusion}.

	\clearpage
	
	\section{Background}
\label{sec-background}

\subsection{Formulation of Scientific Languages}\label{sec:formula}

Language is a sophisticated system of communication, typically with well-defined vocabulary and grammatical rules~\cite{evans2009myth}. It enables the transmission of intricate semantic information, serving as a remarkable feature that distinguishes humans from animals~\cite{le2004wave}. One typical language system is the human language, or called natural language, such as English, Chinese and Spanish. In addition to these, there exist non-natural language systems, such as programming languages like Java and Python, and chemical molecule languages like SMILES. Each language system possesses its unique set of vocabulary and grammar, often entirely disparate from others. Consequently, proficiency in one language system does not necessarily equate to proficiency in another~\cite{ai4science2023impact}. In this survey, we delve into the realm of scientific languages, concentrating not only on domain-specific languages like those of molecules, proteins, and genomes but also on the textual languages that are central to various scientific domains.

\paratitle{Molecular Language.} The chemical molecule language system encompasses specialized languages utilized to represent molecular structures and compositions in chemistry.  
A prominent example in chemical molecule language systems is SMILES (Simplified Molecular Input Line Entry System)~\cite{weininger1988smiles}. It describes molecular structures by encoding atoms, bonds, and connectivity patterns using ASCII strings. For instance, the SMILES notation for Ethanol is ``CCO''. Each letter represents an atom (e.g., `C' for carbon, `O' for oxygen), while special characters denote different bond types (e.g., `-', `=', `\#'). 
SELFIES (Simplified Molecular Input Line Entry System)~\cite{krenn2020self} is a recent molecular language that utilizes a syntax designed with specific rules to represent molecular structures. It employs another set of predefined symbols to encode molecular information. For instance, the SELFIES representation of ethanol is ``[C][C][O]''. The grammar dictates the rules and syntax for constructing valid representations of molecular structures, and governs how atoms, bonds, and other chemical entities are arranged and connected within the encoded strings. 
InChI (International Chemical Identifier)~\cite{heller2013inchi} is another identifier for chemical substances that provides a machine-readable description of the substance. InChI comprises layers of information that can represent molecular connectivity, hydrogen position, isotope composition, stereochemistry, and electronic charge. The InChI algorithm converts input structural information of molecules into a unique identifier through normalization to remove redundant information, canonicalization to generate a unique number label for each atom, and serialization to produce a string of characters.
The International Union of Pure and Applied Chemistry (IUPAC) defines a set of rules that converts chemical structures to natural languages~\cite{favre2013nomenclature}.

\paratitle{Protein Language.} The protein language system is a specialized framework used to represent the primary structures of proteins. It offers a method to encode amino acid sequences, which are the building blocks of proteins, into a structured format that can be interpreted and analyzed computationally.
The vocabulary in the protein language system consists of symbols representing the 20 standard amino acids. Each amino acid is denoted by a specific one-letter or three-letter code~\cite{gdr1984nomenclature}. For instance, `A' represents alanine, `R' represents arginine, and `L' represents leucine in the one-letter code, while `ALA' represents alanine, `ARG' represents arginine, and `LEU' represents leucine in the three-letter code. In terms of grammatical rules, the protein language system does not have a traditional grammar like natural languages. The protein language system enables scientists to describe and analyze protein structures and functions computationally. It allows for the systematic representation of amino acid sequences, aiding in the understanding of protein characteristics, interactions, and biological roles in various biological processes.

\paratitle{Genomic Language.} 
The genomic language system refers to the code used to represent genetic information in living organisms, encompassing both DNA (DeoxyriboNucleic Acid) and RNA (RiboNucleic Acid)~\cite{soukup2001nucleic}. These systems serve as the fundamental languages for encoding genetic instructions and hereditary traits. The vocabulary in the genomic language system includes nucleotide base pairs. In DNA, these are Adenine (A), Thymine (T), Cytosine (C), and Guanine (G). In RNA, thymine is replaced by Uracil (U). Combinations of these bases form the genetic code, representing sequences in genes. Similar to the protein language system, the genomic language does not have a clear set of grammatical rules to organize the tokens in the vocabulary. The genomic language system provides a standardized way to encode and interpret genetic information. It allows scientists to study DNA and RNA sequences, decode genetic instructions, and understand the processes of inheritance, gene expression, and molecular biology.

It is noteworthy that, in addition to the aforementioned sequence-based languages (one-dimensional), alternative forms can be employed to represent molecules, proteins and genomes, such as graphs (two-dimensional) and structures (three-dimensional).
However, considering our focus on language modeling with sequences (i.e., LLMs), we have intentionally excluded the models that are solely based on other modalities in this survey.

\begin{figure}[t]
    \centering
    \includegraphics[width=0.7\linewidth]{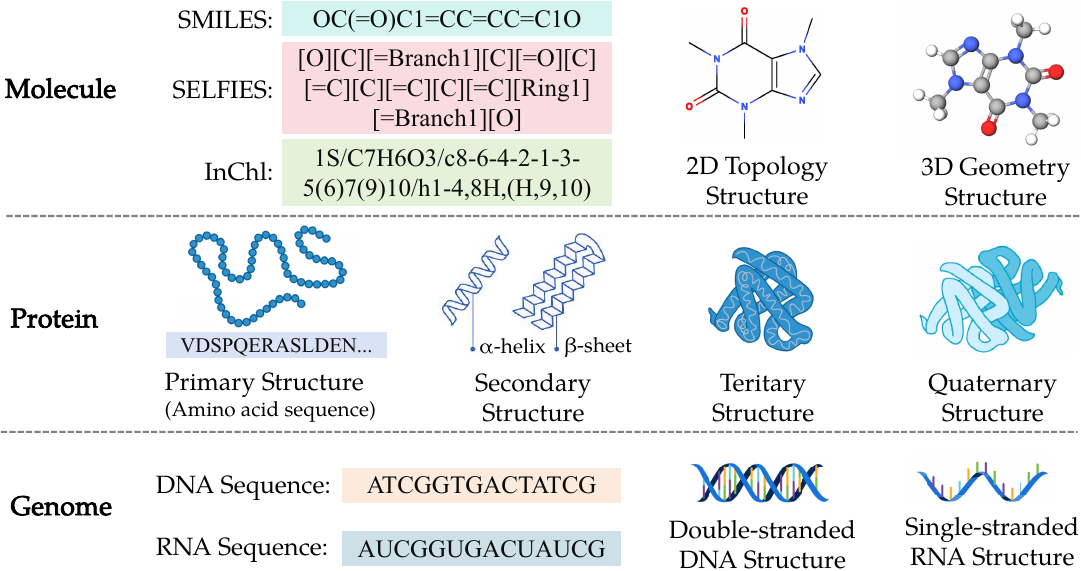}
    \caption{Illustration of molecular, protein and genomic languages. Molecular languages include SMILEs, SELFIES and InChl sequences, and 2D topology and 3D geometry structures. Protein languages consist of primary structure (i.e., amino acid sequence), secondary, tertiary, and quaternary structures (3D). Genomic languages include DNA and RNA sequences/structures. This survey focuses solely on sequence modeling of molecular, protein and genomic languages.
    }
    \label{fig:languages}
    \vspace{-1em}
\end{figure}

\subsection{Taxonomy of Model Architectures}\label{sec:arch}
The taxonomy of model architectures in LLMs can be broadly categorized into three types: encoder-only, decoder-only, and encoder-decoder. Each of these architectures serves different purposes and applies to different kinds of tasks.

\paratitle{Encoder-only.}  Encoder-only models are a type of neural network architecture that solely consists of an encoder part without a corresponding decoder. These models are designed to process input data and generate a fixed-size representation or encoding of that data. The encoder processes the input sequence and compresses the information into a latent space or encoding which can be used for various downstream tasks such as classification, clustering, or any other task requiring an understanding of the input data. A common example of an encoder-only model is the BERT~\cite{Devlin-NAACL-2019-BERT} (Bidirectional Encoder Representations from Transformers) model which is used in natural language processing tasks to generate meaningful representations of text data. In the context of protein, encoder-only models can be employed to learn representations of protein sequences which can then be used for various bioinformatics tasks.

\paratitle{Decoder-only.}
Decoder-only models are neural network architectures that consist solely of a decoder part without a corresponding encoder. These models are typically used for generating sequences or other forms of output data based on given inputs. Unlike encoder-only models, which aim to create a compact representation of the input data, decoder-only models focus on generating or expanding information, often based on a given seed input or condition. A well-known example of a decoder-only model is the Generative Pre-trained Transformer (GPT~\cite{OpenAI-blog-2022-ChatGPT}) series, which is designed to generate text sequences based on provided prompts or initial text fragments. In the field of genomics, decoder-only models could be used for tasks like sequence generation, prediction of missing or subsequent data based on given sequence fragments, or other generative tasks. 

\paratitle{Encoder-decoder.}
Encoder-decoder models are a class of neural network architectures that are particularly powerful in tasks involving sequence-to-sequence transformations. These models consist of two primary components: an encoder and a decoder. The encoder's function is to process the input data and compress it into a context-rich representation, often referred to as the context vector or encoded state. The decoder, on the other hand, takes the encoded representation and generates the output sequence.  
A distinguished example is the Text-to-Text Transfer Transformer (T5~\cite{raffel2020exploring}), which proposes a unified framework that converts all of natural language processing tasks into a sequence-to-sequence format. 
The separation of the encoding and decoding processes enables the model to specialize in understanding and generating sequences, making it ideal for various downstream tasks, such as molecule captioning, editing and summarization.

\begin{figure}[t]
    \centering
    \includegraphics[width=0.75\linewidth]{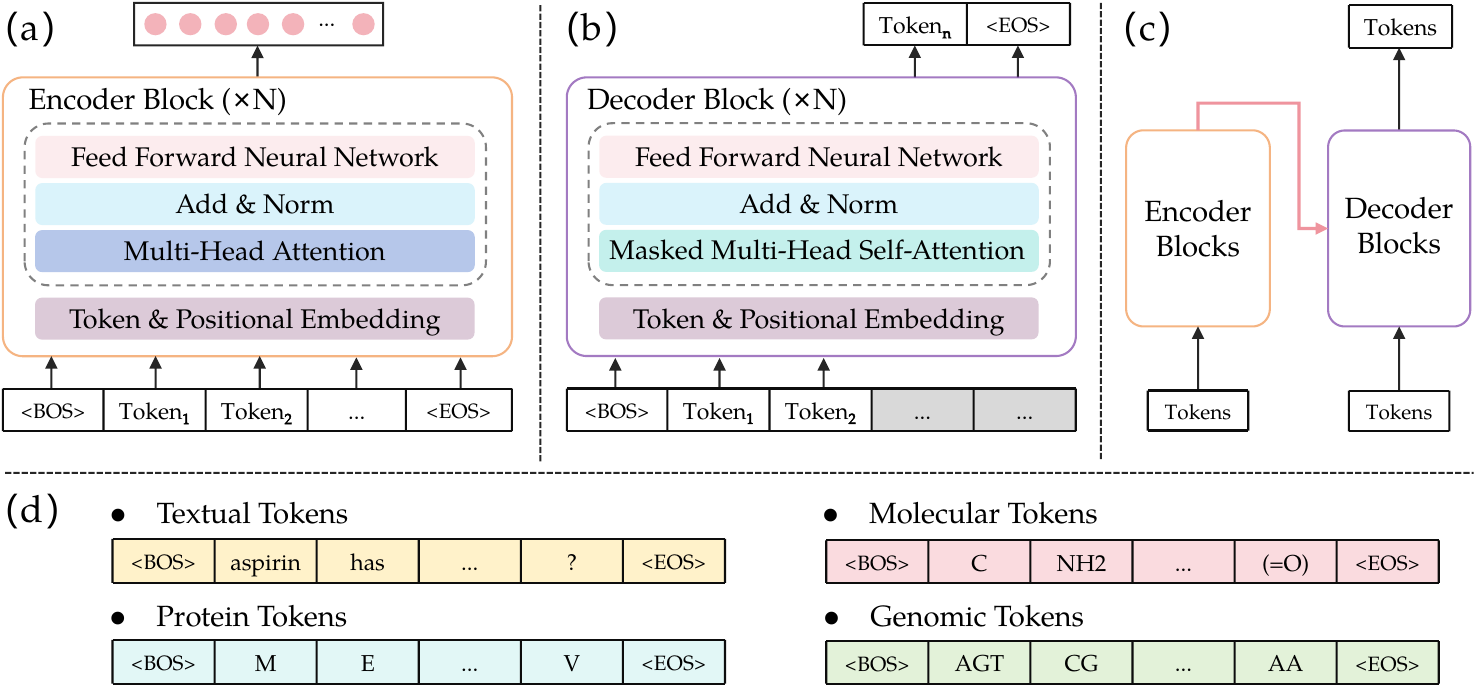}
    \caption{Illustration of common architectures of Sci-LLMs, including (a) encoder-only, (b) decoder-only, and (c) encoder\&decoder-based models, with (d) representing the tokens of scientific languages.}
    \vspace{-1em}
    \label{fig:model_arch}
\end{figure}

\subsection{Pre-training and Fine-tuning}\label{sec:train}
The training of LLMs, particularly those designed for the scientific domain, typically involves a two-stage process: pre-training and fine-tuning. 

\paratitle{Pre-training}.
During the pre-training phase, the model is exposed to a vast and diverse corpus of natural and scientific language. For instance, in the natural language-based Sci-LLMs, this corpus may be heavily weighted toward scientific literature, including research papers, academic journals, and technical documents, to imbue the model with a strong grasp of scientific terminology and concepts. 
The objective function of pre-training is dependent on the chosen model architectures. For example, Sci-LLMs based on BERT~\cite{Devlin-NAACL-2019-BERT} often employ masked language modeling, where the goal is to predict the original identity of these masked tokens. This approach allows Sci-LLMs to learn a bidirectional representation of language by considering the contextual information from both sides of the masked token.
For Sci-LLMs based on GPT~\cite{OpenAI-blog-2022-ChatGPT},  the objective can be causal (or autoregressive) language modeling, which predicts the next token given all the previous tokens.

\paratitle{Fine-tuning}.
The fine-tuning process is crucial for adapting pre-trained language models to specific tasks, such as function prediction and targeted sequence generation. During this stage, Sci-LLMs are fine-tuned with varying objectives and datasets that contain labels. The choice of objective functions also relies on model architectures and specific downstream tasks at hand. For instance, in the chemical reaction classification task, the fine-tuning dataset would consist of reaction samples annotated with type labels, and the optimization objective is to minimize the cross-entropy between the output and ground-truth. For generation tasks, the fine-tuning datasets typically comprise textual instruction data, consisting of pairs of input instructions and corresponding outputs that guide the model to comprehend specific tasks. The instruction data can be sourced from the benchmarks listed in Table \ref{tab:bench_text_LLM}, or generated from the scientific corpora as described in DARWIN \cite{xie2023darwin}. In this context, the objective of instruction fine-tuning could be causal language modeling, which is often used in GPT-based Sci-LLMs.

It should be noted that the pipeline of training in Sci-LLMs (including textual, molecular, protein, genomic and multimodal) is consistent with that of general LLMs; therefore, we will not cover the training part (except for training datasets) in their respective chapters.

\subsection{Notions and Terms} \label{sec:terms}
In this subsection, we clarify the key concepts that will be utilized throughout our discussion. We will not define the terminology like Transformer and pretraining as we assume the readers have a preliminary knowledge of LLMs.

\begin{tcolorbox}[colback=red!5!white, colframe=white]
\paratitle{Language}. Language is a sophisticated system of communication with well-defined vocabulary and grammatical rules.
It includes natural languages like English, used by humans, and non-natural languages such as programming languages crafted by computer scientists, SMILES notation used in chemistry, and amino acid sequences in biology.

\paratitle{Scientific Language}. Scientific Language is a specialized subset of language, crafted specifically to articulate scientific knowledge. In the context of this survey, we define scientific languages in two broad categories: one, as textual language that is focused on various science domains; and two, as domain-specific languages, which include the languages of molecules, proteins, and genomes. This bifurcation helps in understanding both the general and specialized aspects of language used in scientific discourse and research.

\paratitle{Large Language Models (LLMs)}. LLMs are large-scale neural networks with millions or even billions of trainable parameters, typically based on the Transformer architectures. These models are pre-trained with vast amounts of language data, enabling them to exhibit proficiency in comprehending, generating, and translating natural language.

\end{tcolorbox}

\begin{tcolorbox}[colback=red!5!white, colframe=white]

\paratitle{Scientific Large Language Models (Sci-LLMs)}. Sci-LLMs are a collective term of LLMs designed to understand, interpret, and generate scientific languages. In this survey, Sci-LLMs include textual, molecular, protein, genomic, and multimodal scientific LLMs.

\paratitle{Textual Scientific Large Language Models (Text-Sci-LLMs)}. Text-Sci-LLMs are a category of LLMs that are specifically trained on vast amounts of textual scientific data. These models excel in understanding, generating, and interacting with human language in its written form.

\paratitle{Molecular Large Language Models (Mol-LLMs)}. Mol-LLMs are specialized LLMs trained on molecular data, allowing them to understand and predict chemical properties and behaviors of molecules. This distinctive expertise renders them invaluable tools in drug discovery, materials science, and understanding complex chemical interactions.

\paratitle{Protein Large Language Models (Prot-LLMs)}. Prot-LLMs are trained specifically on protein-related data, including amino acid sequences, protein folding patterns, and other biological data relevant to proteins. Consequently, they possess the capability to accurately predict protein structures, functions, and interactions. 

\paratitle{Genomic Large Language Models (Gene-LLMs)}. Gene-LLMs, with a primary focus on genomic data, are trained to understand and predict aspects of genetics and genomics. They can be used to analyze DNA sequences, understand genetic variations, and assist in genetic research endeavors such as identifying genetic markers for diseases or understanding evolutionary biology.

\paratitle{Multimodal Scientific Large Language Models (MM-Sci-LLMs)}. MM-Sci-LLMs are the most advanced and versatile models, capable of processing and integrating multiple types of scientific data. They can handle text, molecules, proteins, and more, making them suitable for complex scientific research that spans different data types. This capability makes them invaluable in interdisciplinary research areas, where they can synthesize knowledge from various scientific domains to provide comprehensive insights.

\end{tcolorbox}

	\clearpage
	
	\section{Textual Scientific Large Language Models} \label{sec-text}

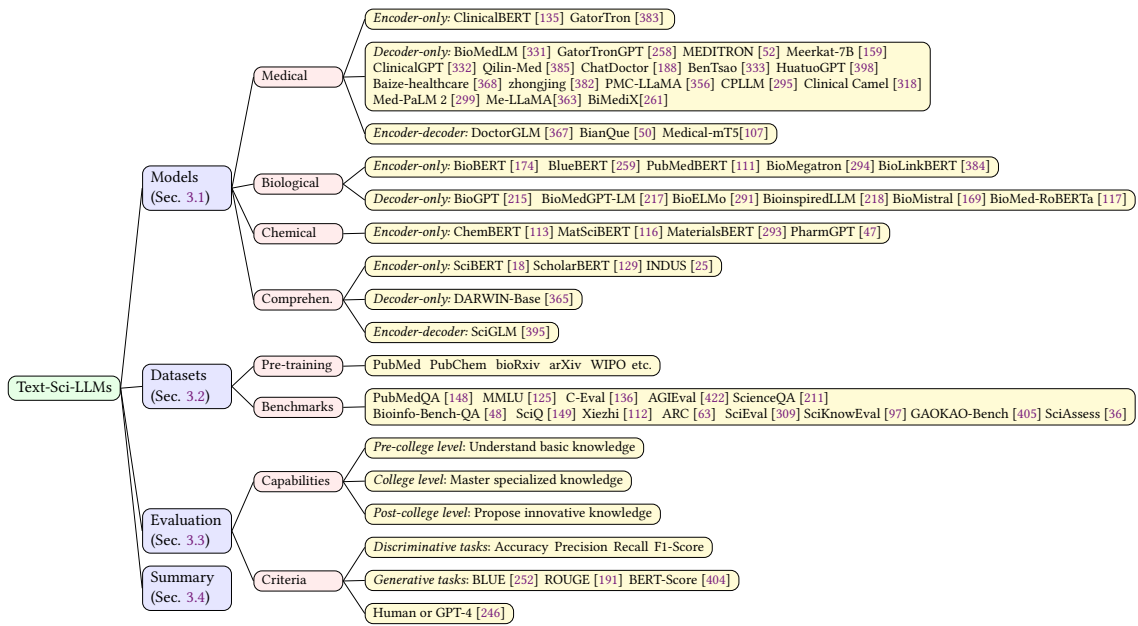
\begin{figure*}[h]
\centering
\resizebox{\textwidth}{!}{
\begin{forest}
  for tree={
  grow=east,
  reversed=true,
  anchor=base west,
  parent anchor=east,
  child anchor=west,
  base=left,
  font=\small,
  rectangle,
  draw,
  rounded corners,align=left,
  minimum width=2.5em,
  inner xsep=4pt,
  inner ysep=1pt,
  },
  where level=1{text width=5em,fill=blue!10}{},
  where level=2{text width=5em,font=\footnotesize,fill=pink!30}{},
  where level=3{font=\footnotesize,yshift=0.26pt,fill=yellow!20}{},
  [Text-Sci-LLMs, fill=green!10
        [Models\\(Sec.~\ref{sec:text-LLM-overview}),text width=4em
            [Medical, text width=4em
              [\emph{Encoder-only:}
              ClinicalBERT~\cite{huang2020clinicalbert}\,
              GatorTron~\cite{yang2022gatortron}\,
              ]
              [\emph{Decoder-only:}
              BioMedLM~\cite{venigalla2022biomedlm}\, 
              GatorTronGPT~\cite{Peng_2023}\, 
              MEDITRON~\cite{chen2023meditron70b}\,
              Meerkat-7B~\cite{kim2024Meerkat7B} \\
              ClinicalGPT~\cite{wang2023clinicalgpt}\,
              Qilin-Med~\cite{ye2023qilinmed}\, 
              ChatDoctor~\cite{li2023chatdoctor}\, 
              BenTsao~\cite{wang2023huatuo}\, 
              HuatuoGPT~\cite{zhang2023huatuogpt} \\ 
              Baize-healthcare~\cite{xu2023baize}\, 
              zhongjing~\cite{yang2023zhongjing}\,
              PMC-LLaMA~\cite{wu2023pmcllama}\, 
              CPLLM~\cite{shoham2023cpllm}\, 
              Clinical Camel~\cite{toma2023clinical}\\ 
              Med-PaLM 2~\cite{singhal2023expertlevel}\, 
              Me-LLaMA\cite{xie2024mellama}\,
              BiMediX\cite{pieri2024bimedix}\,
              ]
              [\emph{Encoder-decoder:}
              DoctorGLM~\cite{xiong2023doctorglm}\, 
              BianQue~\cite{chen2023bianque}\,
              Medical-mT5\cite{garcíaferrero2024medicalmT5}
              ]
            ]
            [Biological, text width=4em
              [\emph{Encoder-only:}
              BioBERT~\cite{lee2020biobert} \,   
              BlueBERT~\cite{peng2020empirical}\, 
              PubMedBERT~\cite{gu2021domain}\,    
              BioMegatron~\cite{shin2020biomegatron} 
              BioLinkBERT~\cite{yasunaga2022linkbert}
              ]
              [\emph{Decoder-only:} 
              BioGPT~\cite{Luo_2022} \, 
              BioMedGPT-LM~\cite{luo2023biomedgpt}
              BioELMo~\cite{sharma2019incorporating}
              BioinspiredLLM~\cite{luu2023bioinspiredllm}
              BioMistral~\cite{labrak2024biomistral}
              BioMed-RoBERTa~\cite{domains}
              ]
            ]
            [Chemical, text width=4em
              [\emph{Encoder-only:} 
              ChemBERT~\cite{guo2021automated} 
              MatSciBERT~\cite{gupta_matscibert_2022}
              MaterialsBERT~\cite{materialsbert}
              PharmGPT~\cite{chen2024pharmgpt}
              ]
            ]
            [Comprehen., text width=4em
              [
              \emph{Encoder-only:} 
              SciBERT~\cite{beltagy2019scibert}
              ScholarBERT~\cite{hong2023diminishing}
              INDUS~\cite{bhattacharjee2024indus}
              ]
              [\emph{Decoder-only:} 
              DARWIN-Base ~\cite{xie2023darwin} 
              ]
              [\emph{Encoder-decoder:} 
              SciGLM ~\cite{zhang2024sciglm}
              ]
            ]    
        ]
        [Datasets\\
        (Sec.~\ref{sec:text-LLM-dataset}),text width=4em
            [Pre-training, text width=4em
              [
              PubMed \,
              PubChem \,
              bioRxiv \,
              arXiv \,
              WIPO\,
              etc.
              ]
            ]
            [Benchmarks, text width=4em
              [PubMedQA~\cite{jin2019pubmedqa} \, MMLU~\cite{hendrycks2021measuring} \, C-Eval~\cite{huang2023ceval} \, AGIEval~\cite{zhong2023agieval} ScienceQA~\cite{lu2022learn}\\
              Bioinfo-Bench-QA~\cite{chen2023bioinfo} \, 
              SciQ~\cite{SciQ} \,Xiezhi~\cite{gu2023xiezhi} \, ARC~\cite{clark2018think} \, SciEval~\cite{sun2023scieval}
              SciKnowEval~\cite{feng2024sciknoweval}
              GAOKAO-Bench~\cite{Zhang2023EvaluatingTP}
              SciAssess~\cite{cai2024sciassess}
              ]
            ]      
        ]
        [Evaluation\\
        (Sec.~\ref{sec:text-LLM-eval}),text width=4em
            [Capabilities, text width=4em
              [\textit{Pre-college level}: Understand basic knowledge
              ]
              [\textit{College level}: Master specialized knowledge
              ]
              [\textit{Post-college level}: Propose innovative knowledge
              ]
            ]   
            [Criteria, text width=4em
              [
              \textit{Discriminative tasks}: Accuracy\, Precision\, Recall\, F1-Score \\
              ]
              [
               \textit{Generative tasks}: BLUE~\cite{papineni2002bleu}\, ROUGE~\cite{lin2004rouge}\, BERT-Score~\cite{zhang2019bertscore} 
              ]   
              [
               Human or GPT-4~\cite{OpenAI-OpenAI-2023-GPT-4}\
              ]  
            ]    
        ]
        [Summary\\
        (Sec.~\ref{sec:text-LLM-summary}),text width=4em
        ]
    ]
\end{forest}
}
\caption{Chapter overview of Text-Sci-LLMs.}
\label{fig:text-LLM-overview}
\end{figure*}

In this section, we aim to explore and delve into scientific large language models specifically trained using textual corpora (i.e., Text-Sci-LLMs), particularly emphasizing their acquisition of chemical and biological knowledge. We will briefly review the existing Text-Sci-LLMs and examine their capabilities, the employed datasets, as well as the evaluation approach. The overview of this section is shown in Figure~\ref{fig:text-LLM-overview}.

\subsection{Models}\label{sec:text-LLM-overview}
To provide a clear overview of Text-Sci-LLMs, we divide them into four distinct categories: medical, biological, chemical, and comprehensive, based on the specific scientific domains they focus on. Table \ref{tab:Text-Sci-LLMs} shows the summary of Text-Sci-LLMs.

\paratitle{Medical Domain.}
{Clinical medicine profoundly influences human health by diagnosing, treating, and preventing diseases. Recent years have witnessed remarkable advances in artificial intelligence technology, leading to the emergence of specialized large language models such as BioMedLM\cite{venigalla2022biomedlm}, GatorTron\cite{yang2022gatortron}, GatorTronGPT\cite{Peng_2023}, PMC-LLaMA\cite{wu2023pmcllama}, MEDITRON\cite{chen2023meditron70b} and Meerkat-7B\cite{kim2024Meerkat7B}. These models are trained with professional materials such as medical literature, clinical records, and medical textbooks, thus significantly advancing the medical industry.}

{Models like Med-PaLM 2\cite{singhal2023expertlevel}, HuatuoGPT\cite{zhang2023huatuogpt}, BianQue\cite{chen2023bianque}, ClinicalGPT\cite{wang2023clinicalgpt}, DoctorGLM\cite{xiong2023doctorglm}, chatDoctor\cite{li2023chatdoctor}, Baize-healthcare\cite{xu2023baize}, zhongjing\cite{yang2023zhongjing}, BenTsao\cite{wang2023huatuo}, Clinical Camel\cite{toma2023clinical}, and the comprehensive Me-LLaMA\cite{xie2024mellama} have facilitated more efficient communication between doctors and patients through training and fine-tuning in real-world medical records and doctor-patient dialogue datasets. Moreover, the landscape of medical AI has expanded with the introduction of multi-lingual models like BiMediX\cite{pieri2024bimedix} and Medical-mT5\cite{garcíaferrero2024medicalmT5}, which facilitate medical interactions across different languages, enhancing the accessibility of medical AI globally. }


{
The innovative Qilin-Med\cite{ye2023qilinmed} stands out with its three-stage training method, addressing the need for specialized knowledge in the medical field and offering insights into tailored training for professional domains. 
Additionally, models like ClinicalBERT\cite{huang2020clinicalbert} predict 30-day readmissions, while CPLLM\cite{shoham2023cpllm} forecasts future disease risks using patients' historical diagnoses, offering doctors scientific references for diagnosis and treatment decisions. Continuous advances in these technologies bring renewed optimism to the field of medicine and make substantial contributions to advancing human health endeavors.}

\paratitle{Biological Domain.}
Large language models trained on extensive biological corpora, such as BERT~\cite{Devlin-NAACL-2019-BERT} and its variations with the encoder-only LLM architecture, have demonstrated significant potential in Natural Language Processing (NLP) tasks within biology. BioBert~\cite{lee2020biobert}, BioMegatron~\cite{shin2020biomegatron}, PubMedBERT~\cite{gu2021domain}, BioM-BERT~\cite{alrowili-shanker-2021-biom}, BioELMo~\cite{sharma2019incorporating} and LinkBERT~\cite{yasunaga2022linkbert} initially trained on broad corpora like Wikipedia and textbooks and then fine-tuned on specific biological NLP tasks, showcase their significant improvements in various downstream tasks such as biological terminology understanding, named entity recognition, text similarity, and relation extraction. BlueBERT~\cite{peng2020empirical} presented a thorough empirical study on the efficacy of multi-task learning on BERT-based models for biomedical texts. Since the encoder-only models lack the capability of textual information generation, GPT and its variants~\cite{radford-openai-2018-improving, radford-blog-2019-language, Brown-NeurIPS-2020-Language} with a decoder-only architecture have emerged as dominant players in the field of biological NLP. BioGPT~\cite{Luo_2022}, an extension of GPT-2~\cite{radford2019GPT2}, has been extensively pretrained on biomedical literature, showcasing remarkable performance in relation extraction and question answering. Moreover, it demonstrates the ability to generate coherent and fluent descriptions within the biomedical context. BioMedGPT-LM~\cite{luo2023biomedgpt}, which is incrementally pretrained on LLaMA2~\cite{touvron2023llama}, enabling the comprehensive understanding of various biological modalities and aligning them with natural language.
BioinspiredLLM~\cite{luu2023bioinspiredllm}, based on the Llama2 model and enhanced with the Orca-2-13b model \cite{mitra2023orca}, specializes in elucidating the mechanics of biological and bio-inspired materials, exhibiting cognitive capabilities such as information retrieval, hypothesis generation, and collaborative interactions with other AI models.
BioMistral~\cite{labrak2024biomistral} is based on the Mistral model and further pre-trained on the corpus sourced from PubMed Central, thereby enhancing its performance in medical question-answering tasks. 
BioMed-RoBERTa~\cite{domains} is a language model based on the RoBERTa-base architecture. It is adapted through continued pretraining on 2.68 million scientific full-text papers from the Semantic Scholar corpus.

\begin{table*}[tbp]
\centering
\caption{Summary of Text-Sci-LLMs}
\label{tab:Text-Sci-LLMs}
\footnotesize
\renewcommand\tabcolsep{10pt}
\begin{tabular}{llllllc}
\toprule
 Domain & Model & Time  &  \#Parameters &   Base model  & Pretraining dataset  &\makecell{Open-\\source}\\
\midrule
 \multirow{27}{*}{Medical}  
 & \href{https://github.com/kexinhuang12345/clinicalBERT}{ClinicalBERT}~\cite{huang2020clinicalbert} & 2019.04 & 110M & BERT & MIMIC-III & \checkmark \\ 
 & \href{https://catalog.ngc.nvidia.com/orgs/nvidia/teams/clara/models/gatortron_og}{GatorTron}~\cite{yang2022gatortron} & 2022.02 & 8.9B & BERT & MIMIC-III, PubMed, etc.& \checkmark  \\ 
 & \href{https://huggingface.co/stanford-crfm/BioMedLM}{BioMedLM}~\cite{venigalla2022biomedlm} & 2022.12 & 2.7B & GPT & PubMed & \checkmark  \\ 
 & \href{https://github.com/uf-hobi-informatics-lab/GatorTronGPT}{GatorTronGPT}~\cite{Peng_2023} & 2023.05 & 5B, 20B & GPT & Pubmed, Custom Data & \checkmark  \\ 
 & \href{https://github.com/epfLLM/meditron}{MEDITRON}~\cite{chen2023meditron70b} & 2023.11 & 7B, 70B & LLAMA2 & Pubmed, PMC, etc. & \checkmark \\ 
 & \href{https://arxiv.org/abs/2404.00376}{Meerkat-7B}~\cite{kim2024Meerkat7B} & 2024.03 & 7B & Mistral & Custom Data & $\times$  \\ 
 & \href{https://arxiv.org/abs/2306.09968}{ClinicalGPT}~\cite{wang2023clinicalgpt} & 2023.06 & 7B & BLOOM & cMedQA2, MedDialog, etc. & $\times$  \\ 
 & \href{https://github.com/williamliujl/Qilin-Med/tree/master}{Qilin-Med}~\cite{ye2023qilinmed} & 2023.10 & 7B & Baichuan & ChiMed & \checkmark \\ 
 & \href{https://github.com/Kent0n-Li/ChatDoctor}{ChatDoctor}~\cite{li2023chatdoctor} & 2023.03 & 7B & LLAMA & HealthCareMagic & \checkmark \\ 
 & \href{https://github.com/SCIR-HI/Huatuo-Llama-Med-Chinese}{BenTsao}~\cite{wang2023huatuo} & 2023.04 & 7B & LLAMA & Custom Data & \checkmark  \\ 
 & \href{https://github.com/FreedomIntelligence/HuatuoGPT}{HuatuoGPT}~\cite{zhang2023huatuogpt} & 2023.05 & 7B & BLOOM & Custom Data & \checkmark \\ 
 & \href{https://github.com/project-baize/baize-chatbot}{Baize-healthcare}~\cite{xu2023baize} & 2023.04 & 7B & LLAMA & MedQuAD & \checkmark  \\ 
 & \href{https://github.com/SupritYoung/Zhongjing}{zhongjing}~\cite{yang2023zhongjing} & 2023.08 & 13B & LLAMA & Medical Books, Wiki, etc. & \checkmark  \\ 
 & \href{https://github.com/chaoyi-wu/PMC-LLaMA}{PMC-LLaMA}~\cite{wu2023pmcllama} & 2023.04 & 13B & LLAMA & PMC, Medical books, etc. & \checkmark  \\ 
 & \href{https://github.com/nadavlab/CPLLM}{CPLLM}~\cite{shoham2023cpllm} & 2023.09 & 2.7B, 13B & LLAMA-2 & eICU-CRD, MIMIC-IV & \checkmark  \\ 
 & \href{https://github.com/bowang-lab/clinical-camel}{Clinical Camel}~\cite{toma2023clinical} & 2023.05 & 13B, 70B & LLAMA-2 & PubMed & $\times$ \\ 
 & \href{https://sites.research.google/med-palm/}{Med-PaLM 2}~\cite{singhal2023expertlevel} & 2023.05 & 340B & PaLM2 & MultiMedQA & $\times$ \\ 
 & \href{https://github.com/xionghonglin/DoctorGLM}{DoctorGLM}~\cite{xiong2023doctorglm} & 2023.04 & 6.2B & ChatGLM & CMD., HealthCareMagic, etc. & \checkmark  \\ 
 & \href{https://github.com/scutcyr/BianQue}{BianQue}~\cite{chen2023bianque} & 2023.10 & 6.2B & ChatGLM & BianQueCorpus & \checkmark \\ 
 & \href{https://arxiv.org/pdf/2404.07613}{Medical-mT5}~\cite{garcíaferrero2024medicalmT5} & 2024.01 & 738M, 3B & T5 & Custom Data & $\times$  \\ 
 & \href{https://github.com/BIDS-Xu-Lab/Me-LLaMA}{Me-LLaMA}~\cite{xie2024mellama} & 2024.02 & 13B, 70B & LLAMA & Custom Data & \checkmark  \\
 & \href{https://github.com/mbzuai-oryx/BiMediX}{BiMediX}~\cite{pieri2024bimedix} & 2024.02 & 7B & Mixtral & BiMed1.3M & \checkmark  \\ 
 \midrule
 &\href{https://github.com/Andy-jqa/bioelmo}{BioELMo}~\cite{jin2019probing} & 2019.04& - & ELMo & PubMed & \checkmark \\

 \multirow{10}{*}{Biology}  &\href{https://github.com/dmis-lab/biobert}{BioBERT}~\cite{lee2020biobert}  &   2019.05  & 117M  &  BERT   & PubMed, PMC & \checkmark\\
  &  \href{https://github.com/ncbi-nlp/bluebert}{BlueBERT}~\cite{peng2020empirical}& 2019.07& 117M &  BERT &PubMed & \checkmark\\
  &\href{ngc.nvidia.com}{BioMegatron}~\cite{shin2020biomegatron}  &   2020.10  & 345M-1.2B& BERT &PubMed, PMC & \checkmark\\ 
 & \href{https://huggingface.co/microsoft/BiomedNLP-BiomedBERT-base-uncased-abstract}{PubMedBERT}~\cite{gu2021domain}&   2020.10  & 117M& BERT &PubMed & $\times$ \\
& \href{https://github.com/salrowili/BioM-Transformers}{BioM-BERT}\cite{alrowili-shanker-2021-biom} & 2021.06 &  235M &  BERT & PubMed, PMC & \checkmark \\
 &\href{https://github.com/michiyasunaga/LinkBERT}{BioLinkBERT}\cite{yasunaga2022linkbert}& 2022.03& 110M, 340M&  BERT& PubMed & \checkmark \\
 & \href{https://github.com/microsoft/BioGPT}{BioGPT}~\cite{Luo_2022}& 2023.03& 347M& GPT &PubMed &\checkmark\\
 &\href{https://github.com/PharMolix/OpenBioMed}{BioMedGPT-LM}~\cite{luo2023biomedgpt}& 2023.08& 7B& LLaMA & PMC, arXiv, WIPO &\checkmark\\

 &\href{https://huggingface.co/lamm-mit/BioinspiredLLM}{BioinspiredLLM}~\cite{luu2023bioinspiredllm}& 2023.09& 13B& Llama-2 & Biological article  &\checkmark\\
 
 &\href{https://github.com/BioMistral/BioMistral}{BioMistral}~\cite{labrak2024biomistral}& 2024.02& 7B& Mistral & PMC &\checkmark\\

 \midrule
 \multirow{4}{*}{Chemistry} &\href{https://github.com/jiangfeng1124/ChemRxnExtractor}{ChemBERT}~\cite{guo2021automated}& 2021.06& 120M& BERT &Chemical journals &\checkmark\\
 & \href{https://github.com/M3RG-IITD/MatSciBERT}{MatSciBERT} ~\cite{gupta_matscibert_2022} & 2021.09 &  117M  & BERT & Elsevier journals &  \checkmark \\
& \href{https://huggingface.co/pranav-s/MaterialsBERT}{MaterialsBERT} ~\cite{materialsbert} & 2022.09 & - &  BERT & Material journals & \checkmark  \\



& PharmGPT~\cite{chen2024pharmgpt} & 2024.02 & 13B, 70B &  LLaMA & Paper, report, book, etc.& $\times$  \\

 \midrule
\multirow{3}{*}{Comprehensive} &\href{https://github.com/allenai/scibert/}{SciBERT}~\cite{beltagy2019scibert}  &  2019.09   & 117M&  BERT   &Semantic Scholar &\checkmark\\
  &\href{https://huggingface.co/globuslabs/ScholarBERT}{ScholarBERT}~\cite{hong2023diminishing}  & 2023.05 & 340M, 770M &BERT & Wiki, Books, etc. & \checkmark \\
  & \href{https://github.com/MasterAI-EAM/Darwin}{DARWIN-Base} \cite{xie2023darwin}& 2023.08& 7B&LLaMA& SciQ, Web of Science &\checkmark\\

& \href{https://github.com/THUDM/SciGLM}{SciGLM} \cite{zhang2024sciglm}& 2024.01& 6B, 32B & ChatGLM3 & SciInstruct  &\checkmark\\

& \href{https://uni-smart.dp.tech/}{Uni-SMART} \cite{cai2024uni}& 2024.03& 7B& - & Patents, news, literature, etc.& $\times$\\

  & INDUS \cite{bhattacharjee2024indus}& 2024.05& 125M & RoBERTa& wikipedia, PubMed, PMC, etc. &$\times$\\
  
  \bottomrule
\end{tabular}
\end{table*}

\paratitle{Chemical Domain.}
In the realm of chemistry, natural language is commonly used to express various attributes and discoveries related to compounds and chemical reactions, making LLMs well-suited for comprehending and reasoning about chemical information~\cite{bran2023transformers}. Built upon the BERT architecture, ChemBERT~\cite{guo2021automated} was pretrained on an extensive dataset comprising 200,000 chemical journal articles. This model is tailored for automated chemical reaction extraction from chemical literatures, employing Masked Language Modeling (MLM) for pretraining and fine-tuning for chemical NLP tasks. 
MatSciBERT \cite{gupta_matscibert_2022} and MaterialsBERT \cite{materialsbert} are materials science-specific models, trained on a large corpus of material literature, and show cutting-edge performance in extracting material property information. It is worth noting that many works utilized ChatGPT \cite{ouyang2022training} or GPT-4 \cite{OpenAI-OpenAI-2023-GPT-4} to extract valuable chemical information from literature, such as synthesis conditions and compound properties. For example, Zheng \textit{et al.}~\cite{Zheng_2023MOF} demonstrated the potential of ChatGPT for text mining and prediction of MOF Synthesis. ChemCrow~\cite{bran2023chemcrow} augmented the performance of ChatGPT in chemistry by incorporating 13 expert-designed tools, specifically tailored for accomplishing tasks related to organic synthesis, drug discovery, and materials design. 

\paratitle{Comprehensive Domain.}
This part introduces general Text-Sci-LLMs, which span diverse domains including biology, chemistry, and other scientific disciplines. 
SciBERT~\cite{beltagy2019scibert} is built upon the BERT architecture and utilizes unsupervised pretraining on a vast corpus of scientific publications to improve performance on downstream scientific NLP tasks. ScholarBERT~\cite{hong2023diminishing} is a series of BERT-based scientific language models trained on a substantial and diverse scientific corpus spanning domains including life sciences, biomedicine and physical sciences. 
INDUS~\cite{bhattacharjee2024indus} is a suite of language models tailored for scientific applications in domains such as Earth science, biology, physics, heliophysics, planetary sciences, and astrophysics, trained using curated scientific corpora. It includes encoder models for natural language understanding tasks and contrastive-learning-based text embedding models for information retrieval, with smaller, efficient versions created using knowledge distillation techniques.
DARWIN-Base~\cite{xie2023darwin}, based on a decoder-only architecture, is trained on a massive corpus from various scientific texts and knowledge bases. Notably, it was fine-tuned with instruction data from the SciQ~\cite{welbl2017crowdsourcing} dataset and additional question-answer pairs, exhibiting proficiency in tasks that require extensive scientific knowledge. We emphasize that, although the general LLMs such as ChatGPT~\cite{ouyang2022training}, GPT-4~\cite{OpenAI-OpenAI-2023-GPT-4} and Gemini~\cite{geminiteam2023gemini} encompass a wealth of scientific knowledge, they are not tailored for Sci-LLMs, and therefore do not fall under the category of general Text-Sci-LLMs.
SciGLM~\cite{zhang2024sciglm} is a suite of scientific language models fine-tuned from the ChatGLM family, using a novel self-reflective instruction annotation framework. This innovative method generates high-quality reasoning data, enabling SciGLM to excel in scientific reasoning and numerical calculations. 
Uni-SMART~\cite{cai2024uni} is a model designed to interpret and analyze multimodal scientific literature, including text, tables, charts, and molecular structures. It uses key methods such as multimodal learning, iterative training, and self-reflective instruction annotation to enhance performance in scientific tasks. 


\subsection{Datasets}\label{sec:text-LLM-dataset}
According to the availability of annotations, datasets can be categorized into two types: unlabeled data and labeled data. The former is often used for self-supervised pre-training, while the latter is utilized for supervised fine-tuning or model evaluation.

\paratitle{Pre-training Datasets.}
In Text-Sci-LLMs, the pre-training corpora are mainly sourced from various scientific articles. We list several widely-used data sources as follows.
\begin{itemize}
    \item PubMed\footnote{\url{https://pubmed.ncbi.nlm.nih.gov/}}is a free and open-access database in the field of biomedical and life sciences, which contains more than 32 million citations and abstracts of biomedical literature. It is managed by the National Center for Biotechnology Information (NCBI) at the U.S. National Library of Medicine.
    \item PubMed Central (PMC)\footnote{\url{https://www.ncbi.nlm.nih.gov/pmc/}}  is a free full-text archive of biomedical and life sciences journals, which has an extensive collection of over 8 million articles. It can be seen as a complementary resource to PubMed, focusing on the storage and free accessibility of full-text papers.
    \item PubChem\footnote{\url{https://pubchem.ncbi.nlm.nih.gov/}}  is a database of chemical molecules and their activities against biological assays. This database includes information on chemical structures, identifiers, chemical and physical properties, biological activities, patents, health, safety, toxicity data, and more. 
    \item bioRxiv\footnote{\url{https://www.biorxiv.org/}} is an open-access preprint repository for biological sciences, which contains over 2 million full-text papers. This repository covers a wide range of topics in biology, including genetics, neuroscience, bioinformatics, and ecology.
    \item arXiv\footnote{\url{https://arxiv.org/}} is an online repository that provides open access to over 2 million scholarly articles in various disciplines including physics, mathematics, computer science, quantitative biology, quantitative finance, statistics, electrical engineering and systems science, and economics. 
    \item Semantic Scholar\footnote{\url{https://www.semanticscholar.org/}} is an advanced, free, AI-powered research tool for scientific literature that provides access to  200 million academic papers sourced from publisher partnerships, data providers, and web crawls. 
    \item World Intellectual Property Organization (WIPO)\footnote{\url{https://www.wipo.int/}} serves as a global forum for intellectual property services, policy, information, and cooperation. It maintains an extensive database of patent information that can be accessed by researchers and inventors around the world. 
    \item Web of Science\footnote{\url{https://www.webofscience.com}}   is a paid-access platform that offers access to diverse databases containing reference and citation data from academic journals, conference proceedings, and other scholarly documents across multiple academic disciplines.
    {
    \item The Medical Information Mart for Intensive Care (MIMIC) series, including MIMIC-III~\cite{Johnson2016MIMICIII}, MIMIC-IV~\cite{Johnson2023MIMICIV}, MIMIC-CXR~\cite{Johnson2019MIMICCXR}, and MIMIC-IV-Note~\cite{johnson2023mimicivnote}, are publicly available medical databases derived from the Beth Israel Deaconess Medical Center. 
    \begin{itemize}
        \item MIMIC-III~\cite{Johnson2016MIMICIII} is a de-identified database containing data from over 40,000 critically ill patients between 2001 and 2012, and it has been extensively utilized in research areas such as epidemiology, machine learning, and clinical informatics. 
        \item MIMIC-IV~\cite{Johnson2023MIMICIV} represents an updated version of MIMIC-III~\cite{Johnson2016MIMICIII}, offering medical data from 2008 to 2019, further enhancing the depth and breadth of healthcare research. 
        \item MIMIC-CXR~\cite{Johnson2019MIMICCXR} focuses on chest X-ray imaging and includes 377,110 chest X-ray images along with their corresponding radiological text reports. It has leveraged natural language processing tools to extract 14 labels from these reports, providing a significant impetus for the development of medical computer vision technology. 
        \item MIMIC-IV-Note~\cite{johnson2023mimicivnote}, contains de-identified discharge summaries for 331,794 patients, totaling 145,915 documents, and radiological reports for 2,321,355 patients, totaling 237,427 documents, offering a rich dataset for clinical research in hospitals and emergency departments. 
    \end{itemize} 
    \item The eICU Collaborative Research Database (eICU-CRD)~\cite{pollard2018eicu} is a publicly available, multi-center critical care database jointly created by Philips Healthcare and the Laboratory for Computational Physiology at the Massachusetts Institute of Technology (MIT). It contains detailed medical data from over 200,000 ICU patients in the United States. 
    \item The cMedQA2~\cite{zhang2018cMedQA2} dataset is a Chinese medical question-and-answer dataset that is aggregated from a Chinese medical question-and-answer online forum\footnote{\url{http://www.xywy.com/}}, consisting of 120,000 questions and 226,000 answers.
    \item MedDialog~\cite{he2020meddialog} is a large-scale medical dialogue dataset comprising 1.1 million conversations and 4 million utterances between patients and doctors.
    \item The ChiMed~\cite{ye2023qilinmed} dataset is a large-scale Chinese dataset designed for training Large Language Models (LLMs) in the medical field. It includes various types of data such as medical question-and-answer, texts, knowledge graphs, and dialogues. The dataset is divided into three stages: Domain-Specific Continuous Pre-training (DCPT), Supervised Fine-Tuning (SFT), and Direct Preference Optimization (DPO).
    \item The HealthCareMagic-100k is a dataset composed of approximately one million real-life doctor-patient conversations, sourced from the online medical consultation website HealthCareMagic\footnote{\url{www.healthcaremagic.com}}.
    \item The MedQuAD~\cite{ben2019question} dataset is for medical question-and-answer systems and supports Recognizing Question Entailment (RQE) methods. It has 47,457 medical question-and-answer sets with annotations from trusted NIH-related sources.
    \item MultiMedQA~\cite{singhal2023large} is a comprehensive benchmark dataset for medical question-answering that integrates multiple existing datasets from medical examinations, medical research, and consumer health inquiries, and also includes the newly added HealthSearchQA dataset, which contains 3,173 commonly searched health-related questions from users.
    \item The OpenI~\cite{demner2016preparing} dataset collected 8,121 images from the Picture Archiving Systems (PACS) of Indiana University Hospitals, along with 3,996 corresponding radiology reports. To improve the relevance of the retrieved clinical documents, manual coding was also added to the radiologists' reports.
    \item Psych8k~\cite{liu2023chatcounselor} is a dataset designed for the field of psychological counseling, based on 260 in-depth individual interviews, each lasting one hour, with an average of 31 question-and-answer pairs per individual. It covers a variety of mental health topics, including anxiety, depression, stress, and interpersonal relationships, and also takes into account minority groups.
    \item The CMD. dataset~\cite{xiong2023doctorglm} is a multi-departmental Chinese medical dialogue dataset that compiles medical consultation records from various departments including surgery, obstetrics and gynecology, pediatrics, internal medicine, and breast surgery, totaling over 900,000 question-and-answer pairs.
    \item BianQueCorpus~\cite{chen2023bianque} is a dataset of multi-turn questioning and health advice that has been polished by ChatGPT. It contains over 2.4 million samples, with questions accounting for 46.2\% of the doctor's responses, aimed at helping the model better understand patient conditions and provide targeted suggestions.
    }

    \item ChemData\cite{zhang2024chemllm} is a large-scale dataset for tuning language models in chemistry, encompassing nine core chemistry tasks and featuring 730,000 high-quality questions and answers, sampled from a subset of 700,000 data pieces.

    \item SciInstruct~\cite{zhang2024sciglm} is a comprehensive scientific instruction tuning dataset. It contains 254k data entries including physics, chemistry problems, math, and formal proofs.

\end{itemize}

\paratitle{Benchmarks.}
We curate a collection of benchmarks for Text-Sci-LLMs, as presented in Table \ref{tab:bench_text_LLM}. These benchmarks include multiple subsets that can be utilized to evaluate LLMs' abilities in grasping scientific knowledge at different levels within the biological, chemical and comprehensive domains. 
\begin{itemize}

    { 
    \item \textbf{MedQA-USMLE}~\cite{jin2020disease} dataset covers USMLE exam questions collected from the web, comprising a total of 10,178/1,272/1,273 train/dev/test standard questions that describe medical scenarios, which doctors should be able to answer, along with four possible answer choices provided for selection. 
    \item \textbf{MedMCQA}~\cite{pal2022medmcqa} is a large-scale multisubject multiple-choice dataset, comprising over 194,000 high-quality questions from Indian medical entrance exams like AIIMS and NEET. This dataset addresses 2.4k healthcare topics in 21 medical topics.
    \item \textbf{JAMA Clinical Challenge dataset}~\cite{chen2024benchmarking}. The JAMA Clinical Challenge dataset is a medical question-answer dataset composed of 1524 real and challenging clinical cases, covering 13 medical fields, with each case including a question, four answer choices, and detailed explanations provided by experts.
    \item \textbf{CMtMedQA}~\cite{yang2023zhongjing} is a Chinese medical dialogue dataset consisting of 70,000 authentic multiturn doctor-patient conversations across 14 medical departments, designed to enhance the dialogue and proactive inquiry capabilities of large language models in the field of traditional Chinese medicine.
    \item \textbf{huatuo-26M}~\cite{li2023huatuo26m} is a large-scale Chinese medical question-answering (QA) dataset, containing 26 million QA pairs. The dataset primarily sources from online medical consultation websites, medical encyclopedias, and medical knowledge bases.
    }

    \item \textbf{MMLU} (Massive Multitask Language Understanding)~\cite{hendrycks2021measuring}. MMLU offers a detailed and challenging benchmark that tests the comprehension and problem-solving capabilities of LLMs across a wide spectrum of tasks and subjects. Specifically, it covers 57 subjects, including STEM (Science, Technology, Engineering, Mathematics), humanities, social sciences, and other fields. Moreover, it ranges in difficulty from elementary to advanced professional levels. We can single out the subjects of biology and chemistry at the high school and college levels to evaluate Text-Sci-LLMs. 
    \item \textbf{C-Eval}~\cite{huang2023ceval}. C-Eval consists of 13,948 multi-choice questions spanning 52 diverse disciplines and four difficulty levels: middle school, high school, college, and professional level. In terms of scientific domains of interest, C-Eval covers the knowledge of chemistry and biology from middle school to undergraduate levels. 
    \item \textbf{AGIEval}~\cite{zhong2023agieval}. AGIEval is derived from rigorous and authoritative human-centric examinations, such as college entrance exams and professional certification tests. These exams ensure a robust and standardized evaluation of language models. AGI encompasses knowledge at both the pre-college and college levels. However, in the specific domains of biology and chemistry, its scope is limited to pre-college level knowledge.
    \item \textbf{ScienceQA}~\cite{lu2022learn}. ScienceQA is a benchmark designed to evaluate the ability and interpretability of LLMs in answering multimodal multiple-choice science questions. It consists of \textasciitilde21k questions covering diverse science topics, including biology and chemistry, as well as other scientific fields.

    \item \textbf{GAOKAO-Bench}~\cite{Zhang2023EvaluatingTP} is a benchmark designed to evaluate the performance of large language models (LLMs) using questions from the Chinese GAOKAO examination. It includes both subjective and objective questions and aims to provide a comprehensive and human-aligned assessment of LLM capabilities across various subjects.

    \item \textbf{Xiezhi}~\cite{gu2023xiezhi}. Xiezhi is a comprehensive evaluation suite for LLMs. It consists of 249,587 multi-choice questions spanning 516 diverse disciplines, which is sourced from a variety of examinations, including those conducted at the elementary school level, middle school entrance evaluations, college entrance assessments, undergraduate examinations, graduate entrance tests, and adult education exams. Xiezhi encompasses a wide range of scientific subjects,  including physics, chemistry, geography, mathematics, physics, biology, medicine, providing over 30k data. 
    \item \textbf{SciEval}~\cite{sun2023scieval}. SciEval consists of 18,000 challenging scientific questions across chemistry, physics, and biology, assessing LLMs' capabilities in basic knowledge, knowledge application, scientific calculation, and research ability. It combines objective and subjective questions and dynamically generates data based on basic scientific principles.    
    \item \textbf{Bioinfo-Bench-QA}~\cite{chen2023bioinfo}. Bioinfo-Bench-QA focuses on evaluating the advanced abilities of foundational models within the context of bioinformatics. It includes 1,500 multiple-choice questions generated by ChatGPT using authoritative bioinformatics literature.    
    \item \textbf{BLURB}~\cite{gu2021domain} focuses on PubMed-based biomedical applications, encompassing a comprehensive compilation of publicly available datasets for various biomedical NLP tasks. These tasks span diverse domains such as named entity recognition, evidence-based medical information extraction, relation extraction, sentence similarity, document classification, and question answering.        
    \item \textbf{PubMedQA}~\cite{jin2019pubmedqa}. PubMedQA is a biomedical Question-Answering (QA) dataset collected from PubMed, which has 1k expert-annotated, 61.2k unlabeled and 211.3k artificially generated QA instances. 
    \item \textbf{ARC}~\cite{clark2018think}. ARC is comprised of 7,787 questions in the field of natural science, specifically questions that were created for standardized tests. ARC has been divided into two sets: the Challenge Set (consisting of 2,590 questions) and the Easy Set (consisting of 5,197 questions), corresponding to the proposed pre-college and college levels, respectively.     
     \item \textbf{SciQ}~\cite{SciQ}. SciQ comprises a collection of 13,679 science exam questions on various subjects such as Chemistry, Biology, and more. These questions are presented in a multiple-choice format, offering four answer options for each question. To enhance understanding and knowledge, the majority of the questions are accompanied by an additional paragraph that provides supporting evidence for the correct answer.

     \item \textbf{SciBench}~\cite{wang2023scibench} is an academic benchmark encompassing physics, chemistry, and mathematics. It comprises a comprehensive set of 695 problems derived from instructional textbooks, specifically tailored for college-level scientific problem-solving. The primary objective of this benchmark is to assess complex reasoning capabilities, strong domain knowledge, and advanced calculation skills of LLMs.

    \item \textbf{SciAssess}~\cite{cai2024sciassess} is a benchmark designed to evaluate LLMs' proficiency in scientific literature analysis across three levels: Memorization, Comprehension, and Analysis \& Reasoning, using tasks from diverse scientific fields. It aims to thoroughly assess LLM capabilities and support their development in scientific literature analysis.

    \item \textbf{SciKnowEval}~\cite{feng2024sciknoweval}  is a benchmark for evaluating large language models' scientific knowledge and abilities across five progressive levels, including knowledge coverage, inquiry and exploration capabilities, reflection and reasoning abilities, ethic and safety considerations, as well as practice proficiency. The SciKnowEval dataset includes 50K biology and chemistry-related problems along with their corresponding solutions. 

\end{itemize}

\begin{table*}[tbp]
\centering
\caption{Summary of the benchmarks for Text-Sci-LLMs}
\label{tab:bench_text_LLM}
\footnotesize
\begin{tabular}{lllrllll}
\toprule
  Dataset  & Last updated & Subset&\#Item&  Domain& Type & Capability & Language\\ 
\midrule
\href{https://github.com/jind11/MedQA}{MedQA-USMLE}~\cite{jin2020disease}&2021.04&General Medicine&61.0K&Medical&Multiple choice&College&Both\\
\href{https://github.com/medmcqa/medmcqa}{MedMCQA}~\cite{pal2022medmcqa}&2022.03&General Medicine&193.1K&Medical&Multiple choice&College&English\\
\href{https://github.com/HanjieChen/ChallengeClinicalQA}{JAMA Clinical}~\cite{chen2024benchmarking}&2024.03&General Medicine&1524&Medical&Multiple choice&College&English\\ 
\href{https://huggingface.co/datasets/Suprit/CMtMedQA}{CMtMedQA}~\cite{yang2023zhongjing}&2023.08&General Medicine&68K&Medical&Multi-turn Dialogue&College&Chinese\\
\href{https://github.com/FreedomIntelligence/Huatuo-26M}{huatuo-26M}~\cite{li2023huatuo26m}&2023.03&General Medicine&26M&Medical&QA&College&Chinese\\

 \midrule
\multirow{4}{*}{\href{https://huggingface.co/datasets/cais/mmlu}{MMLU}~\cite{hendrycks2021measuring}} &  \multirow{4}{*}{2020.09}&  High-school-biology& 344& Biology& \multirow{4}{*}{Multiple choice} &Pre-college&\multirow{4}{*}{ English}\\
                        & & High-school-chemistry & 227& Chemistry &      &Pre-college& \\
                        & & College-biology &  162& Biology &     & College&\\
                        &  &College-chemistry&  110& Chemistry &  &College&\\ 
 \midrule
\multirow{5}{*}{\href{https://huggingface.co/datasets/ceval/ceval-exam}{C-Eval}~\cite{huang2023ceval}}& \multirow{5}{*}{2023.05}  &Mid-school-biology & 218  & Biology &\multirow{5}{*}{Multiple choice}& Pre-college& \multirow{5}{*}{Chinese}\\
                       & &Mid-school-chemistry & 210  & Chemistry &  & Pre-college&\\
                     &  & High-school-biology &  199  & Biology & & Pre-college&\\
                    &  &High-school-chemistry & 196  & Chemistry &  & Pre-college&\\
                    &  &College-chemistry   &   253  & Chemistry &  & College&\\
  \midrule
  \multirow{2}{*}{\href{https://huggingface.co/datasets/baber/agieval}{AGIEval}~\cite{zhong2023agieval}}&\multirow{2}{*}{2023.04} & Gaokao-biology & 210  & Biology  & \multirow{2}{*}{Multiple choice} & \multirow{2}{*}{Pre-college} & \multirow{2}{*}{Chinese}  \\
                                                &  &Gaokao-chemistry & 207  & Chemistry  &                                &        &  \\
\midrule
 \multirow{2}{*}{\href{https://huggingface.co/datasets/derek-thomas/ScienceQA}{ScienceQA}~\cite{lu2022learn}} &  \multirow{2}{*}{2022.09} & Natural-science-biology & 4098  & Biology & \multirow{2}{*}{Multiple choice / QA} & \multirow{2}{*}{Pre-college} & \multirow{2}{*}{English}\\
   &   & Natural-science-chemistry & 1194  & Chemistry &   & & \\  
  \midrule
  \multirow{2}{*}{\href{https://github.com/MikeGu721/XiezhiBenchmark}{XieZhi}~\cite{gu2023xiezhi}}&  \multirow{2}{*}{2023.06}&Science-biology& 2831 & Biology &\multirow{2}{*}{Multiple choice}& \multirow{2}{*}{Mixed} &\multirow{2}{*}{Both}   \\
                          &   &Science-chemistry&  399 &Chemistry & &   \\
  \midrule   
\multirow{7}{*}{\href{https://github.com/OpenDFM/SciEval}{SciEval}~\cite{sun2023scieval}} & \multirow{7}{*}{2023.08} & Basic-biology & 2142 & Biology& \multirow{7}{*}{Multiple choice / QA} & \multirow{7}{*}{Mixed}  &\multirow{7}{*}{English} \\
 &  & Knowledge-biology & 1369 & Biology&    &   &   \\
 &  & Calculation-biology & 299 & Biology&    &   &   \\
  &  & Research-biology & 995 & Biology&    &   &   \\
   &  & Basic-chemistry & 2909 & Chemistry&    &   &   \\
   &  & Knowledge-chemistry & 1700 & Chemistry&    &   &   \\
 &  & Calculation-chemistry & 3396 & Chemistry&    &   &   \\
\midrule

\multirow{2}{*}{\href{https://github.com/OpenLMLab/GAOKAO-Bench}{GAOKAO-Bench}~\cite{Zhang2023EvaluatingTP}} & \multirow{2}{*}{2023.11} & Biology & 266  & Biology& \multirow{2}{*}{Multiple choice / QA}  & \multirow{2}{*}{Mixed}  &\multirow{2}{*}{Chinese} \\
 &  & Chemistry & 133 & Chemistry&    &   &   \\
\midrule

\multirow{2}{*}{\href{https://github.com/hicai-zju/sciknoweval}{SciKnowEval}~\cite{feng2024sciknoweval}} & \multirow{2}{*}{2024.06} & Biology & 27730 & Biology& 
\multirow{2}{*}{\makecell{Multiple choice / QA \\ / True or flase}} & 
\multirow{2}{*}{Mixed}  &\multirow{2}{*}{English} \\
 &  & Chemistry & 22250 & Chemistry&    &   &   \\
\midrule   


\href{https://github.com/cinnnna/bioinfo-bench}{Bioinfo-Bench-QA}~\cite{chen2023bioinfo} & 2023.10 & - & 150 & Biology& Multiple choice &   Post-college & \\
\href{https://huggingface.co/datasets/EMBO/BLURB}{BLURB}~\cite{gu2021domain} &2020.07 & - &648k & Biology & Multiple NLP tasks& Mixed &  \multirow{5}{*}{English}\\
\href{https://huggingface.co/datasets/pubmed_qa}{PubMedQA}~\cite{jin2019pubmedqa}& 2019.09& - & 273.2k   & Biology & True or false &  College & \\

\href{https://github.com/mandyyyyii/scibench}{SciBench}~\cite{wang2023scibench}&2023.07&-&272&Chemistry&QA&College\\

 \href{https://huggingface.co/datasets/ai2_arc}{ARC}~\cite{clark2018think}& 2018.03& - & 7.78k & Natural Science& Multiple choice& Pre-college &     \\
  \href{https://huggingface.co/datasets/sciq}{SciQ}~\cite{SciQ} &  2017.07& - & 13.7k & Natural Science & Multiple choice  & Mixed &   \\ 

\href{https://huggingface.co/datasets/sciq}{ChemData}~\cite{zhang2024chemllm} &  2024.02& -& 727k & Chemistry & QA  & Mixed &   \\ 
    
 \bottomrule
\end{tabular}
\end{table*}

\subsection{Evaluation}\label{sec:text-LLM-eval}

\paratitle{Capabilities.}
The evaluation of LLMs is often structured around Bloom's taxonomy~\cite{krathwohl2002revision,forehand2010bloom},  which outlines six cognitive levels of educational learning objectives: Remember, Understand, Apply, Analyze, Evaluate and Create. More recently, SciEval~\cite{sun2023scieval} has suggested assessing scientific LLMs across four dimensions: basic knowledge, knowledge application, scientific calculation, and research ability. Each dimension aligns with one or more cognitive levels in Bloom's taxonomy.
While these frameworks offer cognitive and research perspectives, our focus is to gauge the depth of knowledge grasped by LLMs,  ideally comparable to human comprehension. Therefore, this survey 
categorizes the capabilities of Text-Sci-LLMs based on the complexity of scientific knowledge, differentiating between Pre-college, College, and Post-college levels. 

\begin{itemize}
\item \paratitle{Pre-college Level.} Scientific knowledge at the pre-college level primarily focuses on fundamental concepts, principles, and theoretical frameworks, serving as a foundation for further comprehensive study and research in scientific disciplines. This predominantly textual knowledge is readily accessible across various corpora, making it easily comprehensible for LLMs. We refer to the understanding of basic knowledge as the pre-college level of LLMs, which corresponds to the Remember and Understand levels in Bloom's taxonomy and the basic knowledge in SciEval. Some evaluation benchmarks such as MMLU~\cite{hendrycks2021measuring} and C-Eval~\cite{huang2023ceval} involve assessing pre-college level's knowledge.
\item \paratitle{College Level.} Scientific knowledge at the college level is typically more specialized, abstract, and necessitates logical reasoning and proof rather than mere memorization. It undergoes frequent updates aligned with scientific and industrial advancements, posing a challenge for LLMs to grasp and adapt to these continual changes.
The master of specialized knowledge is designated as the college level of LLMs, which typically aligns with the Apply and Analyze levels in Bloom's taxonomy, as well as the knowledge application and the scientific calculation in SciEval. Most evaluation benchmarks such as PubMedQA~\cite{jin2019pubmedqa} and SciQ~\cite{SciQ} focus on this level.
\item \paratitle{Post-college Level.} At this level, LLMs need to not only master up-to-date knowledge but also create innovative ideas. The ability to answer questions goes beyond multiple-choice, true/false, and short-answer formats. It also requires the capability to address generative questions, such as summarizing technological advancements and designing novel experimental schemes. This post-college level aligns with the Evaluate and Create levels in Bloom's taxonomy and the research ability in SciEval. A subset in the SciEval dataset~\cite{sun2023scieval} is one of the few benchmarks that can reflect such capabilities of LLMs.
\end{itemize}


\paratitle{Evaluation Metric.}
\label{sec:general-criteria}
To quantitatively measure the performance of Text-Sci-LLMs, one can utilize the standard metrics commonly employed in the field of machine learning. These include Accuracy, Precision, Recall, and F1-score, which are particularly well-suited for discriminative tasks such as multiple choice and judgment questions.
For generative tasks such as question-answering, it is evaluated by measuring the similarity of the LLM's output and the ground-truth. These similarity metrics include (but not limited to):  
\begin{itemize}
    \item \paratitle{BLEU} (Bilingual Evaluation Understudy)~\cite{papineni2002bleu} is a popular metric for evaluating machine translation. 
    BLEU gives a score between 0 and 1 (often expressed as a percentage), where higher scores indicate better translation quality. However, It does not account for the semantic meaning of words and can sometimes rate unnatural translations highly if they have a high n-gram overlap with the reference.
    \item \paratitle{ROUGE} (Recall-Oriented Understudy for Gisting Evaluation)~\cite{lin2004rouge} is primarily used to evaluate automatic text summarization and machine translation. It compares the overlap of n-grams, word sequences, and word pairs between the generated text and a set of reference texts. Higher ROUGE scores signify stronger alignment between the automatic summary and the reference summaries.
    \item \paratitle{BERT-Score}~\cite{zhang2019bertscore} leverages the contextual embeddings from BERT to evaluate the quality of text generation. It compares the similarity of BERT embeddings between the generated candidate and reference texts. BERT-Score addresses some limitations of BLEU and ROUGE, as it considers the context and semantic meaning of words, leading to a potentially more accurate assessment of text generation quality.
    \item \paratitle{LLM-Score} involves using an LLM with a carefully designed prompt to evaluate the quality of generated text. The LLM is prompted to assess the coherence, relevance, and accuracy of the output compared to the reference answer.
    
\end{itemize}

When the ground-truth is absent, one could sort to experts or strong LLMs like GPT-4 \cite{OpenAI-OpenAI-2023-GPT-4} for assessing LLMs. Human evaluation plays a pivotal role in assessing the quality and accuracy of model-generated results, as it closely aligns with real-world application scenarios and provides comprehensive and precise feedback. However, due to extensive human involvement, this process is time-consuming and labor-intensive. Therefore, there is a growing tendency that employ GPT-4 as a ``judge" to evaluate different models \cite{bubeck2023sparks}. Overall, these metrics are often used complementarily in various NLP tasks to provide a comprehensive evaluation of LLMs. 

\subsection{Summary}\label{sec:text-LLM-summary} 
This section focuses on Sci-LLMs trained with textual scientific corpora, particularly those acquiring extensive chemical and biological knowledge. We carefully review the existing Text-Sci-LLMs, their datasets, and evaluation methods. One can conclude that the development of Text-Sci-LLMs is built upon the advancements of general LLMs, with the key difference being the incorporation of more domain-specific corpora during training and a greater emphasis on evaluating the accuracy of scientific knowledge.  However, in contrast to the proliferation of LLMs focused on general domains, the number of Text-Sci-LLMs remains limited. Additionally, there are few benchmarks for evaluating LLMs at the post-college level which is a crucial capability for advanced scientific research.

	\clearpage
	
	\section{Molecular Large Language Models} \label{sec-molecule}
\begin{figure*}[h]
\centering
\resizebox{1.0\textwidth}{!}{
\begin{forest}
  for tree={
  grow=east,
  reversed=true,
  anchor=base west,
  parent anchor=east,
  child anchor=west,
  base=left,
  font=\small,
  rectangle,
  draw,
  rounded corners,align=left,
  minimum width=2.5em,
  inner xsep=4pt,
  inner ysep=1pt,
  },
  where level=1{text width=5em,fill=blue!10}{},
  where level=2{text width=5em,font=\footnotesize,fill=pink!30}{},
  where level=3{font=\footnotesize,yshift=0.26pt,fill=yellow!20}{},
  [Mol-LLMs, fill=green!10
        [Models\\(Sec.~\ref{sec:mol-LLM-overview}),text width=4em
            [Encoder-only, text width= 5.5em
              [\emph{BERT-based:} 
              SMILES-BERT~\cite{wang2019smiles} \, 
              MTL-BERT~\cite{zhang2022pushing}\,
              MolBERT~\cite{fabian2020molecular}\,\\
              rxnfp-BERT~\cite{schwaller2021mapping}\,      
              Mol-BERT~\cite{li2021mol}\,              
              MolFormer~\cite{ross2022large} \,             
              MolRoPE-BERT~\cite{liu2023molrope}
              ]              
              [\emph{RoBERTa-based:} 
              ChemBERTa~\cite{chithrananda2020chemberta}\,
              MFBERT~\cite{abdel2022large}\, \\
              SELFormer~\cite{yuksel2023selformer} \,
              Semi-RoBERTa~\cite{tran2023molecular}\,
              ]              
              [\emph{Integrating 2D or 3D graphs:}\             
              GROVER~\cite{rong2020self}\, 
              MAT~\cite{maziarka2020molecule}\,
              MG-BERT~\cite{zhang2021mg}\, 
              R-MAT~\cite{maziarka2024relative}\, \\      
              KPGT~\cite{li2022kpgt}\, 
              AGBT~\cite{chen2021algebraic} \,
              Molformer~\cite{wu2023molformer}\,
              Uni-Mol~\cite{zhou2023uni}\,      GTMGC~\cite{xugtmgc} \,     
              ]                                 
            ]            
            [Decoder-only, text width= 5.5em
              [\emph{GPT-based:} 
              MolGPT~\cite{bagal2021molgpt}\,
              SMILESGPT~\cite{adilov2021generative}\,
              cMolGPT~\cite{wang2023cmolgpt} \, Taiga~\cite{mazuz2023molecule}\,
              iupacGPT~\cite{cho2023iupacgpt}
              ]
            ]            
            [Encoder-decoder, text width=5.5em
              [\emph{Vanilla Transformer-based:} \\              
              Molecular Transformer~\cite{schwaller2019molecular}\,
              SMILES Transformer~\cite{honda2019smiles}
              \,
              SCROP~\cite{zheng2019predicting}\,\\
              Retrosynthesis Transformer~\cite{karpov2019transformer}\,              
              ChemReactNet~\cite{tetko2020state} \,
              X-MOL~\cite{xue2020x}\,\\
              Two-way Transformer~\cite{kim2021valid} \,
              RetroSynth-Diversity~\cite{toniato2023enhancing}\,
              GO-PRO~\cite{mann2021predicting}\,\\              Transmol~\cite{zhumagambetov2021transmol} \,            RetroTRAE~\cite{ucak2022retrosynthetic}\,                    
              Disconnection aware model~\cite{thakkar2023unbiasing}           
              ]  
              [\emph{BART-based:}              
              Chemformer~\cite{irwin2022chemformer}\,
              BARTSmiles~\cite{chilingaryan2022bartsmiles}\,
              MOLGEN~\cite{fang2023domain}
              ]
              [\emph{Transformer+Graph:}              
              GET~\cite{mao2021molecular}\,
              Graph2SMILES~\cite{tu2022permutation}
              ]                
              [\emph{Others:}              
              GCT~\cite{kim2021generative} 
              ]
            ]    
        ]
        [Datasets\\
        (Sec.~\ref{sec:mol-LLM-dataset}),text width=4em
            [Pre-training, text width=4em
              [ZINC~\cite{sterling2015zinc}\,
              PubChem~\cite{wang2012pubchem}\,
              USPTO~\cite{schneider2016s}\,
              PCQM4M~\cite{hu2021ogb,ying2021first}\,
              GEOM~\cite{axelrod2022geom}  \,\\
              MolTLU~\cite{beaini2023towards}\,
              ChEMBL~\cite{gaulton2012chembl}  \,
              DrugBank~\cite{wishart2018drugbank}\,
              GDB-17~\cite{ruddigkeit2012enumeration}\,
              ExCAPE-DB~\cite{sun2017excape}\,
              ]
            ]
            [Benchmarks, text width=4em
             [MoleculeNet~\cite{wu2018moleculenet}\,
             MARCEL~\cite{zhu2023learning}\,
             GuacaMol~\cite{brown2019guacamol} \, \\               
             MOSES~\cite{polykovskiy2020molecular}\,              
             ADMETlab 2.0 ~\cite{xiong2021admetlab}\,             Molecule3D~\cite{xu2021molecule3d}\,
             ] 
            ] 
        ]
        [Evaluation\\
        (Sec.~\ref{sec:mol-LLM-eval}),text width=4em   
        [{Property prediction},text width=7em    
        [LogP \, Solubility\, Toxicity\, ADMET \, Lipophilicity \, Stability \, etc.]
        ]
        [{Interaction prediction},text width=7em    
        [Drug-drug interactions]
        ]
        [{Reaction prediction},text width=7em   
        [Forward reaction prediction and Retrosynthetic analysis]
        ]
        [{Molecule generation},text width=7em    
       [Template-based design and De novo design]
        ]         
        [Evaluation metric, text width=7em
          [\textit{Discriminative tasks}: Accuracy \, Recall \, F1-score\, etc.
          ]
          [\textit{Generative tasks}: Validity \, Unique \, Novelty \, Frag \, Scaff \, IntDiv$_p$ \, FCD\, etc.
          ]     
        ]
        ]
        [Summary\\
        (Sec.~\ref{sec:mol-LLM-summary}),text width=4em
        ]
    ]
\end{forest}
}
\caption{Chapter overview of Mol-LLMs.}
\label{fig:mol-LLM-overview}
\end{figure*}
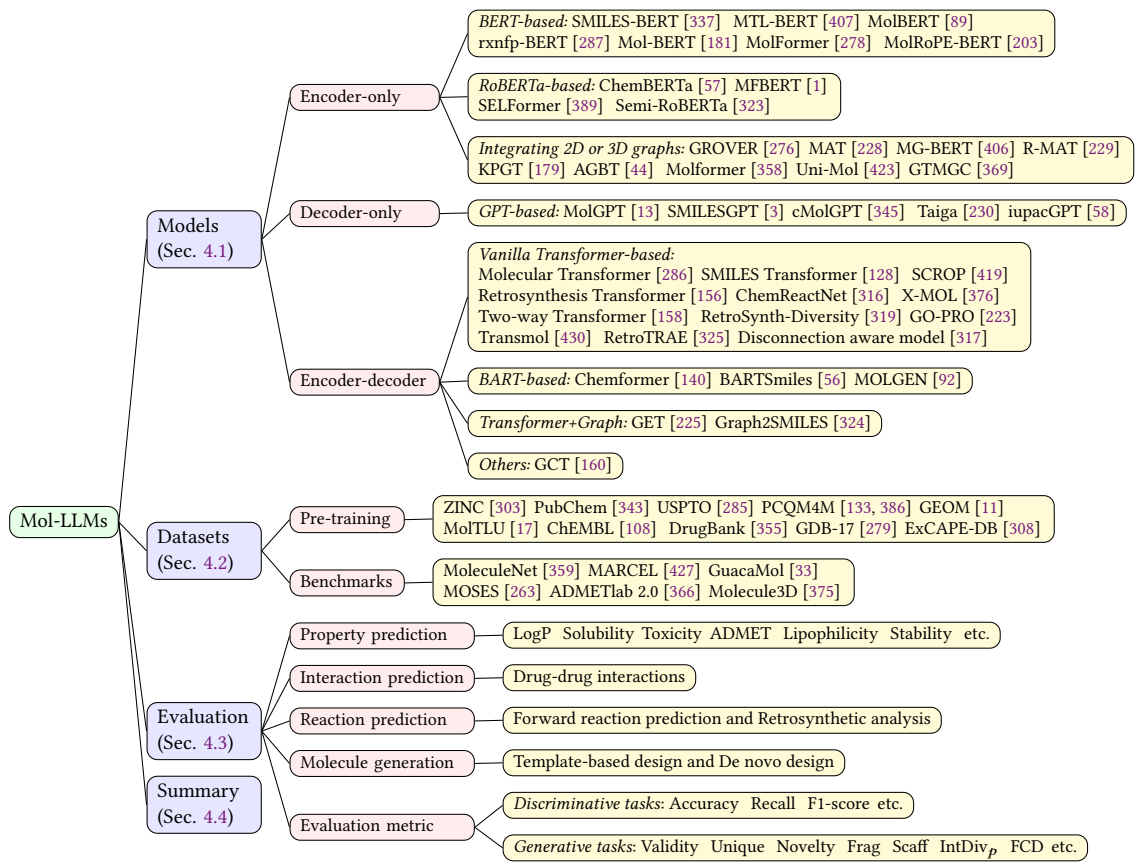

Large language models have shown great potential in accelerating chemical molecular discovery. In this section, we provide a review of LLMs trained in molecular language (Mol-LLMs), including insights into their model architectures, capabilities, utilized datasets, and evaluation criteria. The overview of this section is shown in Figure~\ref{fig:mol-LLM-overview}.

\subsection{Models}
\label{sec:mol-LLM-overview}

As Transformer \cite{Vaswani-NIPS-2017-Attention} can effectively model sequential data, numerous Transformer-based Mol-LLMs have been proposed for predicting molecular properties and structures. Janakarajan et al.~\cite{janakarajan2023language}, Bran and Schwaller~\cite{bran2023transformers} had extensively reviewed the application of Transformer technology in molecular discovery, underscoring its significant impact on advancing the field. Here we categorize Mol-LLMs into three distinct types based on their specific architectures: encoder-only, decoder-only, and encoder-decoder models, as shown in Table \ref{tab:Mol_LLMs}.

\subsubsection{Encoder-only Models} \ \\

Encoder-only models focus on understanding and interpreting the input molecules, making them well-suited for tasks that require a deep comprehension of molecular structures and properties. A typical example is SMILES-BERT~\cite{wang2019smiles} that leverages the BERT architecture to interpret SMILES representations of molecules. It undergoes pre-training on an extensive dataset of unlabeled data through masked SMILES recovery tasks, enabling it to learn nuanced patterns and relationships within molecular structures. Nevertheless, the fine-tuning phase for individual downstream tasks is inefficient. MTL-BERT~\cite{zhang2022pushing} mitigates this limitation by employing a multitask learning framework, which substantially increases the accuracy of predictions for multiple molecular properties. MolBERT~\cite{fabian2020molecular} revolutionizes molecular representation learning by integrating SMILES Equivalence  and physical-chemical property prediction tasks, significantly enhancing performance in virtual screening and QSAR benchmarks. The rxnfp-BERT~\cite{schwaller2021mapping} model draws on the principles of BERT to process and learn from chemical reactions. It is designed to generate reaction fingerprints, which are representations of chemical reactions used for various predictive tasks. However, these models predominantly concentrate on contextual information in molecular sequences while giving less attention to molecular substructures, such as functional groups. Mol-BERT~\cite{li2021mol} extracts atomic-level and substructural features centered on the current atom. MolFormer~\cite{ross2022large} further enhances this approach by integrating relative position encoding and rotational position embedding, thereby facilitating the comprehension of spatial relationships between atoms in a molecule. MolRoPE-BERT~\cite{liu2023molrope} augments this method by employing Rotary Position Embedding (RoPE) to efficiently encode positional information in SMILES, aiming to provide a more robust understanding of molecules.

\begin{table*}[htbp]
\centering
\caption{Summary of Mol-LLMs}
\label{tab:Mol_LLMs}
\footnotesize
\renewcommand\tabcolsep{2.5pt}
\begin{tabular}{llcccccc}
\toprule
  & Model & Time  &  \#Parameters & Base model &  Pre-training Dataset &  Capability  & Open-source \\
  
\midrule
\multirow{23}{*}{Encoder-only} 
   & \href{https://github.com/uta-smile/SMILES-BERT}{SMILES-BERT}~\cite{wang2019smiles} & 2019.09 & - & BERT & ZINC & Prop. pred. & \checkmark  \\
   & \href{https://github.com/ardigen/MAT}{MAT}~\cite{maziarka2020molecule} & 2020.02 & - & Trans. enc. & ZINC-15 & Prop. pred. & \checkmark \\
   & \href{https://huggingface.co/seyonec/ChemBERTa-zinc-base-v1}{ChemBERTa}~\cite{chithrananda2020chemberta} & 2020.10 & - & RoBERTa & PubChem & Prop. pred. & \checkmark \\
   & \href{https://github.com/tencent-ailab/grover}{GROVER}~\cite{rong2020self} & 2020.10 & 48/100M & Trans. enc.+GNN & ZINC-15, ChEMBL & Prop. pred. & \checkmark  \\
   & \href{https://github.com/BenevolentAI/MolBERT}{MolBERT}~\cite{fabian2020molecular} & 2020.11 & 85M & BERT & ChEMBL & Prop. pred. & \checkmark  \\
   & \href{https://rxn4chemistry.github.io/rxn_yields/}{rxnfp-BERT}~\cite{schwaller2021mapping} & 2021.01 & - & BERT& Pistachio, USPTO & React. pred. & \checkmark\\
   & \href{https://github.com/zhang-xuan1314/Molecular-graph-BERT}{MG-BERT}~\cite{zhang2021mg} & 2021.05 & -  & BERT+GNN   & ChEMBL & Prop. pred. & \checkmark \\
   & \href{https://github.com/ChenDdon/AGBTcode}{AGBT}~\cite{chen2021algebraic} & 2021.06 & - & BERT & ChEMBL & Prop. pred. & \checkmark \\
   & \href{https://github.com/cxfjiang/MolBERT}{Mol-BERT}~\cite{li2021mol} & 2021.09 & - & BERT & ZINC-15, ChEMBL 27 & Prop. pred. & \checkmark \\
   & \href{https://github.com/gmum/huggingmolecules}{R-MAT}~\cite{maziarka2024relative} & 2021.10 & - & Trans. enc. & ZINC-15, ChEMBL & Prop. pred. & \checkmark \\
   & \href{https://github.com/lihan97/KPGT}{KPGT}~\cite{li2022kpgt} & 2022.08 & 100M & Trans. enc. & ChEMBL 29 & Prop. pred. & \checkmark \\
   & \href{https://huggingface.co/seyonec/ChemBERTa-zinc-base-v1}{ChemBERTa-2}~\cite{ahmad2022chemberta} & 2022.09 & 5/46M & RoBERTa & PubChem & Prop. pred. & \checkmark \\ 
   & \href{https://github.com/GouldGroup/MFBERT}{MFBERT}~\cite{abdel2022large} & 2022.10 & 88M & RoBERTa & \makecell{GDB-13, ZINC-15, \\ PubChem, ChEMBL, USPTO} & Prop. pred. & \checkmark \\
   & \href{https://github.com/IBM/molformer}{MolFormer}~\cite{ross2022large} & 2022.12 & - & Trans. enc. & PubChem, ZINC  & Prop. pred. & \checkmark \\
   & \href{https://github.com/zhang-xuan1314/MTL-BERT}{MTL-BERT}~\cite{zhang2022pushing} & 2022.12 & - & BERT & ChEMBL & Prop. pred. & \checkmark \\
   & MolRoPE-BERT~\cite{liu2023molrope} & 2023.01 & - & BERT & ZINC-15, ChEMBL 27 & Prop. pred. & $\times$ \\  
   & \href{https://github.com/smiles724/Molformer}{Molformer}~\cite{wu2023molformer} & 2023.01 & - & Trans. enc. & ZINC, PubChem & Prop. pred. & \checkmark \\
   & \href{https://github.com/dptech-corp/Uni-Mol}{Uni-Mol}~\cite{zhou2023uni} & 2023.02 & 47.61M & Trans. enc. & \makecell{ZINC, ChemBL \\ PDBbind} & \makecell{Prop. pred. \\ Conform. gen.} & \checkmark \\
   & \href{https://github.com/HUBioDataLab/SELFormer}{SELFormer}~\cite{yuksel2023selformer} & 2023.05 & 58/87M & RoBERTa & ChEMBL & Prop. pred. & \checkmark \\   
   & Semi-RoBERTa~\cite{tran2023molecular} & 2023.07 & - & RoBERTa & PubChem & Prop. pred. & $\times$ \\ 
   & \href{https://github.com/Rich-XGK/GTMGC}{GTMGC}~\cite{xugtmgc} & 2023.10 &5/12/26M &Trans. enc.+GNN & Molecule3D, QM9& Conform. gen. & \checkmark \\
\midrule 
\multirow{6}{*}{Decoder-only} 
   & \href{https://github.com/devalab/molgpt}{MolGPT}~\cite{bagal2021molgpt}   & 2021.05  & 6M & GPT & MOSES, GuacaMol & Mol. gen. & \checkmark \\  
   & \href{https://github.com/sanjaradylov/smiles-gpt}{SMILES GPT}~\cite{adilov2021generative} & 2021.09 & 13.4M & GPT & PubChem &  Mol. gen., Prop. pred. & \checkmark \\
   & cTransformer~\cite{wang2022pre} & 2022.10 & - & Trans. dec. & MOSES &  Mol. gen. & $\times$ \\
   & \href{https://github.com/AspirinCode/iupacGPT}{iupacGPT}~\cite{cho2023iupacgpt} & 2023.05 & 1500M & GPT & PubChem & Mol. gen., Prop. pred. & \checkmark \\ 
   & \href{https://github.com/VV123/cMolGPT}{cMolGPT}~\cite{wang2023cmolgpt} & 2023.05 & - & GPT & MOSES & Mol. gen. & \checkmark  \\ 
   & \href{https://github.com/eyalmazuz/MolGen}{Taiga}~\cite{mazuz2023molecule} & 2023.05 & - & GPT & \makecell{MOSES, ZINC, GDB 13} & Mol. gen., Prop. pred. & \checkmark \\   
\midrule
\multirow{26}{*}{Encoder-Decoder} 
   & \href{https://github.com/pschwllr/MolecularTransformer}{\makecell[l]{Molecular- \\ Transformer~\cite{schwaller2019molecular}}} & 2019.08 & 12M & Transformer & USPTO, Pistachio & React. pred. & \checkmark \\
   & \href{https://github.com/bigchem/retrosynthesis}{\makecell[l]{Retrosynthesis- \\ Transformer~\cite{karpov2019transformer}}} & 2019.09 & - & Transformer & USPTO & React. pred. & \checkmark \\
   & \href{https://github.com/DSPsleeporg/smiles-transformer}{\makecell[l]{SMILES- \\Transformer~\cite{honda2019smiles}}} & 2019.11 & - & Transformer & ChEMBL 24 & Prop. pred. & \checkmark \\
   & \href{https://github.com/Jh-SYSU/SCROP}{SCROP}~\cite{zheng2019predicting} & 2019.12 & - & Transformer & USPTO & React. pred. & \checkmark \\
   & \href{https://github.com/bigchem/synthesis}{ChemReactNet}~\cite{tetko2020state} & 2020.11 & - & Transformer & USPTO & React. pred. & \checkmark \\ 
   & \href{https://github.com/bm2-lab/X-MOL}{X-MOL}~\cite{xue2020x} & 2020.12 & - & Transformer & ZINC-15 & \makecell{Prop. pred., React. pred.\\Mol. optim., Mol. gen.} & \checkmark \\
   & \href{https://github.com/ejklike/tied-twoway-transformer/}{\makecell[l]{Two-way- \\ Transformer~\cite{kim2021valid}}} & 2021.01 & - & Transformer & USPTO & React. pred. & \checkmark \\ 
   & GO-PRO~\cite{mann2021predicting} & 2021.03 & 5M & Transformer & USPTO & React. pred. & $\times$ \\ 
   & \href{https://gitlab.com/cheml.io/public/transmol}{Transmol}~\cite{zhumagambetov2021transmol} & 2021.07 & - & Transformer & MOSES &  Mol. gen. & \checkmark \\
   & \href{https://github.com/knu-lcbc/RetroTRAE}{RetroTRAE}~\cite{ucak2022retrosynthetic} & 2021.08 & - & Transformer & USPTO, PubChem, ChEMBL & React. pred. & \checkmark \\
   & \href{https://github.com/papercodekl/MolecularGET}{GET}~\cite{mao2021molecular} & 2021.10 & -  & \makecell{Transformer\\+GNN} & USPTO &  React. pred. & \checkmark \\
   & \href{https://github.com/coleygroup/Graph2SMILES}{Graph2SMILES}~\cite{tu2022permutation} & 2021.10 & - & Transformer & \makecell{USPTO} & React. pred. & \checkmark \\
   & \href{https://github.com/Hyunseung-Kim/molGCT}{GCT}~\cite{kim2021generative} & 2021.12 & - & Transformer & MOSES &  Mol. gen. & \checkmark \\
   & \href{https://github.com/MolecularAI/Chemformer}{Chemformer}~\cite{irwin2022chemformer}   & 2022.01 & 45/230M   & BART & ZINC-15 & Prop. pred., React. pred. & \checkmark \\
   & \href{https://github.com/YerevaNN/BARTSmiles/}{BARTSmiles}~\cite{chilingaryan2022bartsmiles} & 2022.11 & - & BART & ZINC 20 & Prop. pred. & \checkmark \\
   & \href{https://github.com/rxn4chemistry/rxn_cluster_token_prompt}{\makecell[l]{RetroSynth- \\Diversity~\cite{toniato2023enhancing}}} & 2023.02 & 12M & Transformer &  USPTO, Pistachio & React. pred. & \checkmark \\
   & \href{https://github.com/rxn4chemistry/disconnection_aware_retrosynthesis}{\makecell[l]{Disconnection aware- \\ model~\cite{thakkar2023unbiasing}}} & 2023.07 & - & Transformer & \makecell{USPTO\\ Pistachio} & React. pred. & \checkmark \\
   & \href{https://github.com/zjunlp/MolGen}{MOLGEN}~\cite{fang2023domain} & 2023.10 & - & BART & ZINC-15 & Mol. gen. & \checkmark \\
   
\bottomrule
\end{tabular}
\end{table*}

RoBERTa~\cite{liu2019roberta}, building upon the BERT's framework,  introduces dynamic token masking for varied learning and eliminates the next sentence prediction task to better learn long-range dependencies. These enhancements address BERT's limitations and elevate RoBERTa's performance across diverse NLP tasks, making it suitable for specialized applications like molecular modeling. A variety of models have been built upon the RoBERTa framework, including chemBERTa~\cite{chithrananda2020chemberta}, chemBERTa-2~\cite{ahmad2022chemberta}, MFBERT~\cite{abdel2022large}, SELFormer~\cite{yuksel2023selformer} and semi-RoBERTa~\cite{tran2023molecular}. MFBERT and semi-RoBERTa both use SMILES sequences as their input, crafting simple molecular representations. MFBERT~\cite{abdel2022large} employs distributed training and a data-driven approach, focusing on generating targeted molecular fingerprints using SMILES sequences. It uses a sentence-piece tokenization model, ensuring robust and accurate molecular representations. Meanwhile, semi-RoBERTa~\cite{tran2023molecular}, uses SMILES inputs to capitalize on 3D compound structures.

Although SMILES is commonly used for molecular representation, it poses challenges in terms of validity and robustness. To address these issues, SELFIES can be an alternative to SMILES. 
As a molecular notation system, SELFIES~\cite{krenn2020self} represents organic molecules in an accessible string format with a recursive syntax, enabling self-referencing to enhance both the accuracy and computational efficiency of representations. 
ChemBERTa~\cite{chithrananda2020chemberta} adopts a broader approach, processing both SMILES and SELFIES inputs. This model delves into the effects of size, tokenizer types, and string representations, employing Byte-Pair Encoding (BPE) to manage these varied inputs. ChemBERTa-2~\cite{ahmad2022chemberta} further integrates multi-task regression and larger datasets, thereby augmenting its proficiency in molecular property analysis. 

The above BERT-based models integrate a vast amount of unlabeled molecular data through a self-supervised learning strategy, but they are based on the sequential format of molecule SMILES or SELFIES, overlooking crucial graph structures and three-dimensional (3D) chemical information. 
The oversight of graph structures can be effectively addressed by integrating graphs with Transformers, such as GROVER~\cite{rong2020self}, MAT~\cite{maziarka2020molecule}, R-MAT~\cite{maziarka2024relative}, MG-BERT~\cite{zhang2021mg}, KPGT~\cite{li2022kpgt} and GTMGC~\cite{xugtmgc}. For example, MG-BERT~\cite{zhang2021mg} combines GNN's local message passing with BERT's architecture, enhancing molecular feature extraction and utilizing a unique pretraining strategy involving atom masking.
Certainly, GNNs alone can deal with molecule graph structures, and representative works include {KCL}~\cite{fang2022molecular}, {KANO}~\cite{fang2023knowledge}, GS-Meta~\cite{zhuang2023graph}, and iMoLD~\cite{zhuang2023learning}. 
Moreover, several approaches have been employed to model 3D molecules, including AGBT~\cite{chen2021algebraic}, Molformer~\cite{wu2023molformer}, and Uni-mol~\cite{zhou2023uni}, and DeepSorption~\cite{cui2023direct}, showcasing the evolution and diversification of computational methods for molecular modeling. For example, DeepSorption leverages a SE(3)-equivalent Transformer to effectively model the 3D structures of molecular crystals, leading to accurate prediction of gas adsorption.
Graph and 3D structure-based molecular representation learning is out of the scope of this survey and we refer interested readers to the surveys~\cite{guo2022graph,xia2023systematic}.

\subsubsection{Decoder-only Models} \ \\
In recent years, decoder-only architectures like GPT have revolutionized molecular generation in cheminformatics. Mol-LLMs based on the GPT architecture, including MolGPT~\cite{bagal2021molgpt}, SMILESGPT~\cite{adilov2021generative}, iupaccGPT~\cite{cho2023iupacgpt}, cMolGPT~\cite{wang2023cmolgpt}, and Taiga~\cite{wang2022pre}, primarily use SMILES strings as input, effectively addressing the challenge of navigating the vast chemical space. These models are pivotal in drug discovery and material science, enabling the synthesis of molecules with specific properties. 

MolGPT~\cite{bagal2021molgpt}, a pioneer in utilizing GPT for molecular generation, uses conditional training for property optimization. It stands out for its efficiency and effectiveness in molecular modeling and drug discovery, showing strong control over multiple properties for accurate generation. 
Moreover, both SMILESGPT and iupacGPT are based on the GPT-2 architecture. SMILESGPT~\cite{adilov2021generative} is a causative Transformer tailored for drug discovery, training on SMILES notation with BPE, and employing special embedding layers for effective molecular representation learning. iupacGPT~\cite{cho2023iupacgpt} diverges by using the more readable IUPAC nomenclature, excelling in learning hierarchical and inner-sequence relationships between atoms and chemical groups. 
The cTransformer~\cite{wang2022pre} and cMolGPT~\cite{wang2023cmolgpt} models are specifically developed for the purpose of generating molecules based on given conditions. cTransformer~\cite{wang2022pre} synergizes the Transformer architecture with conditional embeddings, allowing for target-specific molecule generation and fine-tuning with target-specific data. cMolGPT~\cite{wang2023cmolgpt} represents an advancement in this domain by generating target-specific compounds by utilizing target embeddings as keys and values within its multi-head attention, adeptly producing molecules active against designated targets. Taiga~\cite{mazuz2023molecule} exemplifies the combination of GPT and reinforcement learning (RL) methodologies. It employs a two-stage approach, initially treating the problem as a language modeling task with the prediction of the next token in SMILES strings. Afterwards, it utilizes RL to optimize chemical features, notably emphasizing qualities such as a quantitative estimate of QED (Quantitative Estimate of Drug-likeness). This demonstrates the smooth integration of GPT's language processing abilities with the accuracy of RL approaches.

\subsubsection{Encoder-Decoder Models}  \ \\
In the encoder-decoder architecture, encoders convert raw molecules into latent vectors, and decoders reconstruct these latent vectors into functional chemical structures. In the majority of Transformer-based encoder-decoder models, SMILES or SELFIES serve as inputs for the encoder, and the outputs vary across tasks. For example, in the context of chemical reaction prediction, the decoder generates the anticipated outcomes for reactants. The following paragraphs describe Mol-LLMs based on the encoder-decoder architecture in chronological order.

In the field of chemical reaction prediction, Molecular Transformer~\cite{schwaller2019molecular} is a pioneering Transformer-based model for reaction prediction, which enables the effective handling of complex, long-range sequence interactions. Retrosynthesis Transformer~\cite{karpov2019transformer} stands out for snapshot learning techniques to improve training efficiency. It utilizes a position encoding matrix to address the issue of distance between string elements, enhancing its ability to process and interpret complex chemical information. SCROP~\cite{zheng2019predicting} excels in inferring a set of original candidate reactants and incorporates a syntax corrector with position encoding to address the challenges of string element distances. 
ChemReactNet~\cite{tetko2020state} excels in chemical reaction prediction by employing advanced data augmentation, notably using SMILES augmentation and beam search algorithms, to significantly reduce neural network data memorization and enhance prediction of new sequences. 
Two-Way Transformer~\cite{kim2021valid} in retrosynthesis features a novel architecture where modules of two Transformers are tied together, sharing the encoder and most decoder components. It employs latent modeling with a multinomial latent variable, significantly diversifying the range of generated reactant candidates. 
GO-PRO~\cite{mann2021predicting} combines Transformer with a context-free grammar (CFG)-based representation of molecules. It encodes chemical reactions using a grammar-based system, utilizing either a greedy or beam-search decoding strategy for output, resulting in efficient, structurally informed predictions with reduced model complexity.

BART~\cite{Lewis-ACL-2020-BART}, a Transformer-based architecture renowned for its bidirectional encoder and autoregressive decoder, has been effectively applied in molecular design. Chemformer~\cite{irwin2022chemformer}, BARTSmiles~\cite{chilingaryan2022bartsmiles} and MOLGEN~\cite{fang2023domain} are representative models based on the BART architecture. 
Chemformer~\cite{irwin2022chemformer}, a BART language model fine-tuned for sequence-to-sequence and discriminative tasks in cheminformatics, showcases the efficacy of multi-task learning. It employs three innovative self-supervised pre-training techniques, applied to a substantial dataset of unlabeled SMILES strings. BARTSmiles~\cite{chilingaryan2022bartsmiles}, a robust generative masked language model for molecules, leverages extensive self-supervised pre-training on over 1.7 billion molecules. It stands out for its innovative training scale and denoising objectives, significantly boosting performance across multiple molecular tasks. MOLGEN~\cite{fang2023domain} combines domain-agnostic pre-training with a self-feedback paradigm, emphasizing the generation of chemically valid molecules and aligning generative probabilities with real-world chemical preferences to address challenges such as molecular hallucinations. 

Additionally, there are several models that use specialized inputs or combine Transformers with distinctive architectures. RetroTRAE~\cite{ucak2022retrosynthetic} not only excels in fragment-based tokenization but also learns changes in atomic environments during chemical reactions. GCT~\cite{kim2021generative} integrates a Transformer-based language model with a conditional variational autoencoder, augmented by a conditional Gaussian latent space. RetroSynth-Diversity~\cite{toniato2023enhancing} and Disconnection Aware model~\cite{thakkar2023unbiasing} both employ prompts to guide their retrosynthesis predictions, enhancing diversity and accuracy in different ways. RetroSynth-Diversity~\cite{toniato2023enhancing} leverages cluster token prompts to diversify retrosynthesis predictions, incorporating classification tokens into the molecular language representation. This approach guides the model towards a wider array of disconnection strategies, significantly broadening the exploration of chemical space. The Disconnection Aware model~\cite{thakkar2023unbiasing} enhances retrosynthesis pathway predictions by accurately identifying disconnection sites in molecules, using atom-tagged prompts integrated into product SMILES. It significantly boosts reaction class diversity and requires specialized training with inputs containing tagged disconnection sites.

However, these molecular representation learning methods often overlook the properties of substructures, neglecting natural atomic connections and molecular topological information. GET~\cite{mao2021molecular} and Graph2SMILES~\cite{tu2022permutation} are graph-based approaches designed to address these gaps. GET~\cite{mao2021molecular} integrates graph-level and sequence-level representations, combining a novel GNN with the Transformer model to enhance retrosynthesis prediction and generate more accurate SMILES outputs. 
Graph2SMILES~\cite{tu2022permutation} merges Transformers with molecular graphs, which incorporates a global attention encoder for long-distance and intermolecular interactions.

\begin{table*}[htbp]
\centering
\caption{Summary of the datasets for Mol-LLMs. }
\label{tab:mol-LLM-dataset}
\footnotesize
\renewcommand\tabcolsep{2.5pt}
\begin{tabular}{llcccc}
\toprule
   & Dataset & Last updated & Subset/Version &  Scale &  Keywords/Tasks \\
  \midrule
\multirow{29}{*}{Pre-training} 
   & \multirow{3}{*}{ZINC} & 2015.10
   & \href{https://ZINC15.docking.org/}{ZINC-15}~\cite{sterling2015zinc}     & 120M & \multirow{3}{*}{Ligand discovery} \\
   & & 2020.12 & \href{https://zinc20.docking.org/}{ZINC-20}~\cite{irwin2020zinc20} & 140M &  \\ 
   & & 2012.07& \href{https://figshare.com/articles/dataset/ZINC_250K_data_sets/17122427/1}{ZINC-250k}\cite{irwin2012zinc}  & 250k & \\  
   \cmidrule{2-6}
   & \multirow{11}{*}{\href{https://pubchem.ncbi.nlm.nih.gov/}{PubChem}~\cite{kim2023pubchem}} & \multirow{11}{*}{2023.01} & Compound & 111.9M & Unique chemical structures \\
   & & & Substance & 296.9M & Chemical entities \\
   & & & BioAssay & 1.5M & Biological experiments \\
   & & & Bioactivity & 296.8M & Biological activity data points \\
   & & & Protein & 185.1K & Tested and identified proteints \\
   & & & Gene & 104K & Tested and identified genes \\
   & & & Pathway & 239K & Interactions between chemicals, genes, and proteins \\
   & & & Cell Line & 2K & Tested cell lines\\
   & & & Taxonomy & 112.6K & Organisms of proteins/genes \\
   & & & Patent & 42.4M & Patents with links in PubChem \\
   & & & Data Sources & 872 & Organizations contributing data to PubChem \\
   \cmidrule{2-6}
   & \multirow{3}{*}{\href{https://developer.uspto.gov/data}{USPTO}~\cite{lowe2012extraction}} & \multirow{3}{*}{2012.10} & \href{https://github.com/wengong-jin/nips17-rexgen}{USPTO MIT} & 480k & \multirow{3}{*}{US Patents, unique reactions} \\
   & & & \href{https://github.com/connorcoley/ochem_predict_nn}{USPTO-15K} & 15k & \\
   & & & \href{https://github.com/dan2097/patent-reaction-extraction}{USPTO-full} & 950k & \\
   \cmidrule{2-6}
   & \multirow{2}{*}{PCQM4M} & 2021.10 & \href{https://ogb.stanford.edu/docs/lsc/pcqm4mv2/}{PCQM4Mv2}~\cite{hu2021ogb}  & 3.7M & Quantum property, molecular graphs pred. \\
   & & 2021.06  & \href{https://ogb.stanford.edu/kddcup2021/pcqm4m/}{PCQM4M-LSC}~\cite{ying2021first} & 3.8M & Large-Scale \\
   \cmidrule{2-6}
   & \multirow{3}{*}{\href{https://www.aicures.mit.edu/}{GEOM}~\cite{axelrod2022geom}} & \multirow{3}{*}{2022.04} & QM9 & 134K & Conformers \\ 
   & & & AICures drug dataset & 317.9K & Organic molecules, high-quality conformers \\
   & & & part of MoleculeNet & 16.9K & Biophysics, physiology, physical chemistry \\
   \cmidrule{2-6}
   & \multirow{3}{*}{\href{https://zenodo.org/record/8372621}{MolTLU}~\cite{beaini2023towards}} & \multirow{3}{*}{2023.10} & ToyMix & 154K & Quantum ML, drug discovery, GNN expressivity  \\
   & & & LargeMix & 5.4M & Biological properties \\
   & & & UltraLarge & 83M & Quantum properties \\
   \cmidrule{2-6}
   & \href{https://www.ebi.ac.uk/chembl/}{ChEMBL}~\cite{zdrazil2023chembl}   & 2023.05 & - & 22.7M   &  Drug-like bioactive compounds   \\
   & \href{https://go.drugbank.com/}{DrugBank 5.0}~\cite{wishart2018drugbank}   & 2017.11 & - & 500K   & Drug's mechanisms, interactions, targets  \\ 
   & \href{https://gdb.unibe.ch/downloads/}{GDB-17}~\cite{ruddigkeit2012enumeration} & 2012.10 & - & 166.4B & Organic small molecules \\
   & \href{https://pubchem.ncbi.nlm.nih.gov/}{ExCAPE-DB}~\cite{sun2017excape}  & 2017.03  & - & 70M & Bioactivity, Chemogenomics \\  
\midrule    
\multirow{27}{*}{Benchmarks} 
   & \multirow{18}{*}{\href{https://github.com/deepchem/deepchem}{MoleculeNet}~\cite{wu2018moleculenet}} & \multirow{18}{*}{2017.10} & QM7 & 7K & \multirow{4}{*}{Quantum mechanics prediction} \\
   & & & QM8 & 22K & \\
   & & & QM9 & 134K & \\
   \cmidrule{4-6}
   & & & ESOL & 1K & \multirow{3}{*}{Physical chemistry prediction} \\
   & & & FreeSoly & 643 & \\
   & & & Lipophilicity & 4K & \\
   \cmidrule{4-6}
   & & & PCBA & 440K & \multirow{5}{*}{Biophysics prediction} \\
   & & & MUV & 93K & \\
   & & & HIV & 42K & \\
   & & & PDBbind & 12K & \\
   & & & BACE & 1.5K & \\
   \cmidrule{4-6}
   & & & BBBP & 2K & \multirow{5}{*}{Physiology prediction}\\
   & & & Tox21 & 8K & \\
   & & & ToxCast & 8.6K & \\
   & & & SIDER & 1.4K & \\
   & & & ClinTox & 1.4K & \\
   \cmidrule{2-6}
   & \multirow{4}{*}{\href{https://github.com/SXKDZ/MARCEL}{MARCEL}~\cite{zhu2023learning}} & \multirow{4}{*}{2023.09} & Drugs-75K & 75k & Highly flexible molecules \\
   & & & Kraken & 1.5k & Organophosphorus ligand sterics \\
   & & & EE & 872 & Chiral catalysts and enantiomers \\
   & & & BDE & 5.9k & Organometallic catalyst conformations \\
    \cmidrule{2-6}
   & \href{https://benevolent.ai/guacamol}{GuacaMol}~\cite{brown2019guacamol} & 2019.03 & - & - & Molecule generation \\
   & \href{https://github.com/molecularsets/moses}{MOSES}~\cite{polykovskiy2020molecular}   & 2020.12  & ZINC Clean Leads & 1.9M   & Molecule generation  \\
   & \href{https://admet.scbdd.com/}{ADMETlab 2.0} ~\cite{xiong2021admetlab} & 2021.04 & -   & 250K   & Property prediction  \\
   & \href{https://github.com/mims-harvard/SPECTRA}{SPECTRA} ~\cite{ektefaie2024evaluating} & 2024.02 & - & 637k & Generalization ability evaluation \\
   & \href{https://github.com/divelab/MoleculeX}{Molecule3D}~\cite{xu2021molecule3d} & 2021.09 & GEOM, QM9 & 3.9M & Ground-state 3D Molecular Geometry Prediction\\
   \bottomrule
\end{tabular}
\end{table*}

\subsection{Datasets}\label{sec:mol-LLM-dataset}
Molecule datasets are generally categorized into two types based on the availability of annotations: unlabeled data and labeled data. Each type plays a different role in the model training and evaluation process. 

\paratitle{Pre-training Datasets.} 
Unlabeled data is often used for self-supervised learning, where the model learns to understand and represent molecular structures without explicit guidance on specific tasks. Here is a collection of pre-training datasets for Mol-LLMs, as summarized in Table \ref{tab:mol-LLM-dataset}.
\begin{itemize}
    \item \paratitle{ZINC}~\cite{irwin2012zinc,irwin2020zinc20,sterling2015zinc}. ZINC serves as a vital repository for researchers, especially in molecular dynamics simulations and virtual screening, offering an extensive array of compounds, including those ready for immediate delivery. ZINC 250K is a curated subset of approximately 250,000 molecules from ZINC, known for its pharmaceutical relevance and moderate size. ZINC-15 is an advanced iteration, offering billions of compounds and enhanced search capabilities for a growing chemical space. ZINC 20 further expands this with an ultralarge-scale chemical database, incorporating novel molecules and sophisticated search technologies to address the challenges of massive, dynamic chemical libraries.
    \item \paratitle{PubChem}~\cite{wang2012pubchem}. 
    PubChem is a comprehensive database that offers free access to information on chemical substances and their biological activities.  It contains over 750 million records covering a wide range of data, including chemical structures, identifiers, bioactivity outcomes, genes, proteins, and patents.  
    PubChem is a crucial public repository for biological activity data of small molecules and RNAi reagents, aimed at providing free and easy access to a vast array of data, including over 130 million bioactivity outcomes, 5000 protein targets, and 30,000 gene targets. This extensive database, housing 500,000 assay protocol descriptions, 1.6 million small molecules, 120 chemical probes, and 60,000 RNAi reagents, is integral to molecular modeling and drug discovery. 
    \item \paratitle{USPTO}\cite{schneider2016s}. The USPTO dataset is a significant compilation of chemical reactions extracted from US patents spanning 1976 to 2015.  The dataset includes about 1.3 million unique reactions, covering a broad range of more than 200 different reaction types. This dataset is particularly useful for developing and evaluating new methods in reaction role assignment and atom-to-atom mapping, offering a diverse and extensive collection of reactions for cheminformatics research.  
    \item \paratitle{PCQM4M}~\cite{hu2021ogb,ying2021first}. The PCQM4Mv2 and PCQM4M-LSC datasets are designed for graph-level prediction tasks in quantum chemistry, primarily focusing on predicting the Density Functional Theory (DFT) calculated HOMO-LUMO energy gaps of molecules from their 2D molecular graphs. These datasets provide a robust platform for developing and benchmarking advanced machine learning models in predicting molecular properties. The primary difference between PCQM4Mv2 and PCQM4M-LSC is that PCQM4Mv2 is an enhanced version with more data and features, whereas PCQM4M-LSC is part of the OGB Large-Scale Challenge, emphasizing the development of state-of-the-art graph ML models with a specific scaffold data split and feature set.   
    \item \paratitle{GEOM}~\cite{axelrod2022geom}. The Geometric Ensemble of Molecules (GEOM) dataset is a comprehensive collection designed for molecular property prediction and molecular generation. It encompasses 37 million molecular conformations for over 450,000 molecules, including 133,000 species from QM9 and 317,000 species with experimental data related to biophysics, physiology, and physical chemistry. The GEOM dataset is instrumental in the field of molecular modeling, especially for developing models that predict properties from conformer ensembles and for generative models sampling 3D conformations. 
    \item \paratitle{MolTLU}~\cite{beaini2023towards}. It is a comprehensive suite of datasets for molecular learning, categorized into ToyMix, LargeMix, and UltraLarge. These categories encompass seven distinct datasets, covering nearly 100 million molecules and over 13 billion labels across more than 3,000 tasks. ToyMix combines QM9, TOX21, and ZINC12K, offering 34 labels per molecule for foundational model training in diverse areas such as quantum machine learning and drug discovery. LargeMix merges datasets from quantum chemistry, bio-assays, and transcriptomics, providing data for millions of compounds. UltraLarge, featuring the PM6\_83M dataset, stands out for its vast scale, covering 83 million molecules for training 2D-GNNs. These datasets, with their blend of quantum and biological properties, are crucial for advancing molecular modeling. 
    \item \paratitle{ChEMBL}~\cite{gaulton2012chembl}. 
    The ChEMBL database is a vast repository of bioactive molecules with drug-like properties. It includes over 20.3 million bioactivity measurements derived from scientific experiments and literature, along with 2.4 million unique chemical compounds. These data are crucial for drug discovery and pharmacological research, providing a comprehensive resource for scientists studying the effects of molecular compounds on biological systems. This contrasts with datasets like ZINC, which contain virtual molecules that may be synthesizable but have not yet been created. Crucially, ChEMBL's data serves as an indispensable training dataset for large molecular model training and evaluation, as exemplified by its use in the GuacaMol benchmarking platform. 
    \item \paratitle{DrugBank 5.0}~\cite{wishart2018drugbank}. 
    DrugBank 5.0 includes over 2,358 approved and 4,501 investigational drugs, along with a dramatic increase in drug-drug interactions and pharmacogenomic data. This comprehensive resource offers detailed information on a broad range of drug targets, including proteins, RNA, and DNA, with notable enhancements in fields such as pharmacometabolomics, pharmacotranscriptomics, and pharmacoproteomics. Additionally, DrugBank 5.0 provides extensive data on clinical trials and integrates advanced search features and spectral data, making it an invaluable tool for pharmacological research and the pharmaceutical industry. 
    \item \paratitle{GDB-17}~\cite{ruddigkeit2012enumeration}. GDB-17 (General Database 17) is a comprehensive database containing 166.4 billion organic small molecules. It includes molecules with up to 17 atoms of C, N, O, S, and halogens, thus considerably expanding the available chemical space beyond that of its predecessors, such as GDB-13~\cite{blum2009970}. The database is characterized by a diverse array of nonaromatic heterocycles and various scaffold types, encompassing millions of isomers of known drugs. GDB-17 is especially significant in molecular modeling and drug discovery, providing a broad spectrum of molecular structures for virtual screening and the discovery of novel bioactive compounds. 
    \item \paratitle{ExCAPE-DB}~\cite{sun2017excape}. ExCAPE-DB, a substantial chemogenomics dataset, amalgamates data from ChEMBL and PubChem, comprising 998,131 unique compounds and 70,850,163 SAR data points across 1,667 targets. It is essential for constructing Molecular Activity Quantification Structure (QSAR) models to predict activities against specific targets and serves as a benchmark for appraising various machine-learning algorithms, particularly those focusing on multi-target learning. Reflecting the pharmaceutical industry's chemogenomics datasets, ExCAPE-DB includes a significant proportion of inactive compounds, offering a more accurate bioactivity profile. It covers key target families like enzymes, membrane receptors, ion channels, and transcription factors, with most compounds adhering to the Lipinski rule of five, denoting their drug-like characteristics. 
   
\end{itemize}

\paratitle{Benchmarks.} We also curate a collection of benchmarks for Mol-LLMs (in Table \ref{tab:mol-LLM-dataset}), which can be effectively utilized for training and evaluating the performance in molecular generation and prediction tasks.
\begin{itemize}   
    \item \paratitle{MoleculeNet}~\cite{wu2018moleculenet}. MoleculeNet, a benchmark dataset comprising over 700K compounds from multiple public databases, is a cornerstone in the molecular sciences for the development and evaluation of machine learning models. 
    It encompasses four key categories: quantum mechanics, physical chemistry, biophysics, and physiology, each targeting a distinct molecular science aspect. 
    Quantum mechanics datasets (QM7, QM7b, QM8, and QM9) emphasize the electronic properties, molecular geometries, and energy profiles of small molecules, which are crucial for computational chemistry. 
    Physical Chemistry datasets (ESOL, FreeSolv, and Lipophilicity) provide insights into solvation and molecular interaction, impacting drug discovery and material science. 
    Biophysics datasets (PCBA, MUV, HIV, PDBbind, and BACE) offer crucial data on protein-ligand binding affinities and various biophysical interactions. Physiology datasets (BBBP, Tox21, ToxCast, SIDER, and ClinTox) focus on molecular physiological effects, vital for assessing drug safety and environmental toxicity. 
    \item \paratitle{MARCEL}~\cite{zhu2023learning}. The MoleculAR Conformer Ensemble Learning (MARCEL) benchmark provides a comprehensive platform for evaluating the potential of learning from molecular conformer ensembles. MARCEL marks a significant shift in molecular representation learning by framing it as an ensemble learning task, focusing on diverse molecular conformer structures. This approach is based on the understanding of molecules' inherent flexibility, characterized by their ability to adopt different conformations due to bond rotations and minor vibrational perturbations. Moreover, MARCEL includes four diverse datasets encompassing a wide range of molecule- and reaction-level properties.
    \item \paratitle{GuacaMol}~\cite{brown2019guacamol}. 
    GuacaMol is an evaluation framework designed for de novo molecular design, which aims to generate molecules with specific property profiles through virtual design-make-test cycles. It offers a suite of standardized benchmarks to measure the models' fidelity in reproducing the property distribution of training sets, their ability to generate novel molecules, and their efficiency in exploring and exploiting chemical space through a range of single and multi-objective optimization tasks.
    \item \paratitle{MOSES}~\cite{polykovskiy2020molecular}. Molecular Sets (MOSES) is a benchmark suite for molecular generation. MOSES includes standardized datasets, data preprocessing tools, evaluation metrics, and molecular generation models. The dataset contains ZINC Clean Leads Dataset, which consists of 4,591,276 molecules. It is becoming a crucial tool for exploring molecular space, where models learn from large datasets to produce novel molecular structures with similar properties.  These structures are useful for virtual screening and training semi-supervised predictive models for downstream tasks. 
    \item \paratitle{ADMETlab 2.0}~\cite{xiong2021admetlab}. ADMETlab 2.0 is an advanced online platform dedicated to the accurate prediction of ADMET (Absorption, Distribution, Metabolism, Excretion, and Toxicity) properties. It hosts an extensive collection of bioactivity data, including 500,000 bioassay records and over 130 million bioactivity summary results, covering small molecules and RNAi reagents. This platform is pivotal in molecular modeling and drug discovery, offering a comprehensive framework for evaluating the ADMET profiles of chemicals. 
    {\item  \paratitle{SPECTRA}~\cite{ektefaie2024evaluating}. The SPECTRA benchmark is a cutting-edge framework designed to assess the generalizability of AI models across molecular and biological datasets by evaluating their performance on diverse training and testing splits. It illuminates the robustness of models in predicting unseen data, showcased through its application to a variety of datasets from tuberculosis to SARS-CoV-2. This targeted analysis helps in refining AI approaches for precise applications in molecular biology and bioinformatics, ensuring the development of more accurate and specialized AI models.
    \item  \paratitle{Molecule3D}~\cite{xu2021molecule3d}. Molecule3D is a benchmark dataset specifically curated for predicting ground-state 3D molecular geometries from molecular graphs, containing approximately 3.9 million molecules with precise geometries obtained via density functional theory (DFT). This dataset facilitates the development and evaluation of machine learning models in capturing the 3D spatial configurations of molecules, which are crucial for accurate molecular property prediction and other downstream tasks.}
\end{itemize}

\subsection{Evaluation} \label{sec:mol-LLM-eval}

To evaluate the capabilities of Mol-LLMs, we categorize them into four areas: property prediction, interaction prediction, reaction prediction, and molecule generation.

\paratitle{Property Prediction}. 
Molecular property prediction stands as a pivotal challenge in computational chemistry, where the goal is to accurately predict the properties of molecules, such as solubility, lipophilicity, affinity, absorption, distribution, metabolism, excretion, toxicity, and biological activity, based on their chemical structure. 
In this domain, most encoder-only models are typically pre-trained on extensive and unlabeled datasets and subsequently fine-tuned using specific data for property prediction. In the traditional domain of property prediction, BERT-based models such as ChemBERTa~\cite{chithrananda2020chemberta}, MAT~\cite{maziarka2020molecule}, Mol-BERT ~\cite{li2021mol} and MG-BERT~\cite{zhang2021mg} are widely recognized. However, in cases of small sample problems, SMILES Transformer~\cite{honda2019smiles} has demonstrated superior performance. 

\paratitle{Interaction Prediction}.
Drug-drug interaction (DDI) prediction is a crucial field in drug discovery. Drug-drug interaction (DDI) prediction refers to the process of identifying and assessing the potential effects that may occur when two or more drugs are administered concurrently. This predictive analysis is essential for understanding how different pharmaceutical compounds might interact within the body. In the domain of DDI prediction, a representative model is X-Mol~\cite{xue2020x}, which enhances the accuracy and reliability of DDI predictions, facilitating the development of safer drug combinations and reducing the risk of adverse reactions in patients. 

\paratitle{Reaction Prediction}.
Chemical reaction prediction is a significant area of study in chemistry, and it generally falls into two main categories: forward reaction prediction and retrosynthetic analysis.   
The forward reaction prediction involves predicting the products of a chemical reaction based on the reactants. It's about understanding how different chemicals react with each other under various conditions. Forward reaction prediction helps chemists anticipate the outcome of a chemical reaction, which is crucial for synthesizing new compounds, developing pharmaceuticals, and various industrial processes. In the domain of forward reaction prediction, Molecular Transformer~\cite{schwaller2019molecular} and GO-PRO~\cite{mann2021predicting} are the two classical models. The retrosynthetic analysis is the reverse of forward reaction prediction. It starts with a complex product and works backward to deduce the simpler starting materials (reactants) and the reactions needed to synthesize the product. This approach is particularly valuable in organic chemistry for designing pathways to synthesize complex organic molecules. It helps chemists identify the most efficient and practical way to produce a desired compound, often used in drug development and the synthesis of new materials.  Retrosynthesis Transformer~\cite{toniato2023enhancing} and SCROP~\cite{zheng2019predicting} are representative models in this domain. However, retrosynthesis prediction is challenging since there are massive possible synthetic routes available, and it is often difficult to navigate the direction of the retrosynthesis process. Therefore, there are GET~\cite{mao2021molecular} and Graph2SMILES~\cite{tu2022permutation} that incorporate critical topological connections between atoms, and Disconnection aware model~\cite{thakkar2023unbiasing} and RetroSynth-Diversity~\cite{toniato2023enhancing} that integrate prompts.

\paratitle{Molecule Generation}.
Molecule Generation is an emerging and critical area in computational chemistry and drug design, harnessing the power of advanced computational techniques to create novel molecular structures. This field is particularly vital in the development of new pharmaceuticals and materials, where the ability to generate molecules with desired properties can greatly accelerate the discovery process. 
Molecule generation encompasses two primary approaches: template-based design and de novo design. The former involves modifying known molecules or scaffolds. This approach starts with existing molecular frameworks, which are then altered to enhance desirable properties or reduce unwanted effects. The de novo design refers to the creation of novel molecular structures from scratch, without relying on pre-existing templates. This method is particularly valuable for discovering unique compounds with potential therapeutic or material applications. It challenges computational models to explore vast chemical spaces, requiring sophisticated algorithms capable of predicting viable and effective molecular structures. To address this task, there are lots of typical Mol-LLMs such as MolGPT~\cite{bagal2021molgpt}, SMILES GPT~\cite{adilov2021generative} and Taiga~\cite{mazuz2023molecule}.

\paratitle{Evaluation Metric}.
To quantify the performance of Mol-LLMs in property prediction, interaction prediction, and reaction prediction tasks, which primarily involve classification and regression, one can employ the standard metrics in machine learning, such as accuracy, F1-score, and correlation coefficient. For molecular generation tasks, their performance is evaluated based on their ability to produce diverse, realistic, and relevant molecules. We present several popular metrics as follows.
\begin{itemize}
    \item \paratitle{Validity}~\cite{polykovskiy2020molecular,brown2019guacamol}. This metric assesses the percentage of generated molecules that are chemically viable. Validity is crucial in ensuring that the output of generative models adheres to fundamental chemical laws, such as proper atomic valences and bond formations.
    \item \paratitle{Uniqueness}~\cite{polykovskiy2020molecular, brown2019guacamol}. It focuses on the uniqueness of the generated SMILES strings.  It is calculated for the first $K$ valid molecules in the generated set, where $K$ is typically 1,000 or 10,000, ensuring the model's capability to produce a broad range of distinct molecular structures.  This metric is particularly vital in the context of high-dimensional chemical space, as it assesses the model's ability to generate a vast array of unique molecules.
    \item \paratitle{Novelty}~\cite{polykovskiy2020molecular, brown2019guacamol}. Novelty is the fraction of the generated molecules that are not present in the training set. It measures the proportion of generated molecules that are unique compared to the training dataset. Low novelty indicates overfitting. This metric is significant in drug discovery, where the generation of novel compounds can lead to the development of new therapeutics.
    \item  \paratitle{Internal diversity (IntDiv$_p$)}~\cite{polykovskiy2020molecular}. This metric ensures that the molecules produced by generative models exhibit a wide range of chemical properties, which is defined by
    \begin{equation}\text{IntDiv}_p(G) = 1 - \sqrt[p]{\left( \frac{1}{|G|^2} \sum_{m_1, m_2 \in G} T(m_1, m_2)^p \right)}.
    \end{equation}
    The formula IntDiv$_p$($G$) evaluates the diversity of a molecule set $G$, where $G$ is the set of molecules, $m_1$ and $m_2$ are individual molecules in $G$, $T$ is the similarity measure between two molecules, and $p$ adjusts the sensitivity to similarity.    
    \item \paratitle{External diversity}~\cite{2017ChemGAN}. The external diversity of a generative model is defined as the relative diversity between the training set and a sufficiently large generated sample. $A_1$ and $A_2$ are two sets of molecules, the relative diversity $E$ of $A_1$, $A_2$ is defined by:
    \begin{equation}
    E\left(A_1, A_2\right)=\frac{1}{\left|A_1\right| \times\left|A_2\right|} \sum_{(x, y) \in A_1 \times A_2} T_d(x, y).
    \end{equation}
    $T_d(x, y)$ is the Tanimoto distance between molecule $x$ from set $A_1$ and molecule $y$ from set $A_2$. The higher the external diversity, the more distinct the generated set is from the reference set, indicating a broader exploration of the chemical space by the generative model.

\item  \paratitle{Fragment similarity (Frag)}~\cite{polykovskiy2020molecular}.
This metric serves as a comparative tool for evaluating the distribution of BRICS fragments within generated and reference molecular sets. 
Note that BRICS is a method used to fragment molecules into smaller, more manageable pieces based on retrosynthetically interesting chemical substructures. The Frag metric is formulated by calculating the cosine similarity between the substructural elements of generated molecules and those within a reference set,
\begin{equation}
    Frag(G, R) = \frac{\sum\limits_{f \in F} [c_f(G) \cdot c_f(R)]}{\sqrt{\sum\limits_{f \in F} c^2_f(G)} \sqrt{\sum\limits_{f \in F} c^2_f(R)}},
\end{equation}
where \( c_f(A) \) represents the frequency of occurrence of a substructure \( f \) within a set of molecules \( A \), and \( F \) denotes the collective set of fragments present in either the generated set \( G \) or the reference set \( R \). 

\item  \paratitle{Scaffold similarity (Scaff)}~\cite{polykovskiy2020molecular}.
This metric is similar to the fragment similarity metric. However, instead of comparing the frequencies of fragments, it compares the frequencies of Bemis-Murcko scaffolds between the generated and reference sets of molecules. Note that Bemis-Murcko scaffolds include all of a molecule’s ring structures and linker fragments connecting rings.

\item  \paratitle{Fréchet ChemNet Distance (FCD)}~\cite{polykovskiy2020molecular, brown2019guacamol}.
It measures the difference in distributions between a set of generated molecules \(G\) and a reference set \(R\) by comparing their mean vectors and covariance matrices of activations. These activations represent chemical and biological properties captured from the molecules. 
The formula for FCD is defined as follows:
\begin{equation}
\text{FCD}(G, R) = \| \mu_G - \mu_R \|^2 + \text{Tr}[\Sigma_G + \Sigma_R - 2(\Sigma_G \Sigma_R)^{1/2}].
\end{equation}
where \(\mu_G\) and \(\mu_R\) are the mean vectors of activations for the molecules in sets \(G\) and \(R\) respectively, while \(\Sigma_G\) and \(\Sigma_R\) are the corresponding covariance matrices. A low FCD value indicates generated molecules closely resemble those in the reference set, both chemically and biologically, suggesting high-quality molecular generation. 
\end{itemize}

\subsection{Summary}
\label{sec:mol-LLM-summary}
In this section, we have meticulously examined the landscape of Sci-LLMs specifically adapted to chemical molecular languages like SMILES and SELFIES. Our exploration encompasses an in-depth analysis of their unique model architectures and capabilities, focusing on how these models interpret and process chemical languages. We also delve into the datasets that are pivotal for training and benchmarking these models, highlighting the criteria that are essential for evaluating their performance, particularly in the context of the increasingly popular field of molecular generation.

	\clearpage
	
	\section{Protein Large Language Models} \label{sec-protein}
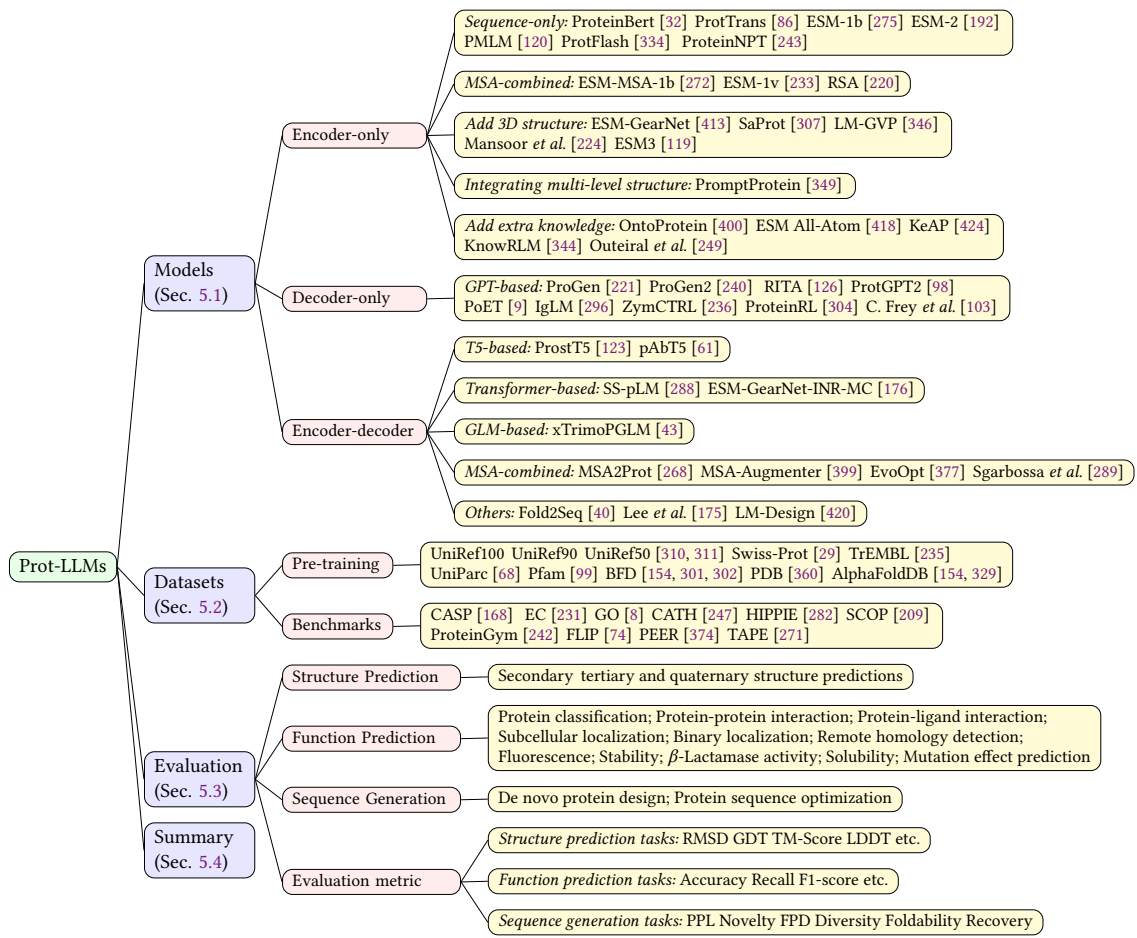
\begin{figure*}[h]
\centering
\resizebox{1.0\textwidth}{!}{
\begin{forest}
  for tree={
  grow=east,
  reversed=true,
  anchor=base west,
  parent anchor=east,
  child anchor=west,
  base=left,
  font=\small,
  rectangle,
  draw,
  rounded corners,align=left,
  minimum width=2.5em,
  inner xsep=4pt,
  inner ysep=1pt,
  },
  where level=1{text width=5em,fill=blue!10}{},
  where level=2{text width=5em,font=\footnotesize,fill=pink!30}{},
  where level=3{font=\footnotesize,yshift=0.26pt,fill=yellow!20}{},
  [Prot-LLMs, fill=green!10
        [Models\\(Sec.~\ref{sec:prot-LLM-overview}),text width=4em
            [Encoder-only, text width= 5.5em
              [\emph{Sequence-only:} 
        ProteinBert~\cite{brandes2022proteinbert}\, 
        ProtTrans~\cite{ProtTrans9477085}\,
              ESM-1b~\cite{rives2019biological}\,
              ESM-2~\cite{lin2022language}\,  \\
              PMLM ~\cite{corr/abs-2110-15527}\,    
              ProtFlash~\cite{WANG2023101600} \,
              ProteinNPT ~\cite{Pascal2023ProteinNPT}\,
              ]              
              [\emph{MSA-combined:} 
              ESM-MSA-1b~\cite{rao2021msa}\,
              ESM-1v~\cite{meier2021language}\,
              RSA~\cite{ma2023retrieved}
              ]              
              [\emph{Add 3D structure:}             
              ESM-GearNet~\cite{zhang2023enhancing}\, 
              SaProt~\cite{su2023saprot}\,
              LM-GVP~\cite{wang2022lm}\,\\
              Mansoor \textit{et al.}~\cite{mansoor2021toward}\,
              ESM3~\cite{thomas2024Simulating}
              ] 
              [\emph{Integrating multi-level structure:}    PromptProtein~\cite{wang2023multilevel}\,
              ]
              [\emph{Add extra knowledge:}             
              OntoProtein~\cite{2022arXiv220111147Z}\,
              ESM All-Atom ~\cite{zheng2024multi}\,
              KeAP~\cite{2023arXiv230113154Z}\,\\
              KnowRLM ~\cite{wangknowledge}\,
              Outeiral \textit{et al.} ~\cite{outeiral2024codon}\,
              ]
            ]            
            [Decoder-only, text width= 5.5em
              [\emph{GPT-based:} 
              ProGen ~\cite{2020arXiv200403497M}\,
              ProGen2~\cite{nijkamp2023progen2} \,
              RITA~\cite{2022arXiv220505789H}\, ProtGPT2~\cite{Ferruz2022.03.09.483666}\, \\ 
               PoET~\cite{truong2023poet}\, 
               IgLM~\cite{shuai2021generative}\, 
               ZymCTRL~\cite{munsamy2022zymctrl}\,
               ProteinRL~\cite{sternke2023proteinrl}\,
               C. Frey \textit{et al.} ~\cite{frey2023protein}\,
              ]
            ]            
            [Encoder-decoder, text width=5.5em
              [\emph{T5-based:}               
              ProstT5 ~\cite{Heinzinger2023.07.23.550085}\,
              pAbT5 ~\cite{chu2023generative}           
              ]  
              [\emph{Transformer-based:}
              SS-pLM~\cite{serrano2023efficient}\,
              ESM-GearNet-INR-MC ~\cite{lee2023pre}
              ]
              [\emph{GLM-based:}               
              xTrimoPGLM~\cite{chen2023xtrimopglm}
              ]
              [\emph{MSA-combined:}              
              MSA2Prot ~\cite{ram2022few}\,              
              MSA-Augmenter ~\cite{2023arXiv230601824Z}\,
              EvoOpt~\cite{yamaguchi2022evoopt}\,
              Sgarbossa \textit{et al.}~\cite{sgarbossa2023generative}
              ]                
              [\emph{Others:}               
              Fold2Seq~\cite{cao2021fold2seq}\,
              Lee \textit{et al.}~\cite{lee2022protein}\,
              LM-Design~\cite{Zheng2023.02.03.526917}
              ]
            ]    
        ]
        [Datasets\\
        (Sec.~\ref{sec:prot-LLM-dataset}),text width=4em
            [Pre-training, text width=4em
             [UniRef100\,
              UniRef90\,
              UniRef50~\cite{suzek2007uniref, suzek2015uniref}\,
              Swiss-Prot~\cite{boutet2016uniprotkb}\,
              TrEMBL~\cite{m1999edittotrembl}\, \\
            UniParc~\cite{uniprot2023uniprot}\,
              Pfam~\cite{finn2006pfam}\,
              BFD~\cite{steinegger2018clustering, steinegger2019protein, jumper2021highly}\,
              PDB~\cite{10.1093/nar/gky949}\,
              AlphaFoldDB~\cite{jumper2021highly, 10.1093/nar/gkab1061}
             ]            
            ]
            [Benchmarks, text width=4em
              [CASP~\cite{kryshtafovych2021critical}  \,
              EC~\cite{10.1093/nar/gkn582}\,
              GO~\cite{ashburner2000gene}\,
              CATH~\cite{orengo1997cath}\,
              HIPPIE~\cite{schaefer2012hippie}\,
              SCOP~\cite{lo2000scop}\,\\
              ProteinGym~\cite{Notin2023.12.07.570727}\,
              FLIP~\cite{dallago2021flip}\,
              PEER~\cite{xu2022peer}\,
              TAPE~\cite{tape2019}       
              ]
            ] 
        ]
        [Evaluation\\
        (Sec.~\ref{sec:prot-LLM-eval}),text width=4em  
            [Structure Prediction, text width=7em   
            [Secondary\, tertiary and quaternary structure predictions]     
            ]
            [Function Prediction, text width=7em   
            [
            Protein classification; 
            Protein-protein interaction;
            Protein-ligand interaction;\\
            Subcellular localization;
            Binary localization;
            Remote homology detection;\\ 
            Fluorescence; 
            Stability; 
            $\beta$-Lactamase activity;
            Solubility; 
            Mutation effect prediction
            ]
            ]
            [Sequence Generation, text width=7em    
            [
            De novo protein design; Protein sequence optimization]
            ]
            [Evaluation metric, text width=7em    
            [
            \textit{Structure prediction tasks:} RMSD GDT TM-Score LDDT etc.
            ]   
            [
            \textit{Function prediction tasks:} Accuracy Recall F1-score etc.
            ]
            [
            \textit{Sequence generation tasks:} PPL Novelty FPD Diversity Foldability Recovery 
            ]
            ]
        ]
        [Summary\\
        (Sec.~\ref{sec:prot-LLM-summary}),text width=4em
        ]
    ]
\end{forest}
}
\caption{Chapter overview of Prot-LLMs.}
\label{fig:prot-LLM-overview}
\end{figure*}

In the past years, large language models have become increasingly influential in protein research, offering novel insights and capabilities in understanding and manipulating proteins. 
In this section, we present a comprehensive review of LLMs for proteins (named Prot-LLMs), encompassing detailed discussions on their model architectures, utilized datasets, various capabilities, and corresponding evaluation criteria. The overview of this section is shown in Figure~\ref{fig:prot-LLM-overview}.

\subsection{Models}
\label{sec:prot-LLM-overview}
In this section, we also classify Prot-LLMs into three main types based on their specific architectures: encoder-only, decoder-only, and encoder-decoder models. These distinct architectures are well-suited for diverse protein research applications. For example, the encoder-only models primarily serve protein function or property prediction purposes, while the decoder-only models are predominantly employed for protein generation tasks. 
Table \ref{tab:Prot-LLMs} provides a summary of Prot-LLMs.

\subsubsection{Encoder-only Models} \ \\
Most of the encoder-only Prot-LLMs are built upon the encoder of Transformer, which enables the encoding of protein sequences or structures into fixed-length vector representations. These representations facilitate the identification of patterns and features within proteins, thereby enhancing subsequent analysis and prediction tasks.
Self-supervised learning is employed to acquire protein representation, such as masked language modeling (MLM) tasks that reconstruct corrupted tokens based on the surrounding sequence. Prominent pre-trained protein sequence encoders include  ESM-1b~\cite{rives2019biological}, ESM-1v~\cite{meier2021language}, and ESM-2~\cite{lin2022language}, ProteinBert~\cite{brandes2022proteinbert}, ProtTrans~\cite{ProtTrans9477085}.
The ESM series~\cite{rives2019biological, meier2021language, lin2022language} mainly employs a Transformer's encoder (i.e., BERT~\cite{devlin2018bert} and RoBERTa~\cite{liu2019roberta}) to predict protein structure and function, which utilizes extensive sequence information from protein databases without relying on manual annotations of the sequences.
ProteinBert~\cite{brandes2022proteinbert} enhances the classical BERT architecture by incorporating a novel pretraining task for predicting protein functionality. 
It separates local (character-level) and global (sequence-level) representations, enabling multitask processing in a principled way. ProtTrans~\cite{ProtTrans9477085} trained several auto-encoder models (includes BERT~\cite{devlin2018bert}, Albert~\cite{lan2019albert}, Electra~\cite{clark2020electra}) on a vast of sequence data. 
Several works have attempted to improve the encoder architecture and the training method. PMLM~\cite{corr/abs-2110-15527} introduced Pairwise Masked Language Model (PMLM), to directly pretrain the encoder with a focus on capturing co-evolutionary information reflected in residue co-variation within the sequence. This approach considers dependencies between masked tokens. ProtFlash~\cite{WANG2023101600} suggested a mixed-chunk attention mechanism that combines multiple positional encodings and a hybrid block attention mechanism incorporating both local and global attention, effectively reducing model complexity. ProteinNPT~\cite{Pascal2023ProteinNPT} developed a non-parametric Transformer specifically designed for protein sequences, making it particularly suitable for scenarios involving sparse labels and multi-task learning.

The Multiple Sequence Alignment (MSA) is a computational method that can reveal common features and variation patterns among sequences.  By aligning multiple sequences, MSA can unveil shared evolutionary relationships between them and facilitate the identification of functional regions and structural domains. Therefore, MSA has been widely used in protein models. 
For example, ESM-MSA-1b (MSA Transformer)~\cite{rao2021msa} extends the self-attention mechanism to the MSA setting, which interleaves self-attention across rows and columns to capture dependencies between amino acids and between sequences. Notably, MSA Transformer has become a crucial component of AlphaFold2~\cite{fang2023method} and AlphaMissence~\cite{cheng2023accurate}.
However, the utilization of MSA incurs significant computational overheads, and cannot handle orphan proteins. An important alternative method, called Retrieved Sequence Augmentation (RSA)~\cite{ma2023retrieved}, eliminates the need for additional alignment or pre-processing steps. RSA effectively links query protein sequences to a set of structurally or functionally similar sequences in the database and aggregates these sequences for subsequent function prediction tasks.

Given that the function of a protein is intricately linked to its structure, another reasonable approach for protein representation would involve the integration of its 3D conformation.
There are some recent efforts in this direction. 
ESM-GearNet~\cite{zhang2023enhancing} combines ESM-1b and GearNet \cite{xie2022gearnet} (the state-of-the-art protein structure encoder) to enhance model performance.
SaProt~\cite{su2023saprot} introduces a structure-aware vocabulary that integrates residue tokens with structure tokens. 
LM-GVP~\cite{wang2022lm} compose Transformer blocks and a graph network derived from the protein 3D structure.
Mansoor \textit{et al.}~\cite{mansoor2021toward} encoded protein sequence and structure through joint training in a semi-supervised manner. 
PromptProtein~\cite{wang2023multilevel} utilizes prompt-guided multi-task pretraining to fuse different levels of protein structure. 
{ESM3~\cite{thomas2024Simulating} is an advanced multimodal generative language model that effectively integrates protein sequence, structure, and function analysis. It exhibits exceptional capability in comprehending intricate prompts by leveraging its evolutionary simulation capabilities, making it highly adaptable for biological alignment tasks.}
It is worth noting that numerous 3D structure encoders (like GearNet \cite{xie2022gearnet} and GBPNet \cite{aykent2022gbpnet}) have been proposed to model protein structural information; however, they are beyond the scope of this survey which primarily focuses on LLMs.

Additionally, some approaches propose to introduce extra knowledge such as Gene Ontology (GO) to enhance the protein representation.
OntoProtein~\cite{2022arXiv220111147Z} considered GO as a factual knowledge graph, which was used to enhance protein representations. 
{ESM All-Atom~\cite{zheng2024multi} achieves atom-scale and residue-scale unified molecular modeling by pretraining on multi-scale protein sequences and utilizing a multi-scale position encoding to capture relationships among residues and atoms.}
KeAP~\cite{2023arXiv230113154Z} proposed a knowledge-injected auto-encoder that performs token-level knowledge graph exploration for protein representation learning.
{KnowRLM ~\cite{wangknowledge} stands out for its ability to utilize contextual amino acid information from knowledge graphs, thus attaining advantages from both statistical patterns of protein sequences and biochemical properties of amino acids.}
{Some studies improve protein learning representation by enriching pre-trained biological data. For example, Outeiral \textit{et al.}~\cite{outeiral2024codon} show that large language models trained on codons, instead of amino acid sequences, provide high-quality representations that outperform comparable state-of-the-art models across a variety of tasks.}

\begin{table*}[tbp]
\centering
\caption{Summary of Prot-LLMs}
\label{tab:Prot-LLMs}
\footnotesize
\renewcommand\tabcolsep{2.5pt}
\begin{tabular}{llcccccc}
\toprule
  & Model & Time  &  \#Parameters & Base model & Pretraining Dataset &  Capability  & \makecell[c]{Open-\\source} \\
\midrule

   \multirow{21}{*}{Encoder-only}
   &\href{https://github.com/facebookresearch/esm}{ESM-1b} ~\cite{rives2019biological}   & 2020.02    & 650M  & RoBERTa & UniRef50 & \makecell{Secondary struct. pred., \\Contact pred., etc.}& \checkmark \\
   &\href{https://github.com/facebookresearch/esm}{ESM-MSA-1b} ~\cite{rao2021msa}   & 2021.02    & 100M  & ESM-1b & UniRef50 & \makecell{Secondary struct. pred., \\Contact pred., etc.}& \checkmark \\
   &\href{https://github.com/facebookresearch/esm}{ESM-1v} ~\cite{meier2021language}   & 2021.02    & 	650M  & ESM-1b & UniRef90 & Mutation effect pred. & \checkmark \\
   &\href{https://github.com/agemagician/ProtTrans}{ProtTrans} ~\cite{ProtTrans9477085}   & 2021.07    & 	-  & \makecell{BERT, Albert, \\Electra} & UniRef, BFD & \makecell{Secondary struct. pred., \\Func. pred., etc}  & \checkmark \\ 
   &PMLM~\cite{corr/abs-2110-15527}   & 2021.07    & 87M - 731M  & Trans. enc. & Uniref50/Pfam & Contact pred.  & $\times$ \\
   &Mansoor \textit{et al.} ~\cite{mansoor2021toward}   & 2021.09    & 100M  & ESM-1b & - & Mutation effect pred.   & $\times$ \\
   
   &\href{https://github.com/nadavbra/protein_bert}{ProteinBERT} ~\cite{brandes2022proteinbert}   & 2022.02    & 16M  & BERT & \makecell{UniRef90} &  \makecell{Func. pred.}  & \checkmark \\ 
   &\href{https://github.com/aws-samples/lm-gvp}{LM-GVP}
   ~\cite{wang2022lm}   & 2022.04    & -  & Trans. enc & - & Func. pred. & \checkmark \\
   &\href{https://github.com/HKUNLP/RSA}{RSA} ~\cite{ma2023retrieved}   & 2022.05    & -  & ESM-1b & - & Func. pred. & \checkmark \\
   &\href{https://github.com/zjunlp/OntoProtein}{OntoProtein} ~\cite{2022arXiv220111147Z}   & 2022.06     & - &  BERT & ProteinKG25 & Func. pred.  & \checkmark\\

&\href{https://github.com/facebookresearch/esm}{ESM-2} ~\cite{lin2022language}   & 2022.07    & \makecell{8M - 15B}  & RoBERTa & UniRef50 & Func. pred., Struct. pred.  & \checkmark \\

&\href{https://github.com/HICAI-ZJU/PromptProtein}{PromptProtein} ~\cite{wang2023multilevel}   & 2023.02    &  650M  & RoBERTa & \makecell{UniRef50, PDB} & Func. pred. & \checkmark \\

   &\href{https://github.com/RL4M/KeAP}{KeAP} ~\cite{2023arXiv230113154Z}   & 2023.02    & -  & RoBERTa & ProteinKG25 & Func. pred.  & \checkmark\\
   
   &\href{https://github.com/ISYSLAB-HUST/ProtFlash}{ProtFlash} ~\cite{WANG2023101600}   & 2023.10    & 79M/174M  & Trans. enc & UniRef50 & Func. pred.  & \checkmark \\
   &\href{https://github.com/DeepGraphLearning/ESM-GearNet}{ESM-GearNet} ~\cite{zhang2023enhancing}   & 2023.10    & -  & ESM-1b, GearNet & - & Func. pred.  & \checkmark \\

   &\href{https://github.com/westlake-repl/SaProt}{SaProt} ~\cite{su2023saprot} & 2023.10    &  650M  & BERT  & -  & Mutation effect pred. & \checkmark \\
   & ProteinNPT ~\cite{Pascal2023ProteinNPT}   & 2023.12    & -  & Trans. enc.  & - & Fitness pred., Redesign & $\times$  \\
   & \href{https://github.com/oxpig/CaLM}{Outeiral \textit{et al.}} ~\cite{outeiral2024codon}   & 2024.02    &  10M - 5B  & Trans. enc.  & \makecell{European Nucleotide\\  Archive} & Protein represent learning & \checkmark  \\
   & ESM All-Atom ~\cite{zheng2024multi}   & 2024.06    &  35M  & RoBERTa   & AlphaFold DB & Unified Molecular Modeling & $\times$  \\
   & KnowRLM ~\cite{wangknowledge}   & 2024.06    &  -  & Trans. enc.  & - & Protein Directed Evolution & $\times$  \\
   &\href{https://github.com/evolutionaryscale/esm}{ESM3} ~\cite{thomas2024Simulating} & 2024.06    &  98B  & RoBERTa   & PDB  & Seq. pred., Func. pred., Struct. pred. & \checkmark \\

   \midrule
   \multirow{10}{*}{Decoder-only} &\href{https://github.com/salesforce/progen}{ProGen} ~\cite{2020arXiv200403497M}   & 2020.03    & 1.2B    & GPT  & \makecell{Uniparc \\SWISS-Prot} & Functional prot. gen. & \checkmark \\
   &\href{https://huggingface.co/nferruz/ProtGPT2}{ProtGPT2} ~\cite{Ferruz2022.03.09.483666}   & 2021.01    & 738M  & GPT  & Uniref50 & \makecell{De novo protein design \\and engineering}  & \checkmark\\
   &\href{https://huggingface.co/AI4PD/ZymCTRL}{ZymCTRL} ~\cite{munsamy2022zymctrl}   & 2022.01    & 738M  & GPT & BRENDA & Functional enzymes gen.  & \checkmark \\
   &RITA ~\cite{2022arXiv220505789H}   & 2022.05    & 1.2B  & GPT   & UniRef100 & Functional prot. gen. & $\times$\\
   &\href{https://github.com/Graylab/IgLM}{IgLM} ~\cite{shuai2021generative}   & 2022.12    & 13M  & GPT & - & Antibody design  & \checkmark \\
   & \href{https://github.com/salesforce/progen}{ProGen2} ~\cite{nijkamp2023progen2}   & 2023.10    & 151M - 6.4B  &GPT & \makecell{Uniref90, \\BFD30, PDB} & Functional prot. gen.  & \checkmark \\
   &ProteinRL~\cite{sternke2023proteinrl}   & 2023.10   &  764M  & GPT  &  - & Prot. design  & $\times$\\
   &PoET ~\cite{truong2023poet}   & 2023.11    & 201M  & GPT  & - & Prot. family. gen.  & $\times$ \\
   
   & C. Frey \textit{et al.} ~\cite{frey2023protein}   & 2024.03    & 9.87M/1.03M  & GPT  & \makecell{hu4D5 antibody\\ mutant} & Functional prot. gen. & $\times$  \\
   \midrule

   \multirow{10}{*}{Encoder-Decoder} &\href{https://github.com/IBM/fold2seq}{Fold2Seq} ~\cite{cao2021fold2seq}   & 2021.01   & -  & Transformer   & - & Prot. design & \checkmark \\
   &MSA2Prot ~\cite{ram2022few}   & 2022.04   &  -  & Transformer  &  - &  Prot. gen., Variant func. pred.  & $\times$\\
   &\href{https://doi.org/10.5281/zenodo.7684052}{Sgarbossa \textit{et al.}}~\cite{sgarbossa2023generative}  & 2023.02   & -  & MSA Transformer  & -  & Prot. gen. & \checkmark \\
   &Lee \textit{et al.}~\cite{lee2022protein} & 2023.02   & 150M  & Transformer   & -  & Prot. design & $\times$ \\
   &\href{https://github.com/BytedProtein/ByProt}{LM-Design} ~\cite{Zheng2023.02.03.526917}   & 2023.02    & 664M  & Transformer  & -  & Prot. design & \checkmark \\
   &\href{https://github.com/Magiccircuit/MSA-Augmentor}{MSA-Augmenter} ~\cite{2023arXiv230601824Z}   & 2023.06    & 260M  & Transformer  & Uniref50 & MSA gen. & \checkmark \\
   &\href{https://github.com/mheinzinger/ProstT5}{ProstT5} ~\cite{Heinzinger2023.07.23.550085}   & 2023.07   & 3B  & T5  &  PDB &  Seq.-struct. translation  & \checkmark \\
   &xTrimoPGLM ~\cite{chen2023xtrimopglm}   & 2023.07   &  100B  & GLM  &  \makecell{Uniref90, \\ColdFoldDB} & Prot. gen., Func. pred.  & $\times$\\
   &SS-pLM~\cite{serrano2023efficient}  & 2023.08   &  14.8M  & Transformer  & Uniref50  & Prot. gen.  & $\times$\\
   &pAbT5 ~\cite{chu2023generative}   & 2023.10   &  -  & T5  &  - & Prot. design  & $\times$\\
   & \makecell[l]{ESM-GearNet-\\INR-MC~\cite{lee2023pre}}   & 2024.04   &  -  &  Transformer  &  \makecell[c]{Swiss-Prot, \\AlphaFoldDB} & Prot. gen  & $\times$\\

\bottomrule
\end{tabular}
\end{table*}

\subsubsection{Decoder-only Models} \ \\

Decoder-based Prot-LLMs play a predominant role in the generation of novel proteins, serving as a crucial tool in protein engineering and drug design. GPT~\cite{radford2019GPT2}, a well-established decoder-only architecture, has gained extensive utilization in Prot-LLMs. 
A representative model is ProGen ~\cite{2020arXiv200403497M}, which utilizes GPT for controllable protein generation. It has been trained on a dataset comprising 280M protein sequences, accompanied by conditioning tags that encode diverse annotations encompassing taxonomic, functional, and locational information.
ProGen2~\cite{nijkamp2023progen2} extends the model to 6.4B parameters and is trained on diverse sequence datasets extracted from over one billion proteins from genomic, metagenomic, and immune repertoire databases. ProtGPT2~\cite{Ferruz2022.03.09.483666} is another GPT-based autoregressive model that effectively generates protein sequences exhibiting amino acid compositions and disorder propensities comparable to those observed in natural proteins.
RITA~\cite{2022arXiv220505789H} presents a suite of autoregressive generative models, which is the first systematic investigation of how capabilities evolve with model size for these protein models. 

The decoder-based protein language models have been widely applied in specific protein designs. 
Timothy \textit{et al.} introduced the protein evolutionary Transformer (PoET), an autoregressive generative model of the distribution over protein families, which can generate sets of related proteins.
Zheng \textit{et al.} proposed the LM-Design~\cite{Zheng2023.02.03.526917} that uses language models for structure-based protein design (i.e., inverse folding). 
ZymCTRL~\cite{munsamy2022zymctrl} trained a conditional language model on the BRENDA enzyme database \cite{chang2021brenda}, aimed at designing customized artificial enzymes by generating specific enzyme classes based on user prompts. 
IgLM~\cite{shuai2021generative} employed autoregressive sequence generation for antibody design. 
{C. Frey \textit{et al.}~\cite{frey2023protein} evaluate the robustness of our approach on generative modeling of antibody proteins and introduce the distributional conformity score to benchmark protein generative models. }
%
{ProteinRL~\cite{sternke2023proteinrl} is a reinforcement learning framework for fine-tuning generative PLMs for proteins optimized for specific sequence and/or structural properties. ProteinRL fine-tuning guides the PLM towards generating sequences optimized for the defined properties, extending to values rarely or never seen in natural sequences or sequences generated without ProteinRL fine-tuning. }

\subsubsection{Encoder-decoder Models} \ \\
The encoder-decoder architecture is commonly used for sequence-to-sequence tasks. 
In the field of protein research, 
ProstT5 ~\cite{Heinzinger2023.07.23.550085}, a protein language model based on the T5 architecture \cite{raffel2020exploring}, is an exemplary encoder-decoder model that facilitates translation between protein sequences and structures. The conversion of protein structure from 3D to 1D is achieved through the utilization of 3Di-tokens introduced by Foldseek \cite{van2022foldseek}.
pAbT5~\cite{chu2023generative} is another T5-based model that takes into consideration the constraints imposed by protein-protein interactions on generation and design. It can generate complementary heavy or light chains from their pairing partners.
xTrimoPGLM~\cite{chen2023xtrimopglm}, based on General Language Model (GLM) architecture \cite{du2021glm},  is a unified protein language model for understanding and generating protein sequences. It has an unprecedented scale of 100B parameters and consumes 1 trillion training tokens.
Considering that the training of protein LLMs involves millions to billions of sequences and billions of parameters, Serrano \textit{et al.}~\cite{serrano2023efficient} introduced the Small-Scale protein Language Model (SS-pLM), which significantly reduces the computational burden, democratizing the use of foundational models in protein generation.

Building upon the Transformer, several models have incorporated MSA modules. MSA2Prot ~\cite{ram2022few} is an MSA-to-protein Transformer, developed axial and cross attentions for encoder and decoder to model sequence probabilities autoregressively. 
MSA-Augmenter ~\cite{2023arXiv230601824Z} utilizes protein-specific attention mechanisms and large-scale MSAs to generate novel protein sequences that are not present in existing databases. 
Sgarbossa \textit{et al.}~\cite{sgarbossa2023generative} proposed an iterative method that directly utilizes the masked language modeling to generate sequences within the MSA Transformer framework. 

In the field of structure-based protein design, Cao \textit{et al.}~\cite{cao2021fold2seq} proposed a Fold2Seq framework, a generative approach based on the Transformer architecture, which aims at designing protein sequences conditioned on specific target folds. To capture intricate sequence-structure relationships, Fold2Seq jointly learns embeddings for both sequences and folding structures, which are derived from the density of secondary structure elements in 3D voxels.

{While recent studies have successfully employed sequence- or structure-based representations to address multiple tasks in protein science, there has been significant oversight in incorporating protein surface information, a critical factor for protein function. Therefore, ESM-GearNet-INR-MC ~\cite{lee2023pre} presents a pre-training strategy that incorporates information from protein sequences, 3D structures, and surfaces to improve protein representation learning. }

Additionally, some approaches utilize reinforcement learning to improve the generation quality. Lee \textit{et al.}~\cite{lee2022protein} efficiently navigates the latent representation space rather than the protein sequence space by formulating the sequence design task as a Markov decision process and employing model-based reinforcement learning. This approach enables them to generate proteins with enhanced functionality and cellular fitness.

\begin{table*}[htbp]
\centering
\caption{Summary of datasets for Prot-LLMs}
\label{tab:prot-LLM-dataset}
\footnotesize
\renewcommand\tabcolsep{2.5pt}
\begin{tabular}{llcccH}
\toprule
    &Dataset & Last updated  &  Scale &  Keywords  & Type \\
\midrule

   \multirow{11}{*}{Pretraining}
   &\href{https://www.uniprot.org/uniref?query=*}{UniRef100}~\cite{suzek2007uniref, suzek2015uniref}  & \multirow{3}{*}{2023.11}   & 314M  & Complete collection of protein sequences from UniProtKB    & \multirow{3}{*}{Regression}  \\
   &\href{https://www.uniprot.org/uniref?query=*}{UniRef90}~\cite{suzek2007uniref, suzek2015uniref}  &     &  150M  &  Cluster UniRef100 sequences at 90$\%$ sequence identity level   &  \\
   &\href{https://www.uniprot.org/uniref?query=*}{UniRef50}~\cite{suzek2007uniref, suzek2015uniref}  &     & 53M  & Cluster UniRef100 sequences at 50$\%$ sequence identity level    &   \\
\cmidrule{2-5}
  &\href{https://www.uniprot.org/uniprotkb?query=*}{UniProtKB/Swiss-Prot}~\cite{boutet2016uniprotkb}  & \multirow{2}{*}{2023.11}   & 570K  &  High-quality, manually curated protein sequence database   & Regression  \\
  &\href{https://www.uniprot.org/uniprotkb?query=*}{UniProtKB/TrEMBL}~\cite{m1999edittotrembl} &    & 251M  &  Computationally annotated protein sequence database  & Regression  \\
  \cmidrule{2-5}
   &\href{https://www.uniprot.org/uniparc?query=*}{UniParc}~\cite{uniprot2023uniprot}  &  2023.11   & 632M  & Comprehensive and non-redundant protein sequence database    & Regression  \\

  &\href{https://www.ebi.ac.uk/interpro/entry/pfam/}{Pfam}~\cite{finn2006pfam} &  2023.09   & 47M  &  Protein family database & Regression  \\
    &\href{https://bfd.mmseqs.com/}{BFD}~\cite{steinegger2018clustering, steinegger2019protein, jumper2021highly} &  2021.07   &2.5B  & Protein sequences from multiple databases and resources & Regression \\
    &\href{https://www.rcsb.org/}{PDB}~\cite{10.1093/nar/gky949}  &  2023.12   & 214K  & Experimentally determined accurate protein structures    & Regression  \\ 
    &\href{https://proteininformationresource.org/}{PIR}~\cite{wu2004pirsf}  &  2024.03   & 513M  & Information on functionally annotated protein sequences    & Regression  \\ 
    
    &\href{https://alphafold.ebi.ac.uk/}{AlphaFoldDB}~\cite{jumper2021highly, 10.1093/nar/gkab1061}  & 2021.11 & 200M & Protein structures predicted by AlphaFold & Regression\\
\midrule
   \multirow{12}{*}{Benchmark} 
   &\href{https://predictioncenter.org/}{CASP}~\cite{kryshtafovych2021critical}  & 2022.01    & -  & Structure prediction competition  & Regression  \\
   &\href{https://www.enzyme-database.org/}{EC}~\cite{10.1093/nar/gkn582}  & 2023.11    & 2.6 M  & Enzymes classification database    & Regression  \\
   &\href{https://geneontology.org/}{GO}~\cite{ashburner2000gene}  & 2023.11    & 1.5M  & Gene Ontology knowledgebase    & Regression  \\
   &\href{http://www.cathdb.info}{CATH}~\cite{orengo1997cath}  & 2023.02    & 151M  & Classification of protein structures   & classification  \\ 
  &\href{http://cbdm-01.zdv.uni-mainz.de/~mschaefer/hippie/}{HIPPIE}~\cite{schaefer2012hippie}   & 2022.04    & 39K   & Protein-protein interaction networks    & Classification   \\
  &\href{http://scop.berkeley.edu}{SCOP}~\cite{lo2000scop}  & 2023.01    & 914K  & Protein structure classification    & Classification  \\
  &\href{https://proteingym.org/}{ProteinGym}~\cite{Notin2023.12.07.570727} & 2022.12    & $\sim300$K   & Predict the effects of protein mutations & Classification   \\
    &\href{https://benchmark.protein.properties}{FLIP}~\cite{dallago2021flip} & 2022.01    & $\sim320$K   &  Fitness landscape prediction (AAV, Thermostability, GB1)   & Classification   \\
    &\href{https://github.com/DeepGraphLearning/PEER_Benchmark}{PEER}~\cite{xu2022peer} & 2022.11    & $\sim390$K   & \makecell{Protein function, Localization, Structure prediction, \\Protein-protein interaction,  Protein-ligand interaction}    & Classification   \\
     &\href{https://github.com/songlab-cal/tape}{TAPE}~\cite{tape2019} & 2021.09    & $\sim120$K   & \makecell{Remote homology detection, Secondary structure, \\ Contact, Fluorescence, Stability prediction}   & Classification   \\
     &\href{https://reactome.org/}{Reactome}~\cite{jassal2020reactome}			&2023.12			& $\sim3$M			&Biological interactions and pathways\\

     &\href{https://cn.string-db.org/}{STRING}~\cite{szklarczyk2023string}			& 2022.11			& 59.3M		& Protein-Protein interaction networks  \\

     &\href{https://thebiogrid.org/}{BioGRID}~\cite{oughtred2019biogrid}			& 2023.12			& 271k		& Genetic and protein interactions \\

     &\href{https://www.ebi.ac.uk/interpro/}{InterPro}~\cite{paysan2023interpro}			& 2024.01			& $\sim41$k		& Classification of protein families \\

     &\href{https://dip.doe-mbi.ucla.edu/dip/Main.cgi}{DIP}~\cite{xenarios2002dip}  &  2020.02   & 81K  & Protein-protein interaction information    \\ 

     &\href{https://prosite.expasy.org/prosite_ref.html}{PROSITE}~\cite{sigrist2012new}  &  2024.03   & 103K  & Collection of signatures that identify patterns or profiles in proteins   \\

\bottomrule
\end{tabular}
\end{table*}

\subsection{Datasets} \label{sec:prot-LLM-dataset}
Protein datasets can be classified into two categories based on the availability of annotations: pre-training datasets and benchmarks. The former, which lacks labels, is commonly employed for self-supervised pre-training, whereas the latter, with labeled data, is utilized for supervised fine-tuning or model evaluation. The following present popular pre-training datasets and benchmarks for Prot-LLMs, as summarized in Table \ref{tab:prot-LLM-dataset}.

\paratitle{Pre-training Datasets:} \vspace{-0.5em}
\begin{itemize}
    \item \paratitle{UniProtKB (Swiss-Prot and TrEMBL)}~\cite{boutet2016uniprotkb, m1999edittotrembl}. It is a central hub for the collection of functional information on proteins, with extensive and accurately annotated protein sequence data. It comprises Swiss-Prot and TrEMBL, where the former is known for its high level of annotations, accuracy, and reliability, because they are manually annotated and reviewed by experts. TrEMBL contains protein sequences that are automatically annotated and have not yet been reviewed by human curators. 
    \item \paratitle{UniRef100, UniRef90, UniRef50}~\cite{suzek2007uniref, suzek2015uniref}. UniRef (UniProt Reference Clusters) is a system of databases that provide clustered sets of protein sequences from UniProtKB and selected UniProt Archive records. The primary purpose of UniRef is to provide a comprehensive and non-redundant dataset that improves the efficiency of sequence similarity searches. The databases are organized into three main clusters: UniRef100, UniRef90, and UniRef50, where the latter two are created by clustering UniRef100 sequences that have at least 90\% and 50\% sequence identity to each other, respectively. 
    \item \paratitle{UniParc}~\cite{uniprot2023uniprot}. UniParc (UniProt Archive) is an important component of the UniProt database family, serving as a resource for storing non-redundant protein sequences. Unlike other members of the UniProt database family, UniParc does not provide protein annotation and functional information; it focuses solely on storing protein sequences themselves.
    \item \paratitle{Pfam}~\cite{finn2006pfam}. Pfam is a widely used protein family database that is used for the classification and annotation of protein sequences. It plays a crucial role in understanding protein function, structure, and evolution. It is built based on the shared conserved domains and functional modules found in protein sequences, providing researchers with classification, annotation, and functional prediction of protein sequences.
     \item \paratitle{BFD}~\cite{steinegger2018clustering, steinegger2019protein, jumper2021highly}. The Big Fantastic Database (BFD)  is a comprehensive collection of protein sequences gathered from various sources. It includes a vast array of sequence data (2.5B), which is crucial for training machine learning models in tasks like protein structure prediction. 
     \item \paratitle{PDB}~\cite{10.1093/nar/gky949}.The Protein Data Bank (PDB) is a global protein structure database used for storing and sharing resolved 3D protein structure data. It archives detailed information about the 3D structures of proteins, nucleic acids, and complex assemblies. These structures are determined using experimental methods like X-ray crystallography, NMR spectroscopy, and, more recently, cryo-electron microscopy.
     \item
     \paratitle{PIR}~\cite{wu2004pirsf}.The Protein Information Resource(PIR) is a popular protein sequence database that provides information on functionally annotated protein sequences. 
PIR maintains three databases, the Protein Sequence Database (PSD), the Non-redundant Reference (NREF) sequence database, and the integrated Protein Classification (iProClass) database, which contains annotated protein sequences, classification information, and protein family, function, and structure information.

    \item \paratitle{AlphaFoldDB} ~\cite{jumper2021highly, 10.1093/nar/gkab1061}
    AlphaFold Database comprises protein structure predictions generated by DeepMind's AlphaFold system, offering accurate predictions of protein 3D structures. While these structures are not experimentally determined, they offer high-confidence models that can be incredibly useful, especially for proteins whose structures have not yet been solved experimentally.

\end{itemize}

\paratitle{Benchmarks:} \vspace{-0.5em}
\begin{itemize}
\item \paratitle{CASP} \cite{kryshtafovych2021critical}.
The Critical Assessment of Structure Prediction (CASP) is an influential initiative in the field of protein structure prediction. The goal of CASP is to advance the methods used in predicting protein structures from their amino acid sequences. CASP is essentially a series of competitions, which provides a unified benchmark for evaluating the performance of different methods in protein structure prediction.

\item \paratitle{EC}~\cite{10.1093/nar/gkn582}.
The Enzyme Commission (EC) dataset is a standardized dataset used to describe the classification and functions of enzymes. Enzymes are protein molecules that are widely present in living organisms and play a crucial role in catalyzing chemical reactions.
The EC dataset contains a large number of enzyme records, each with a unique EC number and corresponding description. These descriptions include the enzyme's name, reaction type, catalytic substrates, and products.

\item \paratitle{GO}~\cite{ashburner2000gene}.
The Gene Ontology (GO) dataset is a widely used gene annotation and functional classification system for describing the biological functions of genes and proteins. It provides a standardized framework for representing gene and gene product attributes across species and databases, helping researchers understand their functions in terms of cellular processes, molecular functions, and cellular components. 

\item \paratitle{CATH}~\cite{orengo1997cath}. The acronym CATH stands for the four main levels at which it classifies protein domains: Class, Architecture, Topology, and Homologous superfamily. 
CATH is widely used for protein structure classification and annotation. It offers a comprehensive framework for systematically classifying protein structures while providing crucial insights into their function and evolutionary aspects.

\item \paratitle{HIPPIE}~\cite{schaefer2012hippie}. The HIPPIE database is a comprehensive database that encompasses information on protein-protein interactions in humans. The data originates from diverse sources, including experimental techniques such as yeast two-hybrid assays, mass spectrometry analysis, and documented interaction relationships reported in the scientific literature. These datasets are meticulously harmonized and standardized to provide precise and dependable insights into protein-protein interactions. This invaluable resource finds extensive applications in functional annotation, systems biology research, drug target discovery, and beyond.

\item \paratitle{SCOP}~\cite{lo2000scop}.
SCOP (Structural Classification of Proteins) is a database used for protein structure classification. It helps researchers understand the structure,  function,  and evolutionary relationships of proteins by assigning protein structures to different levels and categories. The SCOP database employs a hierarchical classification system consisting of four levels: Class,  Fold,  Superfamily,  and Family. Each level has specific definitions and criteria to group protein structures into their respective categories.

\item \paratitle{ProteinGym}~\cite{pmlr-v162-notin22a}.
The ProteinGym platform comprises a comprehensive collection of Deep Mutational Scanning (DMS) assays meticulously curated to evaluate mutation effect predictors in determining the fitness of mutated proteins. It encompasses two distinct benchmarks: a substitution benchmark encompassing experimental characterization of 1.5 million missense variants across 87 DMS assays, and an indel benchmark comprising analysis of 300,000 mutants across 7 DMS assays.

\item \paratitle{FLIP}~\cite{dallago2021flip}.  
FLIP (Functional Learning for Protein Engineering) is a benchmark designed to encourage rapid scoring of representation learning for protein engineering. Currently, FLIP covers experimental data on adenovirus stability for gene therapy, stability of protein domain B1, immunoglobulin binding, and thermal stability of multiple protein families.

\item \paratitle{PEER}~\cite{xu2022peer}.
The PEER benchmark is a comprehensive and multi-task benchmark for protein sequence understanding, encompassing 17 tasks across five categories:  protein function, localization, structure predictions, protein-protein interaction prediction and protein-ligand interaction prediction. 
This benchmark evaluates diverse sequence-based methodologies for each task, including conventional feature engineering approaches, various sequence encoding methods, as well as large-scale pre-trained protein language models under both single-task learning and multi-task learning settings.

\item \paratitle{TAPE}~\cite{tape2019}.
The Tasks Assessing Protein Embeddings (TAPE) dataset is a collection of five biologically relevant semi-supervised learning tasks spanning different domains of protein biology, including secondary structure prediction, contact prediction, remote homology prediction, stability prediction, and fluorescence prediction.
TAPE divides the tasks into specific training, validation, and testing sets to ensure the evaluation of each task's biological relevance and generalizability to real-life scenarios. TAPE serves as a benchmark for a range of semi-supervised protein representation learning methods, encompassing recent advances as well as established sequence-based learning techniques.
\item \paratitle{Reactome}~\cite{jassal2020reactome}. Reactome is a database focused on human biomolecular reactions and pathways. It provides detailed networks of biochemical reactions, including proteins, small molecules, gene expression events, and their associated biological processes.  Reactome can be used to validate the model's predictions about the roles of proteins in biological processes.
\item \paratitle{STRING}~\cite{szklarczyk2023string}. STRING (Search Tool for the Retrieval of Interacting Genes/Proteins) is a comprehensive protein-protein interaction network database. It contains a wealth of information about known and predicted direct (physical) and indirect (functional) interactions. Through STRING, researchers can explore whether the proteins are in known interaction networks, or if they participate in unknown yet biologically plausible new interactions, which is crucial for understanding protein functions and complex biological pathways.
\item \paratitle{BioGRID}~\cite{oughtred2019biogrid}. BioGRID (Biological General Repository for Interaction Datasets) is an online database that specializes in recording and curating protein-protein interactions, genetic interactions, chemical associations, and post-translational modifications from scientific literature. BioGRID serves as a critical tool for validating the predictive capabilities of these models regarding protein interactions and modifications. By leveraging the extensive interaction data in BioGRID, researchers can assess the accuracy of the model’s predictions about how proteins interact with each other and their roles in cellular processes. 
\item \paratitle{InterPro}~\cite{paysan2023interpro}. InterPro is a comprehensive resource that offers functional analysis of protein sequences by classifying them into families and predicting the presence of domains and important sites. It integrates diverse information about protein families, domains, and functional sites from multiple reference databases. LLMs can utilize InterPro to annotate predicted protein sequences with potential functional domains, sites, and family memberships. Furthermore, InterPro's comprehensive collection of protein signatures aids in validating and refining the predictions made by protein sequence models, ensuring a higher level of accuracy and reliability in protein function prediction and analysis.
         \item
         \paratitle{DIP}~\cite{xenarios2002dip}. The Database of Interacting Proteins(DIP) is a database that contains protein-protein interaction information that has been compiled through both manual curations and computational methods.
It is useful for understanding protein functions, and their relationships with other proteins. It can also be used to study the properties of networks of interacting proteins, evaluate predictions of protein-protein interactions, and explore the evolution of these interactions.
         \item
         \paratitle{PROSITE}~\cite{sigrist2012new}. The PROSITE is a collection of signatures that identify patterns or profiles in proteins, which can provide information on their biological functions. 
The signatures in the database are linked to annotation documents that provide information on the protein family or domain detected, including its name, function, 3D structure, and references.

\end{itemize}

\subsection{Evaluation}\label{sec:prot-LLM-eval}
In the realm of protein computational models, Prot-LLMs are primarily evaluated based on their capabilities in three critical areas: protein function prediction, protein sequence generation, and protein structure prediction.
\paratitle{Protein Structure Prediction.} 
It is a crucial area within the fields of bioinformatics and computational biology. It involves the prediction of the three-dimensional structure of proteins given an input sequence, which includes determining the atomic coordinates and the topological relationships between atoms. Predicting protein structures is of significant importance for understanding protein function, drug design, and biomedical research.  This task includes tertiary and quaternary structure predictions. The former is critical for understanding the functional mechanisms of a protein, and the latter plays a vital role in understanding the functionality of multi-subunit complexes like hemoglobin.
The models for protein structure prediction often rely on encoder-based Prot-LLMs (i.e., ESM series~\cite{rives2019biological, meier2021language, lin2022language}) to extract the sequence information, followed by structure prediction modules.

\paratitle{Protein Function Prediction.} 
This task aims to determine the biological function of proteins, including how they operate within an organism and their interactions with other biomolecules. The prediction of protein function encompasses a multitude of subtasks, enumerated below.

\begin{itemize}
\item \paratitle{Protein Classification}. This refers to the process of categorizing and grouping proteins based on their structure, function, sequence similarity, or other features. 
\item \paratitle{Protein-Protein Interaction Prediction}. This refers to the focus on identifying and predicting interactions between different proteins. Proteins often interact with each other to perform various biological functions such as signal transduction, enzyme catalysis, and assembly of protein complexes. 
\item \paratitle{Protein-Ligand Interaction Prediction}. This refers to the process of predicting the binding interactions between a protein and a small molecule ligand. 
\item \paratitle{Subcellular Localization Prediction}. This refers to the prediction of the subcellular location of a protein within a cell. The subcellular localization of a protein is crucial for its function and interactions. 
\item \paratitle{Binary Localization Prediction}. This refers to the process of predicting the subcellular localization of a protein into one of two possible categories.
\item \paratitle{Remote Homology Detection}. This refers to the process of identifying distant relationships between protein sequences. In the protein world, homology refers to the shared ancestry or evolutionary relationship between two or more proteins.
\item \paratitle{Fluorescence Landscape Prediction}. This refers to the prediction of protein sequences to evaluate their fluorescence spectral characteristics under different environmental conditions. 
\item \paratitle{Stability Prediction}. This refers to the process of predicting the stability of a protein to assess its stability under specific conditions. Protein stability represents its ability to maintain the correct folding state within a biological organism or specific environment. 
\item \paratitle{$\beta$-Lactamase Activity Prediction}. This refers to the process of predicting the activity of a protein as a $\beta$-lactamase enzyme, which is capable of hydrolyzing $\beta$-lactam bonds and is commonly involved in antibiotic degradation and the development of antibiotic resistance.
\item \paratitle{Solubility Prediction}. This refers to the process of predicting the solubility of a compound in a solution. Solubility represents the maximum amount of a compound that can dissolve in a solvent at a specific temperature and pressure. 
\item \paratitle{Mutation Effect Prediction}. It focuses on understanding how genetic mutations can affect protein function, a critical aspect due to their potential to induce diseases or modify biological processes. Prot-LLMs aim to identify potential mutation sites and predict the functional consequences of these mutations. 

\end{itemize}

\paratitle{Protein Sequence Generation.}
It refers to the computational process of creating novel amino acid sequences that could potentially form functional proteins, which can be used for various applications such as drug design, enzyme engineering, and fundamental biological research. The task of protein sequence generation can be roughly classified into two categories:
\begin{itemize}
    \item \paratitle{De Novo Protein Design}. It involves creating entirely new protein sequences that are not based on existing proteins. This process relies heavily on computational methods to design proteins that can perform desired functions, such as binding to a particular target. The autoregressive generative models (e.g., ProGen series) are often employed for the tasks of protein sequence generation. 
    \item \paratitle{Protein Sequence Optimization}. It focuses on modifying existing protein sequences to enhance or alter their functions, employing techniques such as directed evolution. Directed evolution is particularly useful for improving the properties of existing proteins, such as increasing their stability, altering their substrate specificity, or enhancing their catalytic efficiency.
\end{itemize}

\paratitle{Evaluation metric}.
The evaluation of protein structure prediction models typically includes Root Mean Square Deviation (RMSD), Global Distance Test (GDT) \cite{zhang2007scoring},  Template Modeling Score (TM-Score)~\cite{zhang2007scoring}, and Local Distance Difference Test (LDDT) \cite{mariani2013lddt}. For more details on evaluating the predicted protein structure, readers can refer to \cite{kryshtafovych2021critical}.
For protein function prediction, the standard metrics in machine learning (e.g., accuracy for classification tasks and correlation coefficient for regression tasks) can be employed to quantitatively assess the performance of these Prot-LLMs. 
To evaluate the performance of protein sequence generation, one can use the following metrics: the model perplexity, novelty, similarity, and diversity of generated protein sequences, as well as the condition consistency if the generation is conditioned on certain inputs (e.g., protein backbone). We list them as follows.
\begin{itemize}
\item 
\textbf{Perplexity (PPL)}. In information theory, perplexity is a measure of uncertainty in the value of a sample from a discrete probability distribution. The larger the perplexity, the less likely it is that a language model can generate the protein sequence. The perplexity of a heldout test set reflects the capability of model to capture the distribution of natural sequences. The formula for PPL is defined as follows:
\begin{equation}
    \text{PPL}(p) = 2^{H(p)} = 2^{-\sum_x p(x)\log_2 p(x)} = \Pi_x p(x)^{-p(x)},
\end{equation}
where $H(p)$ is the entropy of the distribution $p$, and $x$ ranges over the protein sequence. 

\item 
\textbf{Novelty}. Novelty refers to the uniqueness or novelty of a protein sequence in comparison to known proteins, which can be reflected by the highest identity between each generated sequence and natural proteins. The calculation involves aligning the amino acid sequences and counting the number of positions in which the amino acids are identical, defined as
\begin{equation}
    \text{Identity}(\%) = (\frac{N_{\text{ident}}}{N_{\text{total}}}) \times 100,
\end{equation}
where $N_{\text{ident}}$ represents the number of identical positions and $N_{\text{total}}$ represents the total number of aligned positions.
A protein with a low percentage identity within all natural proteins implies a high degree of novelty.

\item 
\textbf{Fréchet Protein Distance (FPD) \cite{jiang2008protein}}. This metric quantifies the similarity between a set of generated proteins ($G$) and a reference set ($R$). Previous research has demonstrated that protein language model embeddings inherently encapsulate information related to both structure and function, suggesting that the FPD metric provides a comprehensive measure of coverage across the distribution of structural and functional properties. The FPD is mathematically defined as follows:
\begin{equation}
    \text{FPD} = ||\mu_G - \mu_R||^2 + \text{Tr}(\Sigma_G+\Sigma_R - 2\sqrt{\Sigma_G\Sigma_R}),
\end{equation}
where, given the embedding space feature vectors for the reference and generated distributions, $\mu$ represents the feature-wise mean for each set of sequences, $\Sigma$ denotes the representative covariance matrix, and $\text{Tr}$ refers to the trace linear algebra operation, defined as the sum of elements along the main diagonal of a square matrix.

\item 
\textbf{Diversity}. 
Sequence diversity assessment involves analyzing the variety of the generated proteins. Tools like BLAST can be used to compare these sequences against known protein databases to identify novel sequences. Metrics such as sequence similarity, percentage of unique sequences, and alignment scores provide a quantitative measure of diversity.

\item 
\textbf{Foldability}. This metric measures the foldability of a generated protein sequence by assessing the average per-residue confidence score, denoted as pLDDT, across the entire protein sequence. pLDDT serves as an indicator of the structural prediction model's confidence in its predictions for individual residues. Lower pLDDT scores are frequently associated with intrinsically disordered regions (IDRs) within proteins, which lack a well-defined three-dimensional structure and thus exhibit inherent unfoldability.

\item 
\textbf{Recovery}. This metric refers to the success or accuracy in predicting the correct amino acid sequence that corresponds to a given 3D structure (i.e., Structure-based conditional generation). A high recovery rate indicates that the designed sequences are likely to fold the desired structures.
\end{itemize}

\subsection{Summary}
\label{sec:prot-LLM-summary}
In this section, we have thoroughly examined the landscape of Sci-LLMs specifically designed for protein languages, namely Prot-LLMs. Our investigation encompasses a comprehensive analysis of their model architectures and capabilities, with a focus on how these models interpret and process protein languages. Additionally, we have curated an extensive collection of datasets, vital for training and benchmarking Prot-LLMs. These datasets are instrumental in providing a rich source of protein-related information, encompassing various aspects like sequence homology, structural configurations, and functional annotations. Finally, we present the evaluation method of Prot-LLMs, which encompasses several downstream tasks that these models are expected to perform, such as functional classification, interaction prediction, and protein sequence design. These tasks are crucial in assessing the models' practical applicability in real-world biological and pharmaceutical scenarios.

	\clearpage
	
	\section{Genomic Large Language Models} \label{sec-genome}
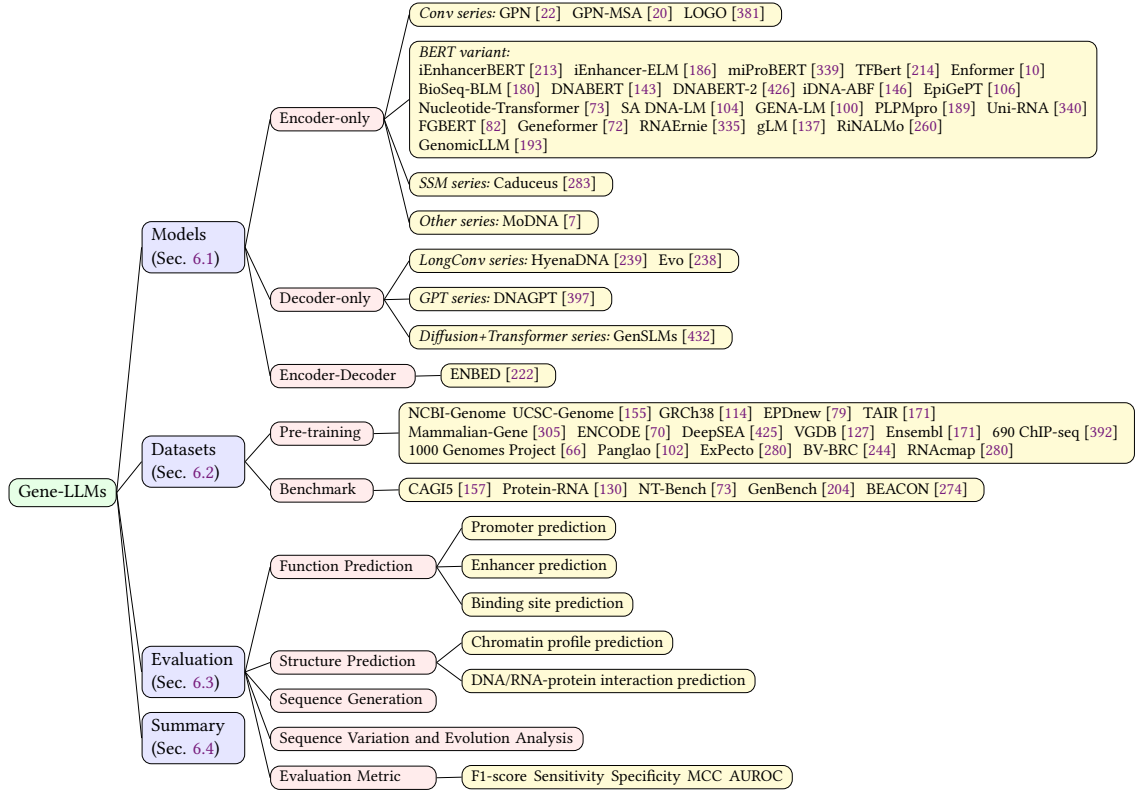
\begin{figure*}[h]
\centering
\resizebox{\textwidth}{!}{
\begin{forest}
  for tree={
  grow=east,
  reversed=true,
  anchor=base west,
  parent anchor=east,
  child anchor=west,
  base=left,
  font=\small,
  rectangle,
  draw,
  rounded corners,align=left,
  minimum width=2.5em,
  inner xsep=4pt,
  inner ysep=1pt,
  },
  where level=1{text width=5em,fill=blue!10}{},
  where level=2{text width=5em,font=\footnotesize,fill=pink!30}{},
  where level=3{font=\footnotesize,yshift=0.26pt,fill=yellow!20}{},
  [Gene-LLMs, fill=green!10
        [Models\\(Sec.~\ref{sec:gene-LLM-overview}),text width=4em
            [Encoder-only, text width=4.5em  
              [\emph{Conv series:} 
              GPN~\cite{benegas2023dna} \,
              GPN-MSA~\cite{benegas2023gpn} \,
              LOGO~\cite{yang2021logo} \,
              ]
              [\emph{BERT variant:} \\
              iEnhancerBERT~\cite{luo2022ienhancer} \,
              iEnhancer-ELM~\cite{li2023ienhancer} \,
              miProBERT~\cite{wang2023miprobert} \,
              TFBert~\cite{luo2023improving} \, 
              Enformer~\cite{avsec2021effective} \,\\
              BioSeq-BLM~\cite{li2021bioseq} \,
              DNABERT~\cite{ji2021dnabert} \,
              DNABERT-2~\cite{zhou2023dnabert}\, 
              iDNA-ABF~\cite{jin2022idna} \, 
              EpiGePT~\cite{gao2023epigept} \,\\
              Nucleotide-Transformer~\cite{dalla2023nucleotide} \,
              SA DNA-LM~\cite{gankin2023species} \,
              GENA-LM~\cite{fishman2023gena} \,
              PLPMpro~\cite{li2023plpmpro} \, 
              Uni-RNA~\cite{wang2023uni} \\
              FGBERT~\cite{duan2024fgbert} \, 
              Geneformer~\cite{cui2022geneformer} \,
              RNAErnie~\cite{wang2024multi} \, 
              gLM~\cite{hwang2024genomic} \,
              RiNALMo~\cite{penic2024rinalmo} \\
              GenomicLLM~\cite{liu2024exploring}
              ]
                [\emph{SSM series:}
                 Caduceus~\cite{schiff2024caduceus} \,
              ]      
              [\emph{Other series:}
                 MoDNA~\cite{an2022modna} \,
              ]
            ]
            [Decoder-only, text width=4.5em
                  [\emph{LongConv series:} 
                    HyenaDNA~\cite{nguyen2023hyenadna} \,
                    Evo~\cite{nguyen2024sequence} \, 
                  ]
                  [\emph{GPT series:} 
                    DNAGPT~\cite{zhang2023dnagpt} \,
                  ]
                  [\emph{Diffusion+Transformer series:} 
                    GenSLMs~\cite{zvyagin2022genslms} \,
                  ]   
            ]  
            [Encoder-Decoder, text width=6em
              [
              ENBED~\cite{malusare2023understanding} \,
              ]
            ]  
        ]
        [Datasets\\
        (Sec.~\ref{sec:gene-LLM-dataset}),text width=4em
            [Pre-training, text width=4em
              [NCBI-Genome\,
              UCSC-Genome ~\cite{karolchik2003ucsc}\,
              GRCh38 ~\cite{guo2017improvements} \, 
              EPDnew~\cite{dreos2013epd} \, 
              TAIR~\cite{lamesch2012arabidopsis}\, \\
              Mammalian-Gene~\cite{strausberg1999mammalian} \,
              ENCODE~\cite{encode2012integrated} \,
              DeepSEA~\cite{zhou2015predicting} \, 
              VGDB~\cite{hiscock2000viral} \,
              Ensembl~\cite{lamesch2012arabidopsis} \,
              690 ChIP-seq~\cite{zeng2016convolutional} \, \\
              1000 Genomes Project~\cite{10002015global} \,
              Panglao~\cite{franzen2019panglaodb} \,
              ExPecto~\cite{rusk2018sequence} \,
              BV-BRC~\cite{olson2023introducing} \, 
              RNAcmap~\cite{rusk2018sequence} \, 
              ]
            ]
            [Benchmark, text width=4em
              [CAGI5~\cite{katsonis2019cagi5} \, 
              Protein-RNA~\cite{horlacher2023systematic} \, 
              NT-Bench~\cite{dalla2023nucleotide} \,
              GenBench~\cite{liu2024genbench} \,
              BEACON~\cite{ren2024beacon} \,
              ]
            ]
        ]
        [Evaluation\\(Sec.~\ref{sec:gene-LLM-eval}),text width=4em
            [Function Prediction, text width=7em
                [Promoter prediction]
                [Enhancer prediction]
                [Binding site prediction]
            ]
            [Structure Prediction, text width=7em
                [Chromatin profile prediction]
                [DNA/RNA-protein interaction prediction]
            ]
            [Sequence Generation, text width=7em
            ]
            [Sequence Variation and Evolution Analysis, text width=14em
            ]  
            [Evaluation Metric, text width=7em
            [F1-score\, Sensitivity\, Specificity\, MCC\, AUROC]
            ] 
        ]
        [Summary\\
        (Sec.~\ref{sec:gene-LLM-summary}),text width=4em
        ]    
    ]
\end{forest}
}
\caption{Chapter overview of Gene-LLMs.}
\label{fig:gene-LLM-overview}
\end{figure*}

In the field of computational biology, genomic data exhibits resemblances to sequence-based information observed in natural language, enabling the application of large language models for analyzing genomic sequences. In this section, we provide a review of LLMs tailored for genomic language (Gene-LLMs), including insights into their model architectures, datasets, and evaluation. The overview of this section is shown in Figure~\ref{fig:gene-LLM-overview}.

\subsection{Models} \label{sec:gene-LLM-overview}
Genomic LLMs, based on the Transformer architecture, proficiently model DNA or RNA sequence information by capturing long-range dependencies to accomplish prediction and generation tasks. 
By employing self-supervised learning techniques on the genomic sequences, Gene-LLMs gradually acquire an understanding of the genome. Once fine-tuned, or through contextual learning, they can be extensively utilized for downstream tasks, thereby improving accuracy and reducing the need for manual intervention.
Here we categorize Gene-LLMs into three distinct types based on their specific architectures: encoder-only, decoder-only, and encoder-decoder models, as shown in Table \ref{tab:Gene-LLMs}.

\begin{table*}[!htp]
    \centering
    \caption{Summary of Gene-LLMs}
    \renewcommand{\arraystretch}{1.2}
    \footnotesize
    \begin{tabular}{clccccccc}
         \toprule
         \multirow{33}{*}{\makecell{Encoder-\\only}}
         &Model & Time & \#Parameters & Base Model & \makecell{Pretraining \\Dataset} &Capability & Data Type & \makecell{Open-\\Source}\\
         \midrule
         &\href{https://github.com/jerryji1993/DNABERT}{DNABERT}~\cite{ji2021dnabert} & 2021.02 & 110M &  BERT & GRCh38 & 
         \makecell{Func. pred.}
         & DNA & \checkmark\\
         &\href{https://github.com/google-deepmind/deepmind-research/tree/master/enformer}{Enformer}~\cite{avsec2021effective} & 2021.10 & 240M & BERT &  \makecell{GRCh38} & 
         \makecell{Func. pred.}
         & DNA & \checkmark\\
         &MoDNA~\cite{an2022modna} & 2022.08 & - & BERT & GRCh38 & 
         \makecell{Func. pred.}
         & DNA & $\times$\\
         &\href{https://github.com/songlab-cal/gpn}{GPN}~\cite{benegas2023dna} & 2022.08 & - & Conv & NCBI-Genome
         & 
         Variant pred.
         & DNA & \checkmark\\
         &iEnhancer-BERT~\cite{luo2022ienhancer} & 2022.08 & 110M &  BERT & - & 
         \makecell{Enhancer. pred.}
         & DNA & $\times$\\
        &\href{https://github.com/FakeEnd/iDNA_ABF}{iDNA-ABF}~\cite{jin2022idna} & 2022.10 & 110M & BERT &  
         -         & 
         \makecell{Func. pred.}
         & DNA & \checkmark\\
         &\href{https://github.com/chen-bioinfo/iEnhancer-ELM}{iEnhancer-ELM}~\cite{li2023ienhancer} & 2022.12 & 110M & BERT & -         & 
         Enhancer. pred.
         & DNA & \checkmark\\
         &\href{https://github.com/DennisGankin/species-aware-DNA-LM}{SA DNA-LM}~\cite{gankin2023species} & 2023.01 & - &  BERT & 
         Ensembl
         & 
         Expression. pred.
         & DNA & \checkmark\\
         & \href{https://github.com/instadeepai/nucleotide-transformer}{\makecell[l]{Nucleotide-\\Transformer~\cite{dalla2023nucleotide}}} & 2023.01 & 50M-2.5B & BERT & \makecell{1000 Genomes\\ Project, etc.}
         & 
         \makecell{Func. pred.}
         & DNA & \checkmark\\
         & \href{https://github.com/biomed-AI/SpliceBERT}{SpliceBERT}~\cite{chen2023self} & 2023.02 & 19.4M & BERT & UCSC 
         & 
         \makecell{Splicing pred.}
         & RNA & \checkmark\\
         & miProBERT~\cite{wang2023miprobert} & 2023.03 & 110M & BERT & EPDnew & 
         Promoter pred.
         & RNA & $\times$\\
         & \href{https://github.com/yikunpku/RNA-MSM}{RNA-MSM}~\cite{zhang2023multiple} & 2023.03 & - & BERT & RNAcmap & 
         \makecell{Structure pred.}
         & RNA & \checkmark\\
        & \href{https://github.com/lhy0322/TFBert}{TFBert}~\cite{luo2023improving} & 2023.05 & - & BERT & 690 ChIP-seq  & 
         Binding pred.
         & DNA & \checkmark\\
         & \href{https://github.com/servehubco/Geneformer}{Geneformer}~\cite{cui2022geneformer} & 2023.05 & 41M-158M & Transformer & NCBI, etc.  & 
         Func. pred.
         & DNA & \checkmark\\
         & \href{https://github.com/zhangying-njust/MRM-BERT}{MRM-BERT}~\cite{zhang2023prediction} & 2023.06 & 110M & BERT & - & 
         \makecell{Modif. pred.}
         & RNA & \checkmark\\
         & \href{https://github.com/AIRI-Institute/GENA_LM}{GENA-LM}~\cite{fishman2023gena} & 2023.06 & 110M-360M & BERT & 
         GRCh38         & 
         \makecell{Func. pred. \\ and profiling}
         & DNA & \checkmark\\
         & DNABERT-2~\cite{zhou2023dnabert} & 2023.06 & 117M &  BERT & GRCh38 & 
         \makecell{Func. pred.}
         & DNA & \checkmark\\
         &PLPMpro~\cite{li2023plpmpro} & 2023.07 & - & BERT & EPDnew & 
         Promoter pred.
         & DNA & $\times$ \\
         &\href{https://github.com/ZjGaothu/EpiGePT}{EpiGePT}~\cite{gao2023epigept} & 2023.07 & - & BERT & ENCODE
         & 
         \makecell{Interpret genome}
         & DNA & \checkmark\\
         &Uni-RNA~\cite{wang2023uni} & 2023.07 & - & BERT & NCBI-Genome
         & 
         Func. pred.
         & DNA & $\times$\\
         &GPN-MSA~\cite{benegas2023gpn} & 2023.10 & 86M & BERT & -
         & 
         \makecell{Variant class. and pred.}
         & DNA & $\times$ \\
         &FGBERT~\cite{benegas2023gpn} & 2024.02 & 954M & BERT & -
         & 
         \makecell{seq-func. relation}
         & DNA & $\times$ \\
         &\href{https://github.com/y-hwang/gLM}{gLM}~\cite{hwang2024genomic} & 2024.04 & 1B & BERT & -
         & 
         \makecell{Func. pred.}
         & RNA & \checkmark \\
         &\href{https://github.com/kuleshov-group/caduceus}{Caduceus}~\cite{schiff2024caduceus} & 2024.04 & 1B & SSM & -
         & 
         \makecell{Func. pred.}
         & RNA & \checkmark \\
         & \href{https://github.com/CatIIIIIIII/RNAErnie}{RNAErnie}~\cite{wang2024multi} & 2024.05 & 105M & BERT & -
         & 
         \makecell{Func. pred.}
         & RNA & \checkmark \\
         & \href{https://github.com/lbcb-sci/RiNALMo}{RiNALMo}~\cite{penic2024rinalmo} & 2024.05 & 650M & BERT & -
         & 
         \makecell{Func. pred.}
         & RNA & \checkmark \\

        \midrule
        \multirow{5}{*}{\makecell{Decoder-\\only}}
         & \href{https://github.com/ramanathanlab/genslm}{GenSLMs}~\cite{zvyagin2022genslms} & 2022.10 & 25M-25B & \makecell{Diffusion+\\GPT} & BV-BRC & 
         \makecell{Evolution \\analysis}
         & RNA & \checkmark\\
         &\href{https://github.com/HazyResearch/hyena-dna}{HyenaDNA}~\cite{nguyen2023hyenadna} & 2023.06 & 0.44M-6.6M & LongConv & GRCh38 & 
         \makecell{Func. pred. \\ and class.}
         & DNA & \checkmark\\
         &\href{https://github.com/Xianjun-Yang/DNA-GPT}{DNAGPT}~\cite{zhang2023dnagpt} & 2023.08 & 100M-3B & GPT & GRCh38 & \makecell{Func. pred. \\ and generation}
         & DNA& \checkmark\\
         &\href{https://github.com/evo-design/evo}{Evo}~\cite{nguyen2024sequence} & 2024.02 & 7B & LongConv & OpenGenome
         & 
         \makecell{Pred. \\ mol interactions.}
         & DNA & $\times$ \\
         \midrule
         \makecell{Encoder-\\Decoder}
         &ENBED~\cite{malusare2023understanding} & 2023.11 & 580M-1.2B & Transformer &  NCBI-Genome & \makecell{Func. pred. \\ and analysis} & DNA& $\times$\\
         \bottomrule
    \end{tabular}
    \label{tab:Gene-LLMs}
\end{table*}

\paratitle{Encoder-only Models.} 
In the realm of encoder-only architecture, significant models such as SpliceBERT~\cite{zheng2019predicting}, DNABERT~\cite{ji2021dnabert}, DNABERT-2~\cite{zhou2023dnabert}, iEnhancer-BERT~\cite{luo2022ienhancer}, scBERT~\cite{yang2022scbert}, BioSeq-BLM~\cite{li2021bioseq}, PLPMpro~\cite{li2023plpmpro}, GPN~\cite{benegas2022dna}, SA DNA-LM~\cite{gankin2023species}, RNA-MSM~\cite{zhang2023multiple}, miProBERT~\cite{wang2023miprobert}, TFBert~\cite{luo2023improving}, MRM-BERT~\cite{zhang2023prediction}, GPN-MSA~\cite{benegas2023gpn}, iDNA-ABF~\cite{jin2022idna} and FGBERT~\cite{duan2024fgbert} are noteworthy attempts employing the Transformer encoder within genomics. These models utilize a mask training mechanism where partial gene sequences are masked, prompting the model to predict and complete them, progressively learning inherent patterns within gene sequences. They demonstrate versatility in fine-tuning for various gene sequence-related tasks like promoter prediction, transcription factor binding site prediction, and single nucleotide variability, achieving commendable performance.
MoDNA~\cite{an2022modna}  employs a Bert-like encoder-only architecture, introducing a unique training paradigm with a stacked Generator-Discriminator structure. During training, the Generator completes masked sections based on sequence and motif information, while the Discriminator assesses the completed sequences. This dual-layer architecture enables the simultaneous integration of sequence features and motif information, facilitating motif-oriented learning. 
GENA-LM~\cite{fishman2023gena} introduces a suite of encoder-based foundational DNA language models designed for genomics research. These models, unique in their ability to handle long sequences of up to 36,000 base pairs, address the challenge of decoding genomic sequences, which requires understanding extensive contextual information across numerous nucleotides. The integration of a newly-developed recurrent memory mechanism allows these models to process even larger DNA segments. 
Nucleotide-Transformer~\cite{dalla2023nucleotide} is a foundation encoder-only model pre-trained on DNA sequences. It integrates data from 3,202 diverse human genomes and 850 genomes from various species. The study focuses on bridging the gap in predicting molecular phenotypes from DNA sequences, a task often limited by the scarcity of annotated data and difficulties in transferring learnings between different prediction tasks. These transformer models, with parameters ranging from 50M to 2.5B, provide transferable, context-specific representations of nucleotide sequences, enabling more accurate molecular phenotype predictions, even in scenarios with limited data. 
Remarkably, these models learn to focus on key genomic elements, such as enhancers that regulate gene expression, without supervision, and can help prioritize functional genetic variants. 
EpiGePT~\cite{gao2023epigept} is a transformer-based language pretrained model designed for epigenomics. It predicts genome-wide epigenomic signals by considering transcriptional regulation mechanisms. EpiGePT uniquely incorporates context-specific activities of transcription factors, offering deeper insights into gene regulation compared to models trained only on DNA sequences. It shows exceptional performance in diverse epigenomic signal prediction tasks and can learn cell-type-specific long-range interactions through self-attention mechanisms. 
Uni-RNA~\cite{wang2023uni} is an encoder-only model that excels in predicting RNA structures and functions, outperforming previous methods in various tasks. It leverages a billion RNA sequences to extract hidden evolutionary and structural information, revolutionizing RNA research and facilitating drug development.
Geneformer~\cite{cui2022geneformer} is a context-aware, attention-based deep learning model developed for network biology applications with limited data. It is pretrained on a large-scale corpus of human single-cell transcriptomes, enabling it to make context-specific predictions. The model leverages transfer learning, allowing it to fine-tune and adapt to various downstream tasks, such as disease modeling and identification of therapeutic targets, with only a small amount of task-specific data.
As gene sequences often significantly exceed the length of natural language text, many models aim to overcome the quadratic time complexity of attention mechanisms. Models like Enformer~\cite{avsec2021effective} and LOGO~\cite{yang2021logo} employ a blend of convolutional downsampling to handle longer DNA sequences efficiently. Tools like BioSeq-BLM~\cite{li2021bioseq} integrate traditional analysis methods with language models. PLPMpro~\cite{li2023plpmpro} marks the advent of pre-training and fine-tuning in this domain, comparing language models' effects with traditional methods.


\paratitle{Decoder-only Models.} 
Decoder-only models have gained significant attention due to their potent generative capacity.
GenSLMs~\cite{zvyagin2022genslms} leverage genome-scale language models comprising multiple layers of attention-based decoders to elucidate the evolutionary dynamics of SARS-CoV-2. GenSLMs thus adeptly capture the evolutionary landscape of the SARS-CoV-2 genome. Pre-trained on prokaryotic gene sequences and further fine-tuned for SARS-CoV-2 genomes, GenSLM demonstrates precise and rapid identification of concerning variants.
Besides commonly used architectures like transformers, niche architectures also find applications in genomic research. For instance, 
Models like DNAGPT~\cite{zhang2023dnagpt} have introduced decoder-only architectures, akin to GPT, into the genomics domain by integrating specific symbols to guide different tasks, enabling zero-shot capabilities in species identification and regulatory factor prediction. HyenaDNA~\cite{nguyen2023hyenadna} incorporates a fast Fourier transform-accelerated recursive long-range convolution, a quadratic time complexity solution, enabling the handling of ultra-long DNA sequences. Employing convolution on the entire DNA sequence preserves single-nucleotide resolution, facilitating research into single nucleotide variability. Evo~\cite{nguyen2024sequence} , a pretrained genome LLM with a modified StripedHyena architecture, is also capable of interpreting the core biological "languages" of DNA, RNA, and proeins, allowing for both predictive analysis and generative design across molecular and genomic scales.

\paratitle{Encoder-Decoder Models.}
Encoder-decoder models in genomics, such as the Ensemble Nucleotide Byte-level Encoder-Decoder (ENBED)~\cite{malusare2023understanding}, represent a significant advancement in bioinformatics. These models combine the strengths of both encoder and decoder components to effectively analyze and interpret complex genomic sequences. The encoder part of the model is responsible for compressing and encoding the input genomic data into a meaningful representation, capturing the essential features and patterns within the DNA sequences. This encoded representation is then passed to the decoder part, which is tasked with generating or reconstructing sequences, making predictions, or performing other relevant bioinformatics tasks. 





\subsection{Datasets}  \label{sec:gene-LLM-dataset}

\begin{table*}[t]
    \centering
    \caption{Summary of datasets for Gene-LLMs} 
    \renewcommand{\arraystretch}{1.2}
    \footnotesize
    \begin{tabular}{lllll}
    \toprule
     &Dataset & Last updated  &Scale& Keywords\\ \midrule
     
     \multirow{16}{*}{Pre-training}&\href{http://mgc.nci.nih.gov}{Mammalian-Gene}~\cite{strausberg1999mammalian} &  1999.10  &-& Mammalian Genomic Data \\
     
     &\href{https://www.ncbi.nlm.nih.gov/assembly/GCF_000001405.26/}{GRCh38}~\cite{guo2017improvements} &  2013.12 &3.2G nucleotides& Genome Reference, Human, Annotations\\
     &\href{https://github.com/zhanglabtools/DNADataAugmentation}{690 ChIP-seq}~\cite{zeng2016convolutional} &  2016.06 &-& DNA-binding Protein Data \\
     
    & \href{http://deepsea.princeton.edu/job/analysis/create/}{DeepSEA}~\cite{zhou2015predicting} & 2017.04 &-&Chromatin Effects \\
        &\href{https://www.internationalgenome.org}{1000 Genomes Project}~\cite{10002015global} &  2017.10 &20.5T nucleotides& Human Genetic Variation \\
 
     &\href{https://epd.expasy.org/epd}{EPDnew}~\cite{dreos2013epd} &  2019.11 &187K promoters& Promoter Prediction \\
     &\href{https://panglaodb.se}{Panglao}~\cite{franzen2019panglaodb} &  2020.03 & 4M cells & scRNA-seq Data \\
     &\href{https://github.com/FunctionLab/ExPecto}{ExPecto}~\cite{rusk2018sequence} &  2020.12 &-& Gene Expression \\
     &\href{https://genome.ucsc.edu}{UCSC-Genome}~\cite{karolchik2003ucsc} & 2022.11  &-& Various Bio Data \\
     
     &\href{https://www.bv-brc.org}{BV-BRC}~\cite{olson2023introducing} &  2023.01 &-& Bacterial and Viral Pathogens \\
     &\href{https://asia.ensembl.org/index.html}{Ensembl}~\cite{lamesch2012arabidopsis}& 2023.02 &-& Various Bio Data\\
     &\href{https://github.com/jaswindersingh2/RNAcmap}{RNAcmap}~\cite{rusk2018sequence} &  2023.07 &-& RNA Evolutionary analysis \\
    
     &\href{https://www.encodeproject.org}{ENCODE}~\cite{encode2012integrated} &  2023.09 &-& Functional Genomics \\
     
     &\href{https://www.ncbi.nlm.nih.gov/genome/}{NCBI-Genome} &  2023.10 &-& Various Bio Data \\
      &\href{https://www.arabidopsis.org}{TAIR}~\cite{lamesch2012arabidopsis}&2023.12  &33602 genes& Arabidopsis Information\\
     &\href{https://www.ncbi.nlm.nih.gov/genome/viruses/}{VGDB}~\cite{hiscock2000viral} & 2023.12 &2847 genes& Viral Genome Sequence \\
     
     \midrule
     &\href{https://onlinelibrary.wiley.com/doi/full/10.1002/humu.23873}{CAGI5}~\cite{katsonis2019cagi5} &  2019.07 &-& Assessing Genetic and Genomic Predictions \\
     
     Benchmark&\href{https://github.com/brianjimenez/protein-RNA-benchmark}{Protein-RNA}~\cite{horlacher2023systematic}          &  2023.08    &-& Evaluating RBP-RNA Interaction Prediction Methods\\
     
     &\href{https://github.com/instadeepai/nucleotide-transformer}{NT-Bench}~\cite{dalla2023nucleotide} &  2023.09 &3.2G+20.5T+174G nucleotides& Evaluating Genomics Foundational Models \\

      &\href{https://github.com/jimmylihui/GenBench}{GenBench}~\cite{liu2024genbench} &  2024.06 & - & Evaluating Genomics Foundational Models \\

      &\href{https://github.com/terry-r123/RNABenchmark}{BEACON}~\cite{ren2024beacon} &  2024.06 & - & Evaluating RNA Models \\
    \bottomrule
    \end{tabular}
    \label{tab:gene-datasets}
\end{table*} 

Currently, the majority of Gene-LLMs are trained on human DNA data. To increase the amount of training data and enhance the model's generalization ability, genomes of organisms closely related to humans, such as chimpanzees or pigs, or even the genomes of all mammals are utilized for training. 
Here we list prevalent pre-training datasets and benchmarks for Gene-LLMs, as summarized in Table \ref{tab:gene-datasets}.

\paratitle{Pre-training Datasets}:
\begin{itemize} 

\item \paratitle{Mammalian-Gene}~\cite{eppig2017mouse}.
In Gene-LLM training, integrating mammalian genomic data alongside human data enhances the model's generalization. These databases provide a plethora of genomic sequences from various mammals, aiding in building a robust Gene-LLM capable of better understanding and interpreting genomic sequences across different species, thereby enriching genomic research.

\item \paratitle{GRCh38 (Human Genome)}~\cite{guo2017improvements}.
Human genome is frequently used as the training corpus for Gene-LLMs. Genome Reference Consortium Human Build 38 (GRCh38) is a substantial assembly of the human genome released after the previous major assembly. The genomic sequences, annotations, and metadata contained within GRCh38 provide a rich source of information that can be utilized to train LLMs to recognize various genomic features, understand genomic syntax, and make predictions related to gene expression, mutation impacts, and other genomic phenomena.

\item \paratitle{690 ChIP-seq}~\cite{zeng2016convolutional} The 690 ChIP-seq datasets are a comprehensive collection of DNA-binding protein data, encompassing 91 human cell types and 161 specific DNA-binding proteins under various conditions. The significance of these datasets lies in their diversity and volume, providing a rich foundation for understanding DNA-protein interactions. This variety and scale are crucial for training advanced models like GeneLLM, as they allow the model to learn from a wide range of scenarios, ensuring robustness and generalizability in predicting DNA-protein bindings. The inclusion of data from different cell types and conditions also aids in capturing the complexity of gene regulation, enhancing the model's ability to make accurate predictions in various biological contexts.

\item \paratitle{DeepSEA}~\cite{zhou2015predicting}.
DeepSEA compiles extensive chromatin profiles from ENCODE and Roadmap Epigenomics projects, comprising data on 690 transcription factor bindings, 125 DNase I sensitivity profiles, and 104 histone-mark profiles. 
It is a significant advance for researchers looking to prioritize functional variants, including eQTLs and disease-associated variants, using purely genomic sequence data, thus offering a robust resource for gene model training and functional genomic studies.

\item \paratitle{1000 Genomes Project}~\cite{10002015global}
The 1000 Genomes Project provided a comprehensive reference for human genetic variation. It sequenced over 2,500 individuals from 26 populations, using whole-genome sequencing, deep exome sequencing, and microarray genotyping. The study identified 88 million variants, including single nucleotide polymorphisms, insertions/deletions, and structural variants. This work expanded our understanding of genetic diversity, particularly in underrepresented populations, and has been instrumental in studies of human evolution, disease biology, and genetic research methodologies. The data serves as a dataset for human genetic studies and has greatly enhanced the catalog of genetic variation, particularly in South Asian and African populations.

\item \paratitle{EPDnew}~\cite{dreos2013epd}.
This is a collection of databases encompassing experimentally validated promoters for selected model organisms, with evidence derived from TSS-mapping from high-throughput experiments like CAGE and Oligo-capping. This data can be particularly useful for tasks related to promoter prediction, gene expression analysis, and other genomic sequence interpretation tasks within the realm of genomics and bioinformatics. 

\item \paratitle{Panglao}~\cite{franzen2019panglaodb}
PanglaoDB is a comprehensive web server database designed for the exploration of mouse and human single-cell RNA sequencing (scRNA-seq) data. This database stands out due to its user-friendly interface, allowing easy access and exploration of extensive scRNA-seq datasets. With over 1054 single-cell experiments, encompassing more than 4 million cells from a diverse range of tissues and organs, PanglaoDB presents an invaluable resource for biological research. 
The availability of pre-processed and quality-controlled scRNA-seq data allows for efficient training and validation of machine learning models aimed at gene expression analysis. The database's broad coverage of cell types and conditions provides a diverse training set that can help improve the accuracy and generalizability of predictive models in genomics.

\item \paratitle{ExPecto}~\cite{rusk2018sequence} The ExPecto dataset is designed for predicting tissue-specific gene expression from DNA sequences. 
This dataset was validated using known expression quantitative trait loci and massively parallel reporter assays, and it has been used to prioritize variants in genome-wide association studies. It facilitates the characterization of genetic variation potential within each gene and enables systematic investigation into tissue-specific effects of mutations on gene expression.

\item \paratitle{UCSC-Genome}~\cite{karolchik2003ucsc}.
The UCSC Genome Database is a comprehensive online tool for genomic research. It provides access to genomic sequences from various species, integrated with a plethora of aligned annotations. This browser facilitates rapid visualization and analysis of genomic data, supported by an extensive database and a suite of interactive tools. With over 108 species' genomes, it allows for in-depth comparative genomics and includes unique features like Assembly Hubs and Custom Tracks, enabling researchers to personalize data analysis. 

\item \paratitle{BV-BRC}~\cite{olson2023introducing}. The BV-BRC dataset is a comprehensive resource integrating genomic and other related data for bacterial and viral pathogens. It hosts a vast collection of genomic sequences, including over 600,000 bacterial genomes, 11,000 archaeal genomes, and 8.5 million viral genomes, which encompasses over 6 million SARS-CoV-2 genomes. This dataset is instrumental in comparing human pathogens with their non-pathogenic relatives, thereby aiding in understanding virulence and pathogenicity. The BV-BRC also includes metadata and non-genomic data like protein structures, enhancing research in host-pathogen interactions and responses to infections. The data is consistently annotated, and its extensive nature supports a wide range of comparative genomic analyses and epidemiological studies.

\item \paratitle{Ensembl}~\cite{martin2023ensembl}.
It offers centralized access to a wide range of data, including DNA sequences, gene models, annotations, and variation data for vertebrates and model organisms. This platform is instrumental for those studying genomics and serves as a repository for large-scale data suited for training complex models. Ensembl’s automated annotation and vast database make it an invaluable resource for machine learning and bioinformatics, providing the raw material for the development of predictive models and analysis tools.

\item \paratitle{RNAcmap}~\cite{rusk2018sequence}. The RNAcmap dataset focuses on RNA evolutionary coupling analysis. It employs a fully automatic pipeline, RNAcmap, which doesn't rely on pre-curated Rfam families. This pipeline uses BLAST-N for initial homolog searches and INFERNAL for building a covariance model based on predicted secondary structures. The dataset supports the generation of base-pairing and distance-based contact maps, using multiple sequence alignment from secondary searches. It demonstrates the less dependency on specific evolutionary coupling tools and more on the accuracy of secondary structure predictors. This dataset is beneficial for RNA structure prediction, offering a significant step in computational RNA analysis by enabling evolutionary coupling analysis for any RNA sequence.

\item \paratitle{ENCODE}~\cite{feingold2004encode}.
The Encyclopedia of DNA Elements (ENCODE) Project aimed at developing a comprehensive map of functional elements within the human genome, which includes genes among other functional genomic elements. This rich dataset offers a vast amount of functional genomics data, which encompasses a variety of experiments and annotations related to stem cell development, immune cells, mouse development, and other crucial biological processes. 

\item  \paratitle{NCBI-Genome}\footnote{\url{https://www.ncbi.nlm.nih.gov}}. 
The National Center for Biotechnology Information (NCBI) is a key platform offering comprehensive data for gene-related research and large model training. NCBI hosts a multitude of databases such as GenBank for DNA sequences, which is crucial for understanding genetic blueprints across various organisms. Their Gene database provides a vital repository for gene-specific information, linking genomic maps, sequences, expression, and function data, making it an invaluable resource for bioinformatics. 

\item \paratitle{TAIR}~\cite{lamesch2012arabidopsis}.
Arabidopsis Information Resource (TAIR) is a comprehensive database for the model plant Arabidopsis thaliana, offering a wealth of data for large-scale model training. It provides annotated genomic sequences, gene function and expression data, metabolic pathways, and gene expression patterns. The database integrates RNA-seq and proteomics data, offering a rich set of annotations for molecular function, cellular components, and biological process aspects of the GO terms. 

\item \paratitle{VGDB}~\cite{hiscock2000viral}.
Viral Genome DataBase is a substantial resource organizing the `sequence space' of viral genomes in a structured manner, aiming to serve as an annotated and curated database of complete viral genome sequences along with value-added derived data and data mining tools. Moreover, VGDB offers detailed gene information and predicted protein sequences from completely sequenced genomes of large viruses, encompassing DNA sequence data, protein sequence data, and other related annotations. 

\end{itemize}
\paratitle{Benchmarks}:
\begin{itemize}

\item \paratitle{CAGI5 Challenge Benchmark (CAGI5)}~\cite{cheng2019cagi}.
The CAGI5 Challenge is a comprehensive benchmark designed to rigorously assess computational methods in predicting a wide array of genetic and genomic outcomes. At the forefront is the prediction of effects of 17,500 single nucleotide variants and small indels in 11 disease-associated enhancers and 10 promoters using a saturation mutagenesis massively parallel reporter assay. Additionally, the challenge delves into the functional effects of nonsynonymous variants in proteins. Another significant aspect of the challenge is the prediction of the impact of variants on exon splicing. The challenge also extends to clinical genomes, where participants are tasked with matching patients' genome sequences to their clinical descriptions and predicting the causal pathogenic variants. This includes data from children with suspected genetic disorders and gene panel sequences associated with intellectual disability or Autism spectrum disorders. 

\item \paratitle{Protein-RNA Interaction Prediction Benchmark (Protein-RNA)}~\cite{horlacher2023systematic}.
While numerous machine-learning-based methods have been developed to predict RNA-binding protein (RBP) interactions, direct comparisons of their performance have been challenging due to the use of varied training and evaluation datasets across different RBP sets and CLIP-seq protocols. It compiled a set of 37 methods designed for \textit{in vivo} RBP-RNA interaction prediction. 
This benchmark systematically evaluates the performance of 11 RBP binding-site prediction methods using binding sites from three large CLIP-seq repositories, which include 313 unique CLIP-seq experiments across 203 RBPs. 

\item \paratitle{Nucleotide Transformer Benchmark (NT-Bench)}~\cite{dalla2023nucleotide}.
This is a comprehensive evaluation framework designed to assess the performance of genomics foundational models. This benchmark pits the Nucleotide Transformer models against other prominent genomics models, such as DNABERT, HyenaDNA (with both 1kb and 32kb context lengths), and Enformer. To ensure a level playing field, all models in this benchmark are subjected to fine-tuning and evaluation using a uniform protocol that spans 18 downstream tasks with the corresponding datasets (e.g., GRCh38, 1000 Genomes project, and the multispecies dataset from NCBI).

\item \paratitle{GenBench: A Benchmarking Suite for Systematic Evaluation of Genomic Foundation Models}~\cite{liu2024genbench}.
GenBench is a benchmarking suite developed to provide a systematic evaluation framework for Genomic Foundation Models (GFMs). It addresses the challenge of equitable assessment by standardizing experimental settings, model complexities, benchmark datasets, and reproducibility. This suite includes evaluations across short- and long-range genomic tasks, covering key areas such as Coding Region, Non-Coding Region, and Genome Structure. GenBench supports a variety of state-of-the-art models, including DNABERT, HyenaDNA, and Nucleotide Transformer, among others. The benchmark spans 43 datasets, offering a thorough analysis of model performance and the interaction between model architectures and dataset characteristics. Through these evaluations, insights are gained into the strengths of different model types, guiding future GFM development. 

\item \paratitle{BEACON: Benchmark for Comprehensive RNA Tasks and Language Models}~\cite{ren2024beacon}.
This benchmark provides a thorough evaluation framework for RNA-related tasks, addressing the gap in standardized benchmarks for RNA models. BEACON assesses various RNA foundational models and traditional approaches, such as CNNs, ResNets, and LSTMs, across 13 tasks spanning structural analysis, functional studies, and engineering applications. The benchmark includes datasets from multiple sources like bpRNA, icSHAPE, and GENCODE, covering RNA sequence-level and nucleotide-level analyses. Key findings highlight the effectiveness of single nucleotide tokenization and the Attention with Linear Biases (ALiBi) positional encoding in RNA language models. BEACON also introduces a robust baseline model, BEACON-B, that demonstrates strong performance with limited data and computational resources.

\end{itemize}

\subsection{Evaluation} \label{sec:gene-LLM-eval}
Gene-LLMs are primarily evaluated based on their capabilities in four critical areas: function prediction, structure prediction, sequence generation, and sequence variation and evolution analysis. 

\paratitle{Function Prediction.}
The prediction of gene function has long been a pivotal focus in genomics research. Traditionally, models were trained directly on specific sequences relevant to these tasks to acquire the requisite capabilities. The advent of LLMs has revolutionized this approach by facilitating pre-training on vast amounts of genomic data, followed by targeted fine-tuning on specific tasks, empowering these models to effectively discern and predict gene expression patterns and their regulatory elements with enhanced accuracy and broader context. Here we present three subtasks that are frequently used for function prediction.
\begin{itemize}
    \item \paratitle{Promoter Prediction}. 
    One crucial task in genomics research involves predicting the locations of promoters, specific DNA segments recognized and initiated for transcription by RNA polymerase. Promoters, typically found upstream of structural genes, contain essential sequences vital for RNA polymerase binding and transcription initiation. DNABERT~\cite{ji2021dnabert}, for instance, effectively utilized human TATA and non-TATA promoter regions to fine-tune downstream models, surpassing existing state-of-the-art models. 
   
    \item \paratitle{Enhancer Prediction}. 
    Enhancers, short DNA regions, play a pivotal role in enhancing the likelihood of gene transcription by providing binding sites for various molecules, facilitating the assembly and stabilization of transcription machinery. Accurate prediction of enhancers holds significant importance in comprehending gene regulation, a critical aspect for understanding health and disease. Identifying enhancers can unveil genetic foundations of diseases and offer potential targets for therapeutic interventions. Notably, encoder-only models like iEnhancer-BERT~\cite{luo2022ienhancer}, iEnhancer-ELM~\cite{li2023ienhancer} basically follow the same method, i.e., pretraining on human genome sequences and fintuning on enhancer data.
    
    \item \paratitle{Binding Site Prediction}.
    Binding site prediction aims to pinpoint regions on DNA, RNA, or proteins where specific molecules, such as transcription factors or peptides, can bind.
    EpiGePT~\cite{gao2023epigept} employs a bin-scanning strategy to detect potential binding sites for a vast set of human transcription factors, selecting the maximum score for each factor. By integrating motif and expression features, EpiGePT offers a comprehensive representation for each DNA bin, significantly enhancing the accuracy of binding site predictions.
\end{itemize}

\paratitle{Structure Prediction.} 
Structure Prediction involves computational tools for identifying and modeling biologically significant nucleic acid structures. These tools also aid in designing novel molecular architectures for nanotechnology and synthetic biology. The field has seen progress in predicting RNA three-dimensional structure directly from sequence and in designing sequences for predefined DNA and RNA nanostructures, indicating that nucleic acid structure is both predictable and controllable. Here we present two subtasks that are frequently used for structure prediction.
\begin{itemize}
 \item \paratitle{Chromatin Profile Prediction.}
    It involves the accurate prediction of the intricate structure of chromatin—a complex formed by DNA and proteins within cells. Understanding these predictions unveils crucial insights into gene regulation mechanisms and the role of chromatin alterations in diseases, notably cancer. 
    HyenaDNA~\cite{nguyen2023hyenadna} concentrates on predicting chromatin profiles and epigenetic markers from DNA sequences, quantifying functional effects caused by non-coding variants. 
    \item \paratitle{DNA/RNA-Protein Interaction Prediction}.
    It aims to understand the intricate interplay between nucleic acids and proteins. This interaction is pivotal for numerous biological processes, including transcriptional regulation, RNA splicing, and translation.   TFBert~\cite{luo2023improving} leverages the power of pre-trained BERT architectures for predicting DNA-protein interactions. 
\end{itemize}

\paratitle{Sequence Generation.}
Generation ability stands as a pivotal task in bioinformatics, focusing on models' capacity to create artificial sequences that closely mimic real biological sequences. This ability becomes particularly crucial when considering the generation of artificial human genomes (AGs), serving as tools to safeguard genetic privacy and reduce costs linked with genetic sample collection. 
DNAGPT~\cite{zhang2023dnagpt} emerges as a model purpose-built for this task. This model generated 5,000 AGs, each spanning a region of 10,000 Single Nucleotide Polymorphisms (SNPs). When compared with other models, DNAGPT showcased superior performance across various metrics.

\paratitle{Sequence Variation and Evolution Analysis.} 
It plays a pivotal role in deciphering DNA sequence alterations and their evolutionary trajectories. Understanding these variations is essential in unraveling the genetic foundations of traits, diseases, and evolutionary patterns. 
GenSLMs~\cite{zvyagin2022genslms} were adapted from large language models to analyze genomic data, specifically focusing on the evolutionary landscape of SARS-CoV-2 genomes. In a different vein, the GPN-MSA~\cite{benegas2023gpn} model introduces a novel framework for DNA language models by harnessing whole-genome sequence alignments across multiple species. 

\paratitle{Evaluation Metric.}
The evaluation metrics for function prediction mainly include accuracy, F1-score, sensitivity, specificity, and precision.
For structure prediction, a common evaluation approach is to assess the accuracy of predicted DNA structures against known structures. Median AUROC (Area Under the Receiver Operating Characteristic Curve) is often used for chromatin profile prediction in structure prediction. 
In sequence generation tasks, the evaluation of generated labels plays a pivotal role. Metrics like Matthews correlation coefficient (MCC), perpendicular distance, true positive rate, and false positive rate values are widely used for label classification.
As for the sequence variation and evolution analysis, AUROC can be used for classifying pathogenic variants. Additionally, the accuracy in identifying variants of concern is also a critical metric. 

\subsection{Summary}\label{sec:gene-LLM-summary} 
This section provides a comprehensive overview of Genomic Large Language Models (Gene-LLMs), detailing their application in computational biology for analyzing DNA and RNA sequences. It categorizes Gene-LLMs into encoder-only, decoder-only, and encoder-decoder models, highlighting their architecture, datasets, and evaluation methods. The discussion emphasizes the significance of these models in understanding gene function, chromatin profile prediction, splice site prediction, binding site prediction, and structure prediction, showcasing their capabilities in generating sequences and analyzing sequence variation and evolution. 
However, it is evident that the availability of generative Gene-LLMs (e.g., decode-based models) is severely limited, necessitating the development of robust generative models to facilitate diverse genomics applications.

	\clearpage
	
	

	\section{Multi-modal Scientific Large Language Models} \label{sec-multimodal}
\begin{figure*}[h]
	\centering
	\resizebox{\textwidth}{!}{
		\begin{forest}
			for tree={
				grow=east,
				reversed=true,
				anchor=base west,
				parent anchor=east,
				child anchor=west,
				base=left,
				font=\small,
				rectangle,
				draw,
				rounded corners,align=left,
				minimum width=2.5em,
				inner xsep=4pt,
				inner ysep=1pt,
			},
			where level=1{text width=5em,fill=blue!10}{},
			where level=2{text width=5em,font=\footnotesize,fill=pink!30}{},
			where level=3{font=\footnotesize,yshift=0.26pt,fill=yellow!20}{},
			[MM-Sci-LLMs, fill=green!10
			[Models\\(Sec.~\ref{sec:mm-sci-LLM-overview}),text width=4em
			[Molecule-Text, text width=5.5em
			[\emph{Encoder-only:} 
			Text2Mol~\cite{edwards2021text2mol} \, 
			KV-PLM~\cite{zeng2022deep}\,
			MoMu~\cite{zeng2022deep}\,
			MolFM~\cite{luo2023molfm}
			GPT-MolBERTa~\cite{balaji2023gptmolberta}
			]
			[\emph{Decoder-only:} 
			DrugChat~\cite{liang2023drugchat}\, 
			MolReGPT~\cite{li2023empowering}\,
			MolXPT~\cite{liu-etal-2023-molxpt}\,
			DrugLLM~\cite{liu2024drugllm}\,
			MolecularGPT~\cite{liu2024moleculargpt}
			]
			[\emph{Encoder-Decoder:}
			MolT5~\cite{edwards2022translation}\,
			Text+Chem T5~\cite{edwards2022translation}\,
			ChatMol~\cite{zeng2023interactive}\,
			GIT-Mol~\cite{liu2023gitmol}\\
			MoleculeSTM~\cite{liu2023multimodal}\,
			GIMLET~\cite{zhao2023gimlet}\,
			Atomas~\cite{zhang2024atomas}\,
			3D-MolT5~\cite{pei20243d}
			]
			]
			[Protein-Text, text width=5.5em
			[\emph{Encoder-only:} 
			ProTranslator~\cite{xu2022protranslator}\,
			ProtST-ProtBert~\cite{xu2023protst} 
			]
			[\emph{Decoder-only:} 
			InstructProtein~\cite{wang2023instructprotein}\,
			ProLLaMA~\cite{lv2024prollama}\,
			ProtLLM~\cite{zhuo2024protllm}
			]
			[\emph{Encoder-Decoder:}
			ProteinDT~\cite{liu2023textguided}\,
			Prot2Text~\cite{abdine2023prot2text}\,
			ProtST-ESM-1B~\cite{xu2023protst}\,
			ProtST-ESM-2~\cite{xu2023protst}\\
			PAAG~\cite{yuan2024functional}\,
			ProtT3~\cite{liu2024prott3}
			]
			]
			[Protein-Molecule, text width=5.5em
			[\emph{Encoder-only:} 
			DrugCLIP~\cite{gao2023drugclip}]
			[\emph{Decoder-only:} 
			DrugGPT~\cite{li2023druggpt}
			]
			[\emph{Encoder-Decoder:} 
			ChemBERTaLM~\cite{uludoğan2022exploiting}\,
			DeepTarget~\cite{chen2023deep}
			]
			] 
			[Gene/Cell-Text, text width=5.5em
			[\emph{Decoder-only:} 
			Cell2Sentence~\cite{levine2023cell2sentence}      
			GenePT~\cite{chen2023genept}        
			ChatCell~\cite{pei2024leveraging}
			GPTCelltype~\cite{hou2024assessing}]
			[\emph{Encoder-Decoder:} 
			CellPLM~\cite{chang2024bidirectional}            CellWhisperer~\cite{schaefer2024joint}
			LangCell~\cite{zhao2024langcell}
			]
			]
			[Comprehensive, text width=5.5em
			[\emph{Encoder-only:}BioTranslator~\cite{xu2023multilingual}\,
			BioBridge~\cite{wang2023biobridge}]
			[\emph{Decoder-only:} 
			Galactica~\cite{taylor2022galactica}\,
			ChatDrug~\cite{liu2023chatgptpowered}\,
			DARWIN-MDP~\cite{xie2023darwin}\,\\
			BioMedGPT-10B~\cite{luo2023biomedgpt}\,
			Mol-Instructions~\cite{fang2023molinstructions}\,
			ChemDFM~\cite{zhao2024chemdfm}\,
			ChemLLM~\cite{zhang2024chemllm}
			]
			[\emph{Encoder-Decoder:} 
			BioT5~\cite{pei2023biot5}\,
			BioT5+~\cite{pei2024biot5+}
			]
			]  
			]
			[Datasets\\ 
			(Sec.~\ref{sec:mm-sci-LLM-dataset}),text width=4em
			[Training, text width=3.5em
			[SwissProtCLAP~\cite{liu2023textguided} \,
			ProtDescribe~\cite{xu2023protst}\,
			InstructProtein~\cite{wang2023instructprotein}\,
			Scientific Knowledge Dataset~\cite{xie2023darwin}\\
			BioLip~\cite{yang2012biolip}\,
			ChEBL-dia~\cite{zeng2023interactive}\,
			PubChemQA~\cite{luo2023biomedgpt}\,
			Prot2Text~\cite{abdine2023prot2text}\,
			ProteinLMDataset~\cite{shen2024fine}\,
			PCdes~\cite{zeng2022deep}\,
			PubChemSTM~\cite{liu2023multimodal}\,\\
			MoMu~\cite{zeng2022deep}\,
			ChEBI-20~\cite{edwards2021text2mol}\,
			Mol-Instructions~\cite{fang2023molinstructions}\,
			GEO~\cite{clough2016gene}\,
			HCA~\cite{regev2017human}\, ARCHS4~\cite{lachmann2018massive}\,\\
			NCBI~\cite{wheeler2007database}\,              cellxgene~\cite{megill2021cellxgene}\,
			SRT~\cite{he2021high}\,   CellTypist~\cite{dominguez2022cross}\,
			scLibrary~\cite{zhao2024langcell}\,
			]
			]
			[Benchmark, text width=3.5em
			[
			MoleculeNet ~\cite{wu2018moleculenet} \, 
			BindingDB~\cite{gilson2016bindingdb} \, 
			DUD-E~\cite{mysinger2012directory}\,
			ProteinLMBench~\cite{shen2024fine}\   
			]
			]      
			]
			[Evaluation\\
			(Sec.~\ref{sec:mm-sci-LLM-eval}),text width=4em
			[Cross-modal Prediction, text width=7.5em
			[Text-based Function prediction across different modalities
			]
			]
			[Cross-modal Retrieval, text width=7.5em
			[Mutual retrieval across different modalities
			]
			]   
			[Cross-modal Generation, text width=7.5em
			[Mutual generation across different modalities
			]
			] 
			[Evaluation Metric, text width=7.5em
			[ Refer to Sec. \ref{sec:text-LLM-eval}; \ref{sec:mol-LLM-eval}; \ref{sec:prot-LLM-eval}; \ref{sec:gene-LLM-eval}
			]
			]
			]
			[Summary\\
			(Sec.~\ref{sec:mm-sci-LLM-summary}),text width=4em
			]            
			]
		\end{forest}
	}
	\caption{Chapter overview of MM-Sci-LLMs.}
	\label{fig:mm-sci-LLM-overview}
\end{figure*}
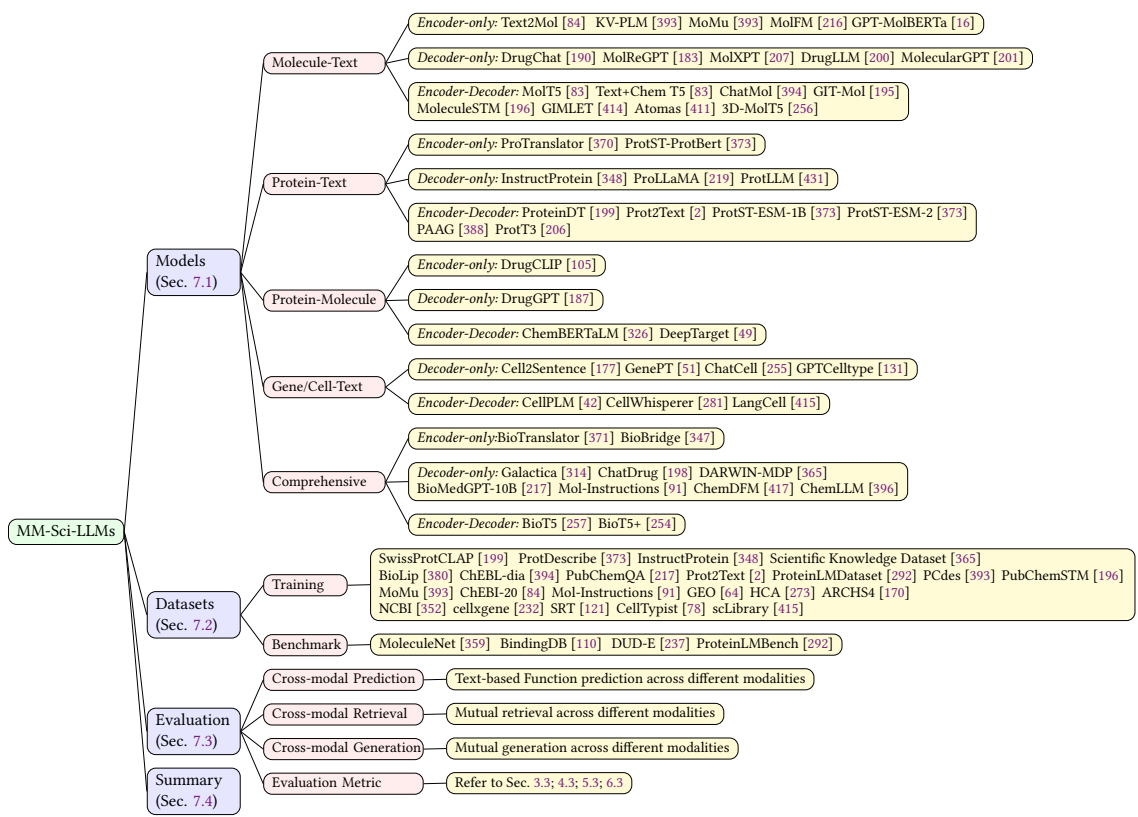

Multimodal Large Language Models have emerged as a prominent research area, utilizing robust LLMs as the core to handle multimodal data. These models possess a distinctive capability to incorporate diverse data types like text, images, audio, and other forms of information, enabling comprehensive exploration and problem-solving across various domains. The multimodal models hold promising prospects, particularly in the realms of biological and chemical sciences encompassing protein, molecular, and genomic studies. In this section, we explore recent advancements in multimodal models within these scientific fields (i.e., MM-Sci-LLMs), highlighting their capabilities and the exploited datasets. Note that, this survey focuses on cross-lingual multimodal models, involving at least two languages from different domains, such as text and molecule. Hence, we exclude the monolingual multimodal approaches from MM-Sci-LLMs, like the joint modeling of protein sequences and structures \cite{su2023saprot, wang2023multilevel, zhang2023systematic}. The diverse forms of different languages and modalities are illustrated in Figure \ref{fig:languages}, and the overview of this section is shown in Figure~\ref{fig:mm-sci-LLM-overview}.

\begin{table*}[htbp]
	\centering
	\caption{Summary of MM-Sci-LLMs}
	\label{tab:MM-Sci-LLM}
	\footnotesize
	\renewcommand\tabcolsep{2.5pt}
	\begin{tabular}{llccccc}
		\toprule
		&Model&Time&\#Parameters&Base Model&Pretraining Dataset& \makecell{Open-\\source}
		\\
		\midrule
		\multirow{21}{*}{Mol.\&Text}       
		&\href{https://github.com/cnedwards/text2mol}{Text2Mol}~\cite{edwards2021text2mol}&2021.11&117M&GCN, SciBERT&PubChem, ChEBI& \checkmark\\
		&\href{https://github.com/thunlp/KV-PLM}{KV-PLM}~\cite{zeng2022deep}&2022.02&110M&SciBERT&
		\makecell{MoleculeNet, USPTO 1k TPL~\cite{schneider2016s}, \\Chemprot~\cite{kringelum2016chemprot}, BC5CDR~\cite{li2016biocreative}}
		& \checkmark\\
		&\href{https://github.com/BingSu12/MoMu}{MoMu}~\cite{su2022molecular}&2022.09&110M&SciBERT, GIN&S2ORC~\cite{lo2020s2orc}, PubChem, OGB~\cite{hu2020open}& \checkmark \\
		&\href{https://github.com/blender-nlp/MolT5}{MolT5}~\cite{edwards2022translation}&2022.11&60M&T5&ChEBI-20& \checkmark \\
		&\href{https://github.com/GT4SD/multitask_text_and_chemistry_t5}{Text+Chem T5}~\cite{christofidellis2023unifying}&2023.05&40M/220M&T5&ChEBI-20& \checkmark \\
		&\href{https://github.com/UCSD-AI4H/drugchat}{DrugChat}~\cite{liang2023drugchat}&2023.05&13B&GNN, Vicuna&PubChem, ChEMBL& \checkmark \\
		&\href{https://github.com/zhao-ht/GIMLET}{GIMLET}~\cite{zhao2023gimlet}&2023.06&-&T5&Chembl& \checkmark \\
		&\href{https://github.com/phenixace/MolReGPT}{MolReGPT}~\cite{li2023empowering}&2023.06&-&ChatGPT&ChEBI-20& \checkmark \\
		&\href{https://github.com/Ellenzzn/ChatMol/tree/main}{ChatMol}~\cite{zeng2023interactive}&2023.06&-&T5&ChEBL-dia~\cite{zeng2023interactive}& \checkmark \\
		&MolXPT~\cite{liu-etal-2023-molxpt}&2023.05&350M& GPT-2 &PubChem, PubMed& $\times$ \\
		
		&\href{https://github.com/PharMolix/OpenBioMed}{MolFM}~\cite{luo2023molfm}&2023.07&138M&GIN, Trans. enc., TransE~\cite{bordes2013translating}&PubChem, S2ORC& \checkmark \\
		&GIT-Mol~\cite{liu2023gitmol}&2023.08&700M&\makecell{SciBERT, GIN,\\SwinTransformer~\cite{liu2021swin}}  &PubChem, ChEBI-20& $\times$ \\
		&\href{https://github.com/Suryanarayanan-Balaji/GPT-MolBERTa}{GPT-MolBERTa}~\cite{balaji2023gptmolberta}&2023.10&-&RoBERTa&MoleculeNet& \checkmark \\
		&\href{https://github.com/chao1224/MoleculeSTM/tree/main}{MoleculeSTM}~\cite{liu2023multimodal}&2023.12&252M&\makecell{GraphMVP~\cite{liu2021pre}, SciBERT \\Chemformer~\cite{irwin2022chemformer}} 
		&PubChemSTM& \checkmark \\
		&\href{https://anonymous.4open.science/r/Atomas-03C3}{Atomas}~\cite{zhang2024atomas}&2024.04&60M&T5
		&PubchemSTM-distll& \checkmark \\
		&DrugLLM~\cite{liu2024drugllm}&2024.05&7B&LLaMA 
		&ZINC, ChEMBL& $\times$ \\
		&3D-MolT5~\cite{pei20243d}&2024.06&255M&T5
		&C4, PubMed Central~\cite{canese2013pubmed, white2020pubmed} & $\times$ \\
		&\href{https://github.com/NYUSHCS/MolecularGPT}{MolecularGPT}~\cite{liu2024moleculargpt}&2024.06&7B&LLaMA2-chat-7B
		&ChEMBL, QM9& \checkmark \\
		
		\midrule
		\multirow{11}{*}{Prot.\&Text}
		&\href{https://github.com/HanwenXuTHU/ProTranslator}{ProTranslator}~\cite{xu2022protranslator}&2022.04&-&PubMedBert&-&  \checkmark \\
		&ProteinDT~\cite{liu2023textguided}&2023.02&-&SciBERT, ProtBERT&SwissProtCLAP&  $\times$ \\
		&\href{https://github.com/DeepGraphLearning/ProtST}{ProtST-ProtBert}~\cite{xu2023protst}&2023.07&420M&ProtBert, PubMedBERT&ProtDescribe& \checkmark \\
		&\href{https://github.com/DeepGraphLearning/ProtST}{ProtST-ESM-1b}~\cite{xu2023protst}&2023.07&650M&ESM-1b, PubMedBERT&ProtDescribe& \checkmark \\
		&\href{https://github.com/DeepGraphLearning/ProtST}{ProtST-ESM-2}~\cite{xu2023protst}&2023.07&650M&ESM-2, PubMedBERT&ProtDescribe& \checkmark \\
		&Prot2Text~\cite{abdine2023prot2text}&2023.07&283M&ESM2, RGCN~\cite{schlichtkrull2018modeling}, Trans. dec.&Prot2Text& $\times$ \\
		&InstructProtein~\cite{wang2023instructprotein}&2023.10&1.3B&OPT~\cite{zhang2022opt}&InstructProtein& $\times$ \\
		&\href{https://github.com/Lyu6PosHao/ProLLaMA}{ProLLaMA}~\cite{lv2024prollama}&2024.02&-&LLaMA2&UniRef50& \checkmark \\
		&\href{https://github.com/ProtLLM/ProtLLM}{ProtLLM}~\cite{zhuo2024protllm}&2024.02&-&LLaMA-7B&InterPT~\cite{zhuo2024protllm}& \checkmark \\
		&\href{https://github.com/DeepGraphLearning/ProtST}{ProtT3}~\cite{liu2024prott3}&2024.05&-&ESM2, Galactica&Swiss-Prot~\cite{bairoch2000swiss}, ProteinKG25~\cite{zhang2022ontoprotein}& \checkmark \\
		&PAAG~\cite{yuan2024functional}&2024.04&1.3B&SciBERT, ProtBert&ProtAnnotation& $\times$ \\
		\midrule
		\multirow{7}{*}{Cell\&Text}               
		&\href{https://github.com/vandijklab/cell2sentence-ft}{Cell2Sentence}~\cite{levine2023cell2sentence} &2023.09 &93M/313M &GPT-2 &CellTypist~\cite{dominguez2022cross}, cellxgene~\cite{megill2021cellxgene} &\checkmark \\
		&\href{https://github.com/OmicsML/CellPLM}{CELLPLM}~\cite{chang2024bidirectional} &2023.10 &80M &Transformer &GEO~\cite{clough2016gene}, HCA, HTCA~\cite{regev2017human}, SRT~\cite{he2021high} &\checkmark \\
		&\href{https://github.com/yiqunchen/GenePT}{GenePT}~\cite{chen2023genept} &2023.10 &- &GPT-3.5 &NCBI~\cite{wheeler2007database} &\checkmark \\
		&\href{https://github.com/zjunlp/ChatCell}{ChatCell}~\cite{pei2024leveraging} &2024.02 &- &GPT-4 &- &\checkmark \\&\href{https://github.com/Winnie09/GPTCelltype}{GPTCelltype}~\cite{hou2024assessing} &2024.03 &- &GPT-4 &- &\checkmark \\     
		&CellWhisperer~\cite{schaefer2024joint} &2024.03 &- &Geneformer, BioBERT &ARCHS4 ~\cite{lachmann2018massive}&$\times$ \\
		&\href{https://github.com/PharMolix/LangCell}{LangCell}~\cite{zhao2024langcell} &2024.06 &- &Geneformer, BERT, PubMedBERT &scLibrary~\cite{zhao2024langcell} &\checkmark \\ 
		\midrule
		\multirow{4}{*}{Mol.\&Prot.}
		&\href{https://github.com/boun-tabi/biochemical-lms-for-drug-design}{ChemBERTaLM}~\cite{uludoğan2022exploiting}&2022.09&-&RoBERTa, ChemBERTa&BindingDB, MOSES~\cite{polykovskiy2020molecular}& \checkmark \\
		&\href{https://github.com/viko-3/TargetGAN}{DeepTarget}~\cite{chen2023deep}&2023.03&-& Transformer&ChEMBL& \checkmark \\
		&\href{https://github.com/LIYUESEN/druggpt}{DrugGPT}~\cite{li2023druggpt}&2023.06&1.5B&GPT-2&ZINC20& \checkmark \\
		&DrugCLIP~\cite{gao2023drugclip}&2023.10&-&SE(3) 3D Transformer&PDBBind~\cite{wang2005pdbbind}, BioLip, ChEMBL& $\times$ \\
		
		\midrule
		\multirow{12}{*}{Comprehensive}
		&\href{https://galactica.org/mission/}{Galactica}~\cite{taylor2022galactica}&2022.11&120B&Trans. dec.&-& \checkmark\\
		&\href{https://github.com/HanwenXuTHU/BioTranslatorProject}{BioTranslator}~\cite{xu2023multilingual}&2023.02&-&PubMedBERT&-& \checkmark\\
		&\href{https://github.com/chao1224/ChatDrug}{ChatDrug}~\cite{liu2023chatgptpowered}&2023.05&-&ChatGPT&-&  \checkmark\\
		&\href{https://github.com/PharMolix/OpenBioMed}{BioMedGPT-10B}~\cite{luo2023biomedgpt}&2023.08&10B&LLaMA2, GIN, ESM-2&PubChemQA, UniProtQA&  \checkmark\\
		&\href{https://github.com/MasterAI-EAM/Darwin}{DARWIN-MDP}~\cite{xie2023darwin}&2023.08&7B&LLaMA&SciQ, FAIR~\cite{wilkinson2016fair}& \checkmark \\
		&\href{https://github.com/QizhiPei/BioT5}{BioT5}~\cite{pei2023biot5}&2023.10&252M&T5&ZINC20& \checkmark \\
		&\href{https://github.com/zjunlp/Mol-Instructions}{Mol-Instructions}~\cite{fang2023molinstructions}&2023.11&-&LLaMA&Mol-Instructions& \checkmark \\
		&\href{https://github.com/RyanWangZf/BioBridge}{BioBridge}~\cite{wang2023biobridge}&2024.01&-&KGs&PrimeKG& \checkmark \\
		& \href{https://huggingface.co/OpenDFM/ChemDFM-13B-v1.0}{ChemDFM}~\cite{zhao2024chemdfm} & 2024.01 & 13B &  LLaMa-13B &  Chemical literature, textbooks & \checkmark  \\
		& \href{https://github.com/keyhsw/ChemLLM}{ChemLLM}~\cite{zhang2024chemllm} & 2024.02 & 7B &  InternLM2 &  ChemData and Multi-Corpus & \checkmark  \\
		&\href{https://github.com/evo-design/evo}{Evo}~\cite{nguyen2024sequence}&2024.02&7B&StripedHyena&OpenGenome& \checkmark \\
		&\href{https://github.com/QizhiPei/BioT5}{BioT5+}~\cite{pei2024biot5+}&2024.02&252M&T5&PubChem, Uniref50, C4& \checkmark \\
		\bottomrule
	\end{tabular}
\end{table*}

\subsection{Models}
\label{sec:mm-sci-LLM-overview}
To provide a comprehensive overview of MM-Sci-LLMs, we categorize them into four groups: molecule-text models, protein-text models, protein-molecule models, and comprehensive models, based on the specific modality they focus on. Table \ref{tab:MM-Sci-LLM} shows the summary of MM-Sci-LLMs.

\subsubsection{Molecule-Text Models}  \ \\
Molecule-text models are designed to understand both the chemical structure and text descriptions of molecules by creating associations between text-molecular structure data pairs. This approach enables the model to effectively discern distinctions between various molecules, enhancing its ability for generalization.

Text2Mol~\cite{edwards2021text2mol}, KV-PLM~\cite{zeng2022deep}, MoMu~\cite{su2022molecular}, MolFM~\cite{luo2023molfm}, GPT-MolBERTa~\cite{balaji2023gptmolberta} and MoleculeSTM ~\cite{liu2023multimodal} represent a suite of encoder-based molecule-text multimodal models. 
Text2Mol~\cite{edwards2021text2mol} is a multimodal embedding approach to establish an aligned semantic space between text and molecules, facilitating cross-modal retrieval. This approach introduces the CHEBI-20 dataset and employs cross-modal attention-based association rules to enhance the performance in downstream tasks.
KV-PLM~\cite{zeng2022deep} represents a cross-modal learning approach that integrates molecular structural information with SMILES into textual knowledge. Similar to BERT's pre-training in an unsupervised fashion, this method facilitates the acquisition of meta-knowledge regarding structural fragments within the textual context. 
MoMu~\cite{su2022molecular} is a molecular multimodal model pretrained on both molecular graphs with Graph Isomorphism Network (GIN)~\cite{xu2018powerful} and their semantically related textual data with SciBERT~\cite{beltagy2019scibert} via contrastive learning, establishing a crucial link between molecular graphs and natural language.
MolFM~\cite{luo2023molfm} serves as a multimodal molecular foundation model, facilitating joint representation learning from molecular structures, descriptive texts, and knowledge graphs. It proposes a cross-modal attention mechanism that enhances the comprehension of molecular structures by incorporating atoms, neighboring molecule entities, and semantically related texts.
GPT-MolBERTa~\cite{balaji2023gptmolberta} utilizes ChatGPT to generate rich textual descriptions of molecules, which is then used to pre-train the RoBERTa model in a self-supervised manner. 
MoleculeSTM~\cite{liu2023multimodal} employs a molecule encoder and a text encoder with a contrastive learning strategy to jointly learn the chemical structures and their textual descriptions. 

DrugChat~\cite{liang2023drugchat}, MolXPT~\cite{liu-etal-2023-molxpt}, MolReGPT~\cite{li2023empowering},  DrugLLM~\cite{liu2024drugllm} and MolecularGPT~\cite{liu2024moleculargpt} are decoder-based molecule-text multimodal models. 
DrugChat~\cite{liang2023drugchat} combines a graph neural network (GNN) and Vicuna~\cite{vicuna2023} (a classical LLM) to enable ChatGPT-like capabilities on drug molecule graphs. 
MolXPT~\cite{liu-etal-2023-molxpt} is a generative pre-trained model that jointly models text and molecules. It aims to enhance the interaction between molecules and text, facilitating tasks such as molecular property prediction and text-guided molecule generation.
MolReGPT~\cite{li2023empowering} framework utilizes in-context few-shot molecule learning to enable language models to perform molecule-caption translation tasks without fine-tuning. 
{DrugLLM~\cite{liu2024drugllm} is an open large language model specialized for few-shot molecule generation, designed to predict and create new molecules with desired pharmacochemical properties based on limited examples, thereby advancing the efficiency of drug discovery processes.
	MolecularGPT~\cite{liu2024moleculargpt} is an innovative LLM fine-tuned for few-shot molecular property prediction, achieving remarkable performance by leveraging a hybrid instruction set and demonstrating the potential of LLMs as universal few-shot learners in the molecular domain.
}

MolT5~\cite{edwards2022translation}, Text+Chem T5~\cite{christofidellis2023unifying}, GIMLET~\cite{zhao2023gimlet}, ChatMol~\cite{zeng2023interactive}, GIT-Mol~\cite{liu2023gitmol}, Atomas~\cite{zhang2024atomas} and 3D-MolT5~\cite{pei20243d} use encoder-decoder architectures. 
MolT5~\cite{edwards2022translation} is a T5-based self-supervised learning framework that allows for the translation between molecules and natural language. 
Text+Chem T5~\cite{christofidellis2023unifying} leverages multi-task learning to unify molecular and textual representations, which can solve language tasks, chemical tasks, and cross-domain tasks, without the need for task-specific fine-tuning or retraining.
GIMLET~\cite{zhao2023gimlet} is a novel model specifically designed for zero-shot learning in molecular property prediction. It distinctively integrates graph and text data using a transformer-based approach, with an emphasis on generalized position embedding.
ChatMol~\cite{zeng2023interactive} is initialized with T5 and conducts multi-task pre-training, facilitating conversational molecular design through natural language interaction.
Based on Q-Former design from BLIP2 \cite{li2023blip}, GIT-Mol~\cite{liu2023gitmol} proposes a novel modality mixer designed with a cross-attention mechanism, enabling seamless fusion of multiple molecular modalities, including SMILES, texts, images, and structure graphs.
{Atomas~\cite{zhang2024atomas} is a multi-modal molecular representation learning framework that employs hierarchical alignment to jointly optimize molecule understanding and generation tasks, achieving state-of-the-art performance without requiring explicit local annotations.
	3D-MolT5~\cite{pei20243d} is a pioneering framework that integrates 1D molecular sequences and 3D structures into a unified model, enhancing language models' capabilities for diverse molecular tasks through innovative 3D molecular tokenization and extensive multi-task pre-training.
}


\subsubsection{Protein-Text Models}  \ \\
Texts from literature and databases have become increasingly important in protein research, leading to the emergence of protein-text multimodal models. 
ProTranslator~\cite{xu2022protranslator} innovatively approaches protein function prediction by redefining it as a machine translation problem. By utilizing textual descriptions, ProTranslator can annotate proteins to functions without the need for pre-known associated proteins.
ProteinDT~\cite{liu2023textguided} adopts contrastive learning to align the representation space of text and protein sequences. The ProteinDT model addresses the problem of aligning text and protein sequences to enable text-guided protein design and editing tasks.
ProtST series~\cite{xu2023protst}, including ProtST-ESM-1B, ProtST-ESM-2, and ProtST-ProtBert, are models enhancing protein sequence pre-training by integrating biomedical texts. Leveraging tasks like unimodal mask prediction and multimodal representation alignment, these models demonstrate superior performance in representation learning benchmarks.
Prot2Text~\cite{abdine2023prot2text} predicts a protein's function in free text format by integrating GNNs and LLMs in an encoder-decoder framework.  
InstructProtein~\cite{wang2023instructprotein} stands out as a decode-only LLM bridging the gap between human and protein languages through bidirectional generative capabilities. 
{
	ProLLaMA~\cite{lv2024prollama} leverages a pre-trained general LLM (i.e., LLaMA2) and transforms it for protein language processing through a two-stage training framework. It introduces a novel approach by incorporating Low-Rank Adaptation (LoRA) for efficient training, allowing the model to retain natural language capabilities while learning protein language. 
	ProtLLM~\cite{zhuo2024protllm} is a versatile crossmodal LLM, pioneers an interleaved protein-text pre-training approach, empowering it with superior performance on both protein-centric and protein-language tasks through a dynamic protein mounting mechanism and protein-as-word language modeling.
	ProtT3~\cite{liu2024prott3} introduces a novel framework for protein-to-text generation that empowers language models to understand and generate textual descriptions of proteins by integrating a protein language model through a cross-modal projector, enhancing text-based protein understanding tasks.
	PAAG~\cite{yuan2024functional} pioneers a novel multimodal protein design approach by aligning textual annotations with protein sequences, enabling the controlled generation of proteins with specific domains and properties guided by rich, functional descriptions from protein databases.
}

\subsubsection{Gene/Cell-Text Models}  \ \\
Cellular behavior is fundamentally driven by gene expression, which dictates various cellular functions and activities. Gene/Cell-Text models employ cutting-edge natural language processing to decode the intricate language of cellular transcriptomes. By translating gene expression data into textual descriptions, these models effectively narrate cellular activities and interactions. This approach offers new insights into cell behavior, making complex biological processes more comprehensible by describing cellular functions through their underlying gene expression data. Cell2Sentence~\cite{levine2023cell2sentence} is a transformative approach that teaches language models to process and generate biologically relevant text, enhancing capabilities in cell generation, complex cell type annotation, and data-driven text generation. CellPLM~\cite{chang2024bidirectional} is a pre-trained transformer model designed to handle single-cell RNA sequencing (scRNA-seq) and spatial transcriptomics (SRT) data. It leverages spatially-resolved transcriptomic data and a Gaussian mixture prior distribution to effectively capture intricate cell-cell relationships, enabling superior performance in tasks like cell type annotation, spatial transcriptomic imputation, and scRNA-seq denoising. GenePT~\cite{chen2023genept}, leveraging GPT-3.5 model, generates comprehensive embeddings for genes and cells by integrating text descriptions from NCBI and gene expression data, enabling advanced genomic analysis. 
GPTCelltype model~\cite{hou2024assessing} is an R software package that harnesses the predictive capabilities of the GPT-4 language model for high-fidelity cell type annotation in single-cell RNA sequencing analysis, significantly streamlining the process and reducing the need for manual expertise. 
CellWhisperer~\cite{schaefer2024joint} leverages the Geneformer model for transcriptome encoding and BioBERT for text annotations, enabling capabilities such as free-text cell retrieval and zero-shot cell type classification in single-cell RNA sequencing datasets. 
LangCell~\cite{zhao2024langcell} pioneers an integrative pre-training paradigm, harnessing multifaceted ontological annotations to decode the semantic intricacies within single-cell RNA sequencing data, thereby enhancing the precision of cell identity inference across diverse biological spectra.

\subsubsection{Protein-Molecule Models}  \ \\
Protein-molecule models focus on converting protein information into molecular information and exploring potential associations between the two. Protein-molecular models typically employ an encoder-decoder architecture, where the encoder converts a protein sequence into fixed-length vectors, and the decoder generates molecular information from those vectors.
ChemBERTaLM~\cite{uludoğan2022exploiting} uses a pre-trained biochemical language model to initialize a target molecule generation model, aiming to address the challenge of limited data on new protein labeling in drug discovery. 
DeepTarget~\cite{chen2023deep} is an innovative deep generative model that designs new drug molecules directly from protein amino acid sequences, integrating adversarial networks and contrastive learning to minimize reliance on existing structural information.
DrugGPT~\cite{li2023druggpt} is built upon the GPT framework for ligand design, enabling precise capture of both structural information and chemical principles governing drug molecules. 
DrugCLIP~\cite{gao2023drugclip} is a novel approach to virtual screening for potential drugs that uses contrastive learning to align representations of binding protein pockets and molecules. 
These methods aim to improve the efficiency and accuracy of drug discovery through AI-assisted methods.


\subsubsection{Comprehensive Models}  \ \\
In this survey, the comprehensive multimodal models include text, proteins, molecules, genomes and other scientific languages. 
Their potential impact spans diverse domains, particularly offering significant support in accelerating fundamental science research.
Galactica~\cite{taylor2022galactica}, BioTranslator~\cite{xu2023multilingual}, ChatDrug~\cite{liu2023chatgptpowered}, BioMedGPT0-10B~\cite{luo2023biomedgpt}, DARWIN-MDP
~\cite{xie2023darwin}, BioT5~\cite{pei2023biot5}, Mol-Instructions~\cite{fang2023molinstructions}, BioBridge~\cite{wang2023biobridge}, ChemDFM~\cite{zhao2024chemdfm}, ChemLLM~\cite{zhang2024chemllm}, Evo~\cite{nguyen2024sequence} and BioT5+~\cite{pei2024biot5+} are representative comprehensive multimodal models.
Galactica~\cite{taylor2022galactica} is a scientific language model based on a decoder-only architecture, trained on a massive corpus of 106 billion tokens from various scientific texts and knowledge bases. Its strength lies in its ability to efficiently process extensive scientific datasets, enhancing its potential utility in specific scientific tasks.
BioTranslator~\cite{xu2023multilingual} leverages PubMedBERT~\cite{gu2021domain} for modeling textual descriptions and fine-tunes it using ontologies across various domains, showcasing its versatility in integrating diverse modalities for sophisticated biological data analysis.
ChatDrug~\cite{liu2023chatgptpowered} utilizes conversational language models to perform drug editing tasks. It addresses the problem of modifying small molecules, peptides, and proteins to achieve specific properties, such as solubility or secondary structure, by leveraging the knowledge extraction and summarization capabilities of the language model.
BioMedGPT-10B~\cite{luo2023biomedgpt} is an open multimodal generative pre-trained transformer, to bridge the modality gap in biomedicine by aligning different biological modalities with natural language.  It addresses the challenge of effectively communicating with diverse biological modalities through free text, including molecular structures, protein sequences, pathways, and cell transcriptomics.
DARWIN-MDP~\cite{xie2023darwin} is a variant of the DARWIN series that involves 16 FAIR datasets to generate instructions for material and device prediction tasks, such as classification, regression, and design. 
BioT5~\cite{pei2023biot5} is a pre-training framework that leverages structured and unstructured data sources to capture the underlying relations and properties of bio-entities. It aims to address the challenges in cross-modal generation and prediction tasks in the field of computational biology.
Mol-Instructions~\cite{fang2023molinstructions} aims to address the challenge of enhancing LLMs' performance in tasks related to molecules, proteins, and biomolecular text by providing task-specific instructions.
BioBridge~\cite{wang2023biobridge} stands out as an innovative model within the biomedical sphere, adeptly connecting unimodal Foundation Models using Knowledge Graphs. This model exploits the extensive yet fragmented biomedical datasets to propose a unique method for multimodal integration, circumventing the need for broad-scale retraining.
ChemDFM~\cite{zhao2024chemdfm}, built on the LLaMa-13B model, is trained on chemical literature to assist with molecular recognition, property prediction, and reaction analysis, and to support research through free-form dialogues.
ChemLLM~\cite{zhang2024chemllm} is a specialized large language model for chemistry, based on the InternLM2 model \cite{cai2024internlm2}. It utilizes chemical data for instruction tuning and has been evaluated by the proposed ChemBench dataset. On the other hand, ChemDFM~\cite{zhao2024chemdfm}, built on the LLaMa-13B model, focuses on training with chemical literature to assist in molecular recognition, property prediction, reaction analysis, and support research through free-form dialogues. PharmGPT~\cite{chen2024pharmgpt}, built on the LLaMa-13B and LLaMa-70B models, is trained using biopharmaceutical and chemical literature to assist in drug discovery, chemical synthesis assistance, and biological pathway analysis.
{Evo~\cite{nguyen2024sequence} utilizes the StripedHyena architecture and is trained on the OpenGenome datasets, which include whole prokaryotic genomes. The model has 7 billion parameters and is capable of modeling and generating DNA sequences at scales ranging from molecular to genome-wide. It is trained on 300 billion nucleotides, allowing it to perform zero-shot function prediction across DNA, RNA, and protein modalities.
	BioT5+~\cite{pei2024biot5+} is an extension of the BioT5 framework,  it integrates IUPAC names for improved molecular understanding and includes an extensive corpus of bio-text and molecule data from sources such as bioRxiv and PubChem.   The model employs multi-task instruction tuning for generality across tasks and introduces a novel numerical tokenization technique for enhanced processing of numerical data. } 

\subsection{Datasets}
\label{sec:mm-sci-LLM-dataset}

\begin{table*}[t]
	\centering
	\caption{Summary of datasets for MM-Sci-LLMs}
	\label{tab:MM-Sci-LLM-dataset}
	\footnotesize
	\renewcommand\tabcolsep{2.5pt}
	\begin{tabular}{llccc}
		\toprule
		&Dataset&Last updated&Scale&Keywords\\
		\midrule 
		\multirow{7}{*}{Mol.\&Text} 
		&\href{https://github.com/cnedwards/text2mol}{ChEBI-20}~\cite{edwards2021text2mol}&2021.11&33K&Cross-Lingual Retrieval\\
		&\href{https://github.com/thunlp/KV-PLM}{PCdes}~\cite{zeng2022deep}&2022.02&15K&SMILES strings, Cross-information Retrieval\\
		&\href{https://github.com/BingSu12/MoMu/tree/main/data}{MoMu}~\cite{su2022molecular}&2022.09&15K&Bidirectional Molecular Graph-to-text Retrieval\\
		&\href{https://github.com/chao1224/MoleculeSTM/tree/main/MoleculeSTM/datasets}{PubChemSTM}~\cite{liu2023multimodal}&2022.12&281K+250K&Structure-Text Pairing\\
		&\href{https://github.com/Ellenzzn/ChatMol/tree/main}{ChEBL-dia}~\cite{zeng2023interactive}&2023.06&10K&Molecule Generation Evaluation\\
		&\href{https://github.com/PharMolix/OpenBioMed}{PubChemQA}~\cite{luo2023biomedgpt}&2023.08& 325K+365K &Molecule Description\\&MoleculeQA~\cite{lu2024moleculeqa}&2024.03&62K+23K&Comprehensive Evaluation of Molecular\\
		\midrule
		\multirow{7}{*}{Prot.\&Text} 
		&SwissProtCLAP~\cite{liu2023textguided}&2023.02&441K&Protein Sequence Analysis\\
		&\href{https://github.com/DeepGraphLearning/ProtST}{ProtDescribe}~\cite{xu2023protst}&2023.07&553K&Protein Sequence Representation\\
		&Prot2Text~\cite{abdine2023prot2text}&2023.07&256K&Protein Function Prediction\\
		&\href{https://github.com/PharMolix/OpenBioMed}{UniProtQA}~\cite{luo2023biomedgpt}&2023.08&569K+1.8M&Protein Function and Property\\
		&InstructProtein~\cite{wang2023instructprotein}&2023.10&2.8M&Structured KG, Bidirectional Protein-Text Generation\\
		&\href{https://huggingface.co/datasets/tsynbio/ProteinLMBench}{ProteinLMDataset}~\cite{shen2024fine}&2024.06&89.3K&Protein Sequence-Text\\
		&\href{https://huggingface.co/datasets/tsynbio/ProteinLMBench}{ProteinLMBench}~\cite{shen2024fine}&2024.06&944&A benchmark for protein understanding\\
		\midrule
		\multirow{8}{*}{Cell\&Text}
		&\href{https://www.ncbi.nlm.nih.gov/geo/}{GEO}~\cite{clough2016gene} &2016 &1.3M & High-throughput Genomics\\
		&\href{https://www.humancellatlas.org/}{HCA}~\cite{regev2017human} &2017.12 &- & Single-cell Transcriptomics \\    
		&\href{https://amp.pharm.mssm.edu/archs4}{ARCHS4}~\cite{lachmann2018massive}&2018.04 &188K & Mouse and human RNA-seq samples\\
		&\href{https://www.ncbi.nlm.nih.gov/}{NCBI}~\cite{wheeler2007database} & 2020.10 &3B & Comprehensive Biotechnological Data\\
		&\href{https://github.com/chanzuckerberg/cellxgene}{cellxgene}~\cite{megill2021cellxgene} &2021.04 &37M &High-dimensional sparse matrices\\   
		&\href{https://nanostring.com/products/cosmx-spatial-molecular-imager/ffpe-dataset/nsclc-ffpe-dataset/}{SRT}~\cite{he2021high} &2021.11 & 2.7M &Spatial transcriptomic data \\    &\href{https://github.com/Teichlab/celltypist#interactive-tutorials}{CellTypist}~\cite{dominguez2022cross} &2022.05 &360K &Immune cell analysis, Cell type annotation\\
		&scLibrary~\cite{zhao2024langcell} &2024.06 &27.5M &Extract from cellxgene\\
		\midrule
		\multirow{3}{*}{Mol.\&Prot.} 
		&\href{https://dude.docking.org/}{DUD-E}
		~\cite{mysinger2012directory}&2012.06& 22k&Ligand Diversity, molecular docking\\
		&\href{https://zhanggroup.org/BioLiP/index.cgi}{BioLiP}~\cite{yang2012biolip}&2012.10& 204k &ligand-protein interactions, molecular docking\\
		&\href{https://www.bindingdb.org/rwd/bind/index.jsp}{BindingDB}~\cite{gilson2016bindingdb}&2016.01&1.1M&protein-ligand interactions, drug design\\
		\midrule
		\multirow{4}{*}{Comprehensive}   
		&Galactica~\cite{taylor2022galactica}&2022.11&62.2M&Large Scientific Corpus\\
		&\href{https://github.com/MasterAI-EAM/Darwin}{Scientific Knowledge Dataset}~\cite{xie2023darwin}&2023.08&13k+6M &biomolecular studies\\
		&\href{https://github.com/zjunlp/Mol-Instructions}{Mol-Instructions}~\cite{fang2023molinstructions}&2023.10&2M& Comprehensive Instruction, Diverse Text Descriptions\\
		&\href{https://huggingface.co/datasets/osunlp/SMolInstruct}{SMolInstruct}~\cite{yu2024llasmol}&2024.02&3.3M& Comprehensive Instruction, Diverse Text Descriptions\\
		\bottomrule
	\end{tabular}
\end{table*}

Similar to the taxonomy of multimodal models, we divide multimodal datasets into four categories. The following present popular datasets for each category, which is also summarized in Table \ref{tab:MM-Sci-LLM-dataset}.

\subsubsection{Molecule-Text Datasets} \ \\
The molecule-text datasets are mainly sourced from chemical literature, patents, drug instructions, etc., which contain descriptions of molecules. After pre-processing (e.g., data clearing), it can be used to train the model to achieve communication between molecular language and natural language. 
\begin{itemize}
	\item \paratitle{ChEBI-20}~\cite{edwards2021text2mol}. 
	CHEBI-20 was created by collecting ChEBI~\cite{hastings2016chebi} annotations of compounds from PubChem~\cite{kim2016pubchem,kim2019pubchem}, leading to a collection of 102,980 compound-description pairs. From this data, a subset of 33,010 pairs was curated to form the CHEBI-20 dataset. It can be applied to assess the model's efficacy in molecular captioning tasks. 
	
	\item \paratitle{PCdes}~\cite{zeng2022deep}. 
	PCdes comprises SMILES notations and corresponding property descriptions for 15,000 molecules from PubChem~\cite{wang2009pubchem}. This dataset facilitates cross-information retrieval tasks. For instance, KV-PLM~\cite{zeng2022deep}, when fine-tuned on PCdes, aims to optimally select matching SMILES strings or property description sentences for each other.
	
	\item \paratitle{MoMu}~\cite{su2022molecular}. 
	MoMu was developed by aggregating SMILES strings from the first 50,000 molecular compounds in the PubChem~\cite{wang2009pubchem} database and transforming them into molecular maps. This process was complemented by incorporating relevant scientific texts from the fields of medicine, biology, chemistry, and computer science, obtained from the S2ORC~\cite{lo2020s2orc} database. The final dataset comprised 15,613 pairs of graphic documents. It can be used for many tasks and applications related to molecular structure and properties, including molecular image description generation, zero-sample molecular generation, molecular property prediction, and molecular similarity calculation.
	
	\item \paratitle{PubChemSTM}~\cite{liu2023multimodal}. 
	PubChemSTM dataset is constructed using data from the PubChem~\cite{kim2021pubchem} database, which consists of 250,000 molecules and 281,000 structure-text pairs. PubChemSTM is used to train the MoleculeSTM model~\cite{liu2023multimodal}, and can be extended to structure-text retrieval tasks and text-based molecule editing tasks.
	
	\item \paratitle{PubChemQA}~\cite{luo2023biomedgpt}. 
	PubChemQA is composed of molecules and their associated textual descriptions sourced from PubChem~\cite{kim2023pubchem}. This dataset is characterized by a singular question format: ``Please describe the molecule." This dataset includes 325,754 unique molecules and 365,129 molecule-text pairs, with an average of 17 words per text description.
	
	\item \paratitle{MoleculeQA}~\cite{lu2024moleculeqa}. 
	MoleculeQA is a comprehensive question-answering dataset consisting of 62K pairs that evaluate factual accuracy for molecular comprehension. It covers a range of molecules with detailed descriptions linked to authoritative sources, providing a robust platform for assessing language models' capabilities in understanding molecular properties, sources, structures, and applications.

\end{itemize}

\subsubsection{Protein-Text Datasets} \ \\
The text-protein dataset focuses on converting text information into protein-related information, including predictions about protein structure, function, and interactions. These datasets typically contain protein sequences, descriptive text, and corresponding annotation information. They have a wide range of applications in bioinformatics and provide a rich resource for the research community to explore new methods and strategies for converting text information into protein information, or vice versa.

\begin{itemize}
	\item \paratitle{SwissProtCLAP}~\cite{liu2023textguided}. 
	The SwissProtCLAP dataset, derived from the SwissProt database~\cite{boutet2007uniprotkb}—a subset of UniProt~\cite{uniprot2021uniprot}, consists of 441,000 text-protein sequence pairs. This dataset finds extensive applications in bioinformatics and molecular biology, particularly in tasks requiring the integration of textual descriptions with protein sequences.
	\item \paratitle{ProtDescribe}~\cite{xu2023protst}. 
	The ProtDescribe dataset, comprising 553,052 aligned pairs of protein sequences and corresponding property descriptions, is constructed using data from the Swiss-Prot database~\cite{bairoch2000swiss}. This database is renowned for its high-quality annotations of diverse protein properties. The dataset facilitates the incorporation of protein property information into protein language models, thereby enhancing their ability to model semantic correlations between known and unknown protein properties.
	
	\item \paratitle{UniProtQA}~\cite{luo2023biomedgpt}. 
	UniProtQA is a dataset comprising proteins and corresponding textual queries about their functions and properties. This dataset, derived from UniProt~\cite{uniprot2023uniprot}, encompasses four distinct categories of inquiries related to functional attributes, official nomenclature, protein families, and sub-cellular localizations. It comprises an extensive compilation of 569,516 proteins along with 1,891,506 question-answering instances.
	
	\item \paratitle{InstructProtein}~\cite{wang2023instructprotein}. 
	The InstructProtein dataset is composed of protein-text instruction data derived from the protein knowledge graph constructed by the UniProt/Swiss-Prot database~\cite{uniprot2019uniprot}, which encompasses a comprehensive collection of 2.8 million data. It is the first high-quality protein instruction dataset for tuning LLMs.
	
	\item \paratitle{ProteinLMDataset}~\cite{shen2024fine}. 
	The InstructProtein dataset is composed of protein-text instruction data derived from the protein knowledge graph constructed by the UniProt/Swiss-Prot database~\cite{uniprot2019uniprot}, which encompasses a comprehensive collection of 2.8 million data. It is the first high-quality protein instruction dataset for tuning LLMs.
	
	\item \paratitle{ProteinLMBench}~\cite{shen2024fine}.
	ProteinLMBench stands as the inaugural benchmark for evaluating the protein comprehension proficiency of LLMs. It encompasses 944 meticulously curated multiple-choice questions that span a spectrum of protein-related tasks. This benchmark is designed to rigorously assess the interpretative and analytical skills of LLMs in the context of protein sequence understanding

\end{itemize}

\subsubsection{Gene/Cell-Text} \ \\
The integration of single-cell transcriptomics and natural language text data transforms gene expression profiles into textual sequences, enabling the application of large language models to biological data. This approach facilitates biologically relevant cell generation, accurate cell type predictions, and meaningful biological insights extraction through natural language processing.

\begin{itemize}
	\item \paratitle{GEO}~\cite{clough2016gene}. The Gene Expression Omnibus (GEO) is a vast public repository that archives high-throughput gene expression and functional genomics data sets, accommodating various technologies such as DNA microarrays, high-throughput sequencing, and genome methylation studies. With over 54,640 studies and more than 1.3 million samples from 2,889 different organisms, GEO provides extensive tools for data retrieval, visualization, and analysis, supporting researchers globally in exploring and interpreting complex genomic data. 
	\item \paratitle{HCA}~\cite{regev2017human}. The Human Cell Atlas dataset is a comprehensive reference map of all human cell types, characterized by high-throughput single-cell transcriptomics. It aims to profile millions of individual cells to understand their molecular signatures, locations, and interactions within healthy human tissues. This dataset provides invaluable insights into human physiology, development, and disease mechanisms.
	\item \paratitle{ARCHS4}~\cite{lachmann2018massive}. The dataset described is a comprehensive RNA-seq collection encompassing 187,946 samples, with 103,083 from mice and 84,863 from humans. This large-scale dataset is processed and accessible through the ARCHS4 web resource, offering extensive tools for gene expression analysis and visualization. It provides a valuable resource for retrospective and integrative analyses in genomic research.
	\item \paratitle{NCBI}~\cite{wheeler2007database}. The National Center for Biotechnology Information (NCBI) offers a comprehensive suite of biological databases, including GenBank for nucleic acid sequences and PubMed for life science literature. The Entrez system provides integrated access to 34 distinct databases, collectively containing billions of records across genomics, genetics, proteins, chemicals, and more. This extensive resource supports a wide range of biological research and data retrieval needs.
	\item \paratitle{cellxgene}~\cite{megill2021cellxgene}. The cellxgene dataset in question consists of high-dimensional sparse matrices, commonly used in single-cell RNA sequencing (scRNAseq) analyses. These matrices typically include millions of individual cell observations and tens of thousands of gene expression measurements, enabling detailed exploration and analysis of cellular heterogeneity and gene expression patterns across different conditions and populations.
	\item \paratitle{SRT}~\cite{he2021high}. The Spatial Transcriptomics (SRT) data set, specifically designed for high-plex molecular imaging, provides comprehensive insights into gene expression and protein localization within tissue sections. This data set allows for the spatial mapping of RNA and protein targets on FFPE (formalin-fixed, paraffin-embedded) samples, enabling detailed analysis at single-cell and subcellular resolutions. The utilization of advanced imaging technologies facilitates a deeper understanding of tissue architecture and cellular interactions in various biological contexts.
	\item \paratitle{CellTypist}~\cite{dominguez2022cross}. The CellTypist dataset comprises nearly 360,000 single-cell transcriptomes obtained from 16 different tissues of 12 deceased individuals. It was generated using single-cell RNA sequencing (scRNA-seq) and VDJ sequencing, focusing on immune cell profiling. The dataset supports comprehensive cell type annotation and the development of the CellTypist tool, providing detailed insights into the heterogeneity of immune cells across human tissues.
	\item \paratitle{scLibrary}~\cite{zhao2024langcell}. The scLibrary dataset, constructed from cellxgene, includes approximately 27.5 million pairs of single-cell RNA sequencing (scRNA-seq) data and their corresponding textual descriptions. It features comprehensive metadata annotations from various biological ontologies, providing detailed information on cell type, developmental stage, tissue, disease status, and more. This dataset enables enhanced understanding of cell identity through integrated transcriptomic and textual data.
\end{itemize}


\subsubsection{Molecule-Protein Datasets} \ \\
Protein-molecule interaction datasets are an important resource in the field of bioinformatics and drug discovery, providing detailed information on interactions between proteins and small molecule ligands. These datasets are critical for understanding complex interactions between proteins and small molecules, developing new drugs, and predicting molecular behavior.

\begin{itemize}
	\item \paratitle{DUD-E}~\cite{mysinger2012directory}. 
	The DUD-E dataset represents an enhanced benchmarking collection developed for evaluating molecular docking. It includes 102 proteins, accompanied by 22,886 clustered ligands sourced from ChEMBL~\cite{gaulton2012chembl}. This dataset was designed to offer a robust and diverse benchmark for well-researched targets, characterized by a substantial number of ligand affinities measurements and a variety of distinct cocrystal ligand structures.
	\item \paratitle{BioLiP}~\cite{yang2012biolip}. 
	The BioLiP dataset is constructed by utilizing protein structures from the Protein Data Bank (PDB)~\cite{rose2010rcsb} as templates, encompassing a total of 204,223 entries. This dataset serves multiple purposes, including aiding in template-based ligand-protein docking, virtual ligand screening, and annotating protein functions.
	\item \paratitle{BindingDB}~\cite{gilson2016bindingdb}. 
	The BindingDB dataset is predominantly constructed using data derived from scientific articles and, to an increasing extent, from US patents. It comprises roughly 1.1 million quantified protein-small molecule affinities, encompassing about 490,000 small molecules and several thousand proteins. This dataset is valuable for various purposes: medicinal chemists can employ it for designing drugs targeted at specific proteins; computational chemists can use it to test and refine algorithms in computer-aided drug design; and pharmacologists can utilize it to screen drug candidates for potential side effects or to formulate hypotheses about the mechanisms of action of new bioactive compounds.
\end{itemize}

\subsubsection{Comprehensive Datasets} \ \\
The comprehensive datasets amalgamate various data modalities, encompassing protein, molecular, genomic compositions, and associated textual data, thereby enabling a holistic perspective in comprehending biological entities. This dataset plays a crucial role in the training of LLMs for tasks such as prediction of protein-molecule interactions, drug discovery, and automated analysis of biomedical texts, effectively bridging the disciplinary divide between molecular biology and computational linguistics.

\begin{itemize}
	\item \paratitle{Galactica}~\cite{taylor2022galactica}. 
	The Galactica dataset, constructed from an extensive scientific corpus, encompasses over 62.22 million documents with 106 billion tokens. It uniquely combines natural language sources, such as academic papers and textbooks, with sequences found in nature, including protein sequences and chemical formulas. Furthermore, this dataset incorporates processed LaTeX to encapsulate intricate scientific data, as well as academic code pertinent to computational science. This composition makes the Galactica dataset an invaluable tool for training neural networks to decode and comprehend the diverse linguistic expressions of scientific disciplines.
	\item \paratitle{Scientific Knowledge Dataset}~\cite{xie2023darwin}. 
	It is constructed using two main sources: the SciQ dataset~\cite{welbl2017crowdsourcing}, which includes 13,679 science exam questions and answers, and a dataset of 6,057,261 full-text scholarly papers from the Web of Science.  This comprehensive dataset is designed to train and enhance large language models for scientific tasks, such as question-answering and knowledge modeling in natural science domains.
	\item \paratitle{Mol-Instructions}~\cite{fang2023molinstructions}. 
	The Mol-Instructions is a comprehensive focus on biomolecular studies and is divided into three main components: 
	\begin{itemize}
		\item \paratitle{Molecule-Oriented Instructions}: This segment comprises approximately 1.484 million instructions, distributed across six distinct tasks. It tackles key challenges in various chemical reactions and molecular design. 
		\item \paratitle{Protein-Oriented Instructions}: This segment encompasses roughly 505,000 instructions, spread over five categories of tasks.  These tasks are specifically designed to predict protein structure, function, and activity, and also support protein design based on textual instructions.
		\item \paratitle{Biomolecular Text Instructions}: This portion includes six information extraction and QA tasks represented through 53,000 instructions.
	\end{itemize}
	In summary, the Mol-Instructions dataset is instrumental for the training and improvement of LLMs, specifically in their ability to comprehend and interpret intricate biomolecular data.

	\item \paratitle{SMolInstruct}~\cite{yu2024llasmol}.
	It is a large-scale and high-quality instruction tuning dataset designed for advancing large language models in the field of chemistry. It encompasses 14 distinct chemistry tasks with over three million samples, facilitating capabilities such as name conversion, property prediction, molecule description, and chemical reaction analysis. This dataset serves as a robust foundation for training and evaluating LLMs on a wide array of chemistry-related tasks, pushing the boundaries of their performance in this scientific domain.
\end{itemize}

\subsection{Evaluation}
\label{sec:mm-sci-LLM-eval} 
When evaluating MM-Sci-LLMs in the realm of multimodal models, their capabilities are primarily assessed based on three critical areas: cross-modal prediction, retrieval and generation. 

\paratitle{Cross-modal Prediction.}
Cross-modal prediction is a task that involves using multimodal models to predict the functionality of molecules, proteins, and genomes based on textual instructions. These approaches combine different types of data, such as text and molecular structures, to make accurate predictions about the function of biological entities. The specific functions can be found in Section \ref{sec:mol-LLM-eval}, \ref{sec:prot-LLM-eval}, \ref{sec:gene-LLM-eval}. Most encoder-based multimodal models such as MoleculeSTM~\cite{liu2023multimodal} are designed for this task. Moreover, decode-only architectures such as Mol-Instructions~\cite{fang2023molinstructions} also possess the capability to directly generate functions.
As a fundamental task, cross-modal function prediction has various applications in bioinformatics and drug discovery, including identifying potential drug targets, understanding the interactions between proteins, and predicting the activity of molecules.

\paratitle{Cross-modal Retrieval.}
Cross-modal retrieval refers to the process of retrieving information from one modality based on a query from another modality, and vice versa. In the context of multimodal scientific language models, cross-modal retrieval primarily involves retrieving molecules, proteins, and genomes that are related to textual descriptions, and vice versa. 
For example, given a textual description of a molecule, a multimodal model can be used to retrieve similar molecules based on their functional descriptions. This can be useful in drug discovery, where researchers may want to find molecules that have specific properties. Models can accomplish these tasks include KV-PLM~\cite{zeng2022deep}, MoMu~\cite{su2022molecular} and MolFM~\cite{luo2023molfm}. 
On the other hand, given a protein or genomic dataset, a multimodal model such as ProtST-ESM-1b~\cite{xu2023protst} can be used to retrieve textual descriptions that provide information about the properties, functions, or interactions of the proteins or genes in the dataset. This can aid in understanding the biological mechanisms or relationships between different proteins or genes. 
Additionally, multimodal models can retrieve relevant protein sequences corresponding to a given molecular structure, or find the appropriate molecular structures based on a protein sequence. For example, DrugCLIP~\cite{gao2023drugclip} uses a dense retrieval approach, employing contrastive learning to maximize similarity between a protein-molecule pair, which enables two-way screening.

\paratitle{Cross-modal Generation}.
This tasks aim to generate data in one modality based on another modality. MM-Sci-LLMs are designed to process data from multiple modalities including texts, molecules, proteins, and genomes.
For example, Text2Molecule refers to generating molecular information (e.g., SMILES), based on input text descriptions. Many models can accomplish this task including Text2Mol~\cite{edwards2021text2mol}, MolT5~\cite{edwards2022translation}, and MOL-Instruction~\cite{fang2023molinstructions}. 
The bidirectional conversion capability of cross-modal generation models allows them to not only generate data in one modality based on data from another modality but also to convert data from one modality to another. For example, Molecule2Text is to produce precise and pertinent textual descriptions derived from molecular data, namely molecule captioning. 
Similar to the mutual generation of molecule and text, Text2Protein aims to generate protein-related information including sequences and structures based on text descriptions. A representative model is ProteinDT~\cite{liu2023textguided} which introduced a text-guided protein design method. Conversely, Protein2Text converts protein sequences into detailed text descriptions to express the complex biological characteristics of proteins. Prot2Text~\cite{abdine2023prot2text}, BioMedGPT-10B~\cite{luo2023biomedgpt} and MOL-Instruction~\cite{fang2023molinstructions} can address this task effectively.  
Additionally, Protein2Molecule involves the process of generating relevant molecules from protein sequences, which can be achieved by ChemBERTaLM~\cite{uludoğan2022exploiting} and DrugGPT~\cite{li2023druggpt}. Overall, cross-modal generation in multi-modal models plays a crucial role in bridging the gap between different modalities and enabling the creation of new and cohesive multi-modal data. 

\paratitle{Evaluation Metric}. The evaluation criteria for MM-Sci-LLMs are similar to those used by uni-modal LLMs introduced in previous sections (Sec. \ref{sec:text-LLM-eval}, \ref{sec:mol-LLM-eval}, \ref{sec:prot-LLM-eval}, and \ref{sec:gene-LLM-eval}), such as the standard classification and retrieval metrics. Hence, we will not elaborate on them again.
In particular, in multimodal generation tasks, we emphasize the crucial alignment between the generated content and the input instruction, which necessitates evaluation using specific cross-modal alignment metrics. These metrics evaluate how well the model's output in one modality (e.g., molecule) aligns with the input or context provided in another modality (e.g., text). However, defining these metrics is challenging due to inherent difficulties in quantifying alignment across different modalities. Therefore, human evaluation often plays a crucial role as expert evaluators can assess biological relevance, novelty, and practical applicability of the generated content, providing insights that automated metrics might overlook.

\subsection{Summary}\label{sec:mm-sci-LLM-summary} 
In this section, we conduct a comprehensive examination of the multimodal scientific models MM-Sci-LLMs, which integrate diverse modalities including molecules, proteins, genomes, and textual data. Our investigation entails a detailed analysis of their distinctive architectural designs and capabilities, with particular emphasis on the interactions across multiple modalities. Additionally, we scrutinize the datasets pivotal for training and evaluating these models. 
MM-Sci-LLMs represent a rapidly growing research direction that integrates multiple modalities of scientific data. Currently, this field is still in its early developmental stages, offering significant opportunities for pioneering works and innovative applications in science discovery.

	\clearpage
	
	\section{Conclusion and Perspective} \label{sec-conclusion}

In this survey, we have explored the latest advancements in Sci-LLMs, particularly in the biological and chemical fields. The survey began by establishing the foundational concepts of scientific languages, encompassing textual, molecular, protein, and genomic categories, as detailed in Sec. \ref{sec-background}. We then delved into each scientific language type, examining their recent developments. This examination included an in-depth look at their model architectures, capabilities, datasets, and evaluation, covered in Sec.~\ref{sec-text}, Sec.~\ref{sec-molecule}, Sec.~\ref{sec-protein}, and Sec.~\ref{sec-genome}, respectively. A notable highlight of the survey is Sec.~\ref{sec-multimodal}, where we presented the emerging multimodal Sci-LLMs, focusing on the integration of multiple scientific language systems. 

In what follows, we discuss the research challenges of Sci-LLMs and potential avenues for future exploration. Given the considerable advantages and growing prominence of generative LLMs in scientific applications, our discussion will primarily focus on these generative Sci-LLMs. This targeted approach allows us to thoroughly examine the unique aspects and possibilities that generative models offer in advancing the field of Sci-LLMs.

\subsection{Critical Challenges}
Although previous research has made notable strides in the field of Sci-LLMs, it is important to acknowledge that this area of study is still in its nascent stages. In the process of preparing this survey, we conducted a thorough examination of existing studies and identified several critical challenges that are yet to be addressed. These challenges highlight the evolving nature of Sci-LLM research and underscore the need for continued exploration in this domain.

\subsubsection{Training Data} \ \\
Data stands as the foundation of AI model development. In exploring Sci-LLMs, we focus on key factors that influence their growth and effectiveness. 
\begin{itemize}
    \item \textbf{Scale of Pre-training Datasets}. The scale of pre-training datasets is a pivotal factor constraining the advancement of larger-scale Sci-LLMs. In the realm of language models, scaling laws delineate the relationship between an LLM's size (quantified by parameters or computational resources) and its performance, indicating that a model's capabilities and effectiveness tend to enhance, often in a nonlinear or exponential manner, as its size increases. Given that data acts as the fuel for AI models, amassing a substantial volume of pre-training data is critical for the development of larger and potentially more proficient LLMs. When contrasting general LLMs with Sci-LLMs, a stark difference emerges in their data scales. General LLMs often comprise hundreds of billions of parameters, while Sci-LLMs typically operate on the scale of billions. This disparity is largely due to the vast abundance of human language data on the web. For instance, ChatGPT, with its 175 billion parameters, is estimated to be trained {on 570 billion tokens. In contrast, a protein language model like ProGen, despite having 1.2 billion parameters, is limited to training with 280 million protein sequences, encompassing tens of billions of tokens}. However, there is a silver lining. With the continuous advancements in sequencing technologies, including DNA and protein sequencing, more gene and protein sequences are being identified. These advancements are expected to significantly contribute to the development of larger and more capable Sci-LLMs.
    \item \textbf{Quality of Finetuning Datasets}. Following pre-training, Sci-LLMs undergo a supervised fine-tuning process using labeled datasets. This stage is pivotal as it trains the models to follow specific instructions and generate the desired outputs. The effectiveness of this fine-tuning is heavily dependent on the quality of the labeled datasets, which can be assessed in terms of scale, diversity, and balance. 
    A larger scale of fine-tuning datasets generally leads to more proficient models, as they are exposed to a wider array of scenarios. However, compiling extensive annotated data can be resource-intensive. Besides, data diversity is also crucial; reports have shown that exposing LLMs to a broader spectrum of instructions enhances their ability to generalize across various downstream tasks. Equally important is the balance within the datasets. An imbalanced distribution of labels can lead Sci-LLMs to disproportionately favor predictions in the majority class, undermining their effectiveness. 
    \item \textbf{Lack of Cross-modal Datasets}. The quality of these cross-modal datasets, predominantly used in the fine-tuning stages, must be rigorously assessed in terms of scale, diversity, and balance. However, beyond these factors, the alignment between modalities emerges as a crucial and more complex consideration. Alignment entails ensuring that each data point contains samples from different language systems which express the same or relevant semantic content. This requirement demands consistent semantic representation across different modalities. 
    This deficiency could result in a narrowed scope of instructional diversity and a limited span of language systems included. To foster the effective development of robust multi-modal Sci-LLMs, it is essential to ensure that these datasets are not just extensive in scale but also rich in diversity and balance. They must also maintain strong semantic alignment across the various modalities, thereby enabling more accurate cross-modal interactions and interpretations.
\end{itemize}
The extensive pre-training corpus of Sci-LLMs provides vital exposure to a diverse array of semantic and linguistic elements. 
In fine-tuning these models, it is imperative to focus on aspects such as scale, diversity, balance and alignment across modalities. Such attention ensures the maximization of Sci-LLMs' performance in various applications.

\subsubsection{Architectures and Learning Objectives} \ \\
While most LLMs rely on Transformer-based architectures to learn semantic relevance in languages, this approach may not be ideally suited for Sci-LLMs. Several reasons underpin this mismatch:
\begin{itemize}
    \item \textbf{Handling Longer Sequences}. Scientific languages often comprise sequences that are significantly longer than typical natural language sentences. For instance, a protein sequence might contain hundreds or thousands of amino acids, while a natural language sentence usually has fewer than 30 words. This disparity in sequence length poses computational efficiency challenges for traditional Transformer-based architectures. To address this, recent models such as RWKV, Hyena, and S4 have been developed, showcasing innovations in handling longer sequences more effectively
    
    \item \textbf{Incorporating 3D Structural Information}. Current biological and chemical language models struggle to explicitly incorporate crucial 3D structural information, which is often more important than sequences alone for understanding functions and properties. 3D structural data, typically represented as atom coordinates, does not seamlessly integrate into traditional language systems. Recent studies have attempted to map 3D structures to learning tokens, aiming to create structural tokens in addition to sequence tokens. These efforts, though promising, are still in their infancy.

    \item \textbf{Autoregressive Learning Objective Limitations}. Natural language generation is generally processed in an autoregressive manner. However, this is not the case with biological and chemical languages, which are not inherently created or interpreted autoregressively. In these domains, sequences function as a whole. Therefore, an ideal learning objective for a generative Sci-LLM should allow the model to capture semantic information from the entire sequence, considering bidirectional context, rather than relying solely on one-way information flow.
\end{itemize}
These considerations highlight the need for specialized adaptations or entirely new model architectures and learning objectives for Sci-LLMs to effectively process and understand scientific languages, taking into account their unique characteristics and requirements.

\subsubsection{Model Evaluation} \ \\
The effective evaluation of LLMs remains a critical area of research focus. As detailed in Sec.~\ref{sec-text}, researchers often utilize Bloom's taxonomy, a framework defining six cognitive levels of educational objectives, to assess learning and understanding in LLMs. However, the unique nature of Sci-LLMs necessitates a tailored approach to evaluation.
Unlike general LLMs, Sci-LLMs are primarily geared towards mastering scientific knowledge and demonstrating utility in research-oriented scenarios. To address this, we propose a novel evaluation framework that categorizes Sci-LLM capabilities based on the complexity of scientific knowledge and research acumen. This framework classifies capabilities into pre-college, college, and post-college levels, aligning the model's understanding and output with the sophistication expected at these educational stages.

However, this methodology predominantly addresses textual outputs. The challenge intensifies when evaluating the generation of biological and chemical languages, such as those used in the de novo design of proteins or molecules. Currently, it is arduous for computational systems, and even more so for human experts like biologists and chemists, to accurately ascertain the characteristics and efficacy of these generated entities based solely on observation. Computational methods like molecular simulation offer some insights, but their reliability can vary, sometimes deviating significantly from actual outcomes, not to mention that they are usually too slow to closely collaborate with Sci-LLMs.
The computational metrics discussed in Sec.\ref{sec-molecule}, Sec.\ref{sec-protein}, and Sec.~\ref{sec-genome} offer indirect indications of the quality of newly generated proteins or molecules. However, these indicators are not always definitive. Wet-lab experiments remain the gold standard for validation, yet they are often beyond the scope and capabilities of many AI research teams. This gap underscores the urgent need for more robust and reliable computational evaluation and benchmarking systems. Such systems would not only augment the accuracy of assessments in the absence of wet-lab experimentation but also enhance the overall development and application of Sci-LLMs in scientific research.

\subsubsection{Ethics} \ \\
The ethical considerations in the development and application of Sci-LLMs are multifaceted and critical. Firstly, data privacy and consent are paramount, especially when dealing with sensitive biological data. Ensuring that genomic data, for instance, is anonymized and used with proper consent is vital to maintaining individual trust. Secondly, there is the risk of misuse of information, such as the potential for creating harmful biological substances or contributing to bioterrorism, necessitating stringent control measures. Bias in algorithmic decision-making also poses a significant ethical challenge. Sci-LLMs, like general LLM systems, can perpetuate and amplify biases present in their training data, leading to skewed outcomes in scientific research and applications. 
Lastly, there is an ethical imperative to ensure equitable access to the benefits of Sci-LLMs, preventing the exacerbation of existing inequalities in scientific research and healthcare.
Addressing these ethical issues requires a collaborative approach involving ethicists, scientists, policymakers, and other stakeholders to develop robust guidelines that ensure responsible and beneficial use of Sci-LLMs.

\subsection{Future Directions}
To propel forward the field of AI-driven scientific discovery, we have pinpointed five key research avenues that hold promise for the advancement of more capable Sci-LLMs in the future.
\begin{itemize}
    \item \textbf{Construction of Larger-scale, High-quality, and Cross-modal Training Datasets}. As we analyzed before, a critical challenge lies in the realm of training data. Expanding the pre-training dataset is contingent upon advancements in biological sequencing technologies and the open-sourcing of chemical molecular data. Concurrently, the responsibility of maintaining high-quality fine-tuning datasets, particularly in the cross-modal domain, falls to AI research teams. These enhanced datasets are instrumental in not only improving the understanding of single-modal language systems by Sci-LLMs but also in enabling them to tackle biological and chemical tasks from a more integrated, cross-modal perspective. This dual focus on expanding and refining training datasets is crucial for realizing the potential of larger and more effective Sci-LLMs in practical scientific research scenarios.

    \item \textbf{Incorporating 3D Stereo-Temporal Information into Scientific Language Systems}. The inherent nature of biological and chemical data is rooted in the three-dimensional physical world, where the functionalities are primarily determined by their constantly changing 3D structures. This highlights the urgent need to integrate 3D structural and temporal information into language-based modeling approaches. In addition to the recent structural tokens, it may be equally crucial to explore structural motifs. These motifs, akin to Lego building blocks in their ability to interlock and compose protein conformations, represent specific 3D structures that hold significant potential for modeling. 
    By incorporating these structural motifs into the scientific lexicon, they can be learned and processed by Sci-LLMs in a manner similar to other tokens. 
    Besides, the learning process is supposed to capture the dynamic alternation of 3D structures. This would enrich the models' understanding and representation of the intricate three-dimensional aspects of biological and chemical data.

    \item \textbf{Sci-LLM's Marriage with External Knowledge Sources}. After centuries of development, scientists have accumulated a wealth of domain knowledge. For instance, biologists have developed the Gene Ontology, a comprehensive knowledge graph that systematically delineates the relationships between protein sequences and their functions. The potential synergy between Sci-LLMs and these external knowledge sources represents a substantial opportunity for enhancement. By integrating Sci-LLMs with such rich, structured external resources, their capabilities can be significantly expanded, overcoming the potential issues such as hallucination. This approach underscores the importance of looking beyond sequence data and embracing a more holistic view in the development of Sci-LLMs.

    \item \textbf{Sci-LLM Interacting with Physical Simulation}. Current Sci-LLMs are primarily adept at capturing the semantics inherent in sequence data. However, the realm of scientific understanding extends far beyond sequence data alone, encompassing a wealth of additional resources that can significantly contribute to a deeper comprehension. For instance, the field of Molecular Dynamics (MD) offers simulations that mirror the physical world, providing insights into molecular behavior and interactions. The interplay between Sci-LLMs and MD would not only complement the existing data-processing strengths of Sci-LLMs but also enrich their analytical power, leading to more nuanced and accurate interpretations of complex scientific phenomena.
    
    \item \textbf{Augmenting Sci-LLMs with Specialized Tools and Agents}. General LLMs have demonstrated remarkable proficiency in leveraging software and tools, such as calculators, to accomplish complex tasks. This capability is equally pertinent to Sci-LLMs, which can be significantly enhanced through integration with science-specific computational tools \cite{ramos2024areview}. A prime example of this is the recent success of GPT-4, which has shown proficiency in web searching, academic paper analysis, and designing experiments, including replicating intricate cross-coupling reactions like the Suzuki and Sonogashira experiments. The realm of scientific tools is vast and intricate, encompassing a wide array of computational resources. The challenge of accurately selecting the appropriate tool or software from tens of thousands of possibilities, precisely inputting data, and effectively interpreting the output, presents a rich area for research exploration. 
    The advancement in this area holds significant promise for augmenting the capabilities of Sci-LLMs, enabling them to perform more sophisticated and specialized scientific tasks with higher precision and efficiency.
    
    \item \textbf{Development of Computational Evaluation Metrics and Benchmarks}. The proper evaluation of model performance is crucial, especially for generative tasks. This extends beyond the capabilities of AI teams alone, necessitating greater collaboration with experts in computational biology. Unlike the time-intensive process of molecular dynamics, the ideal evaluation metrics for these models should be both rapid to compute and accurate in reflecting outcomes. Establishing such metrics would facilitate the quick construction of evaluation benchmarks, significantly speeding up research iterations. This collaborative approach, merging AI with computational biology, is essential for developing effective, efficient evaluation strategies that can keep pace with the rapid advancements in model development.

    \item \textbf{Super-alignment with Human Ethics.} Alignment with human ethics is pivotal in the development of Sci-LLMs. Super-alignment, in this context, extends beyond basic adherence to ethical norms and encompasses a deep integration of ethical reasoning capabilities within Sci-LLMs. This involves equipping these models with a nuanced understanding of ethical principles and the ability to apply them contextually in diverse scientific domains. For instance, in genomic research, Gene-LLMs should be able to weigh the benefits of genetic discoveries against potential risks like genetic discrimination or privacy breaches.
   Future research in this area could focus on developing frameworks and algorithms that enable Sci-LLMs to recognize and respect diverse ethical viewpoints, allowing them to suggest or evaluate actions based on ethical principles. Ultimately, the goal is to ensure that the advancements driven by Sci-LLMs are both scientifically robust and ethically sound, thereby fostering trust and acceptance in their applications across various scientific fields.
    
\end{itemize}
 
In this survey, we have conducted a systematic review of scientific large language models, particularly emphasizing the biological and chemical domains. We established the foundational concepts of scientific languages including scientific texts, molecules, proteins, and genomes. Then we delved into each scientific language type, examining the recent developments in Sci-LLMs. This examination included an in-depth look at their model architectures, capabilities, datasets, and evaluation methods. Moving forward, we pinpointed four key challenges and proposed seven promising research directions for future exploration. Our goal is to serve as a comprehensive and insightful resource for both the AI and fundamental science communities, fostering collaborative efforts and driving forward the `AI for Science' research agenda. By effectively modeling scientific languages, LLMs pave a more stable path towards the realization of artificial general intelligence.

\clearpage
\subsection*{Contributions}
This survey is led and organized by Qiang Zhang, Keyan Ding and Huajun Chen, and the contributions of all authors for each section are listed as follows:
\begin{itemize}
    \item Sec.\ref{sec:introduction}: Qiang Zhang, Keyan Ding, Huajun Chen 
    \item Sec.\ref{sec-background}: Qiang Zhang, Keyan Ding, Renjun Xu, Huajun Chen 
    \item Sec.\ref{sec-text}: Jing Yu, Keyan Ding, Qiang Zhang, Kehua Feng, Tao Huang, Pengju Yan, Xiaolin Li, Huajun Chen 
    \item Sec.\ref{sec-molecule}: Tianwen Lyu, Keyan Ding, Qiang Zhang, Xiang Zhuang, Jiyu Cui, Huabin Xing, Huajun Chen 
    \item Sec.\ref{sec-protein}: Yiwen Zhang, Keyan Ding, Qiang Zhang, Zeyuan Wang, Zhuoyi Xiang, Hongyang Chen, Huajun Chen 
    \item Sec.\ref{sec-genome}: Qingyu Yin, Keyan Ding, Qiang Zhang, Mengyao Zhang, Jinlu Zhang, Xiaohui Fan, Huajun Chen 
    \item Sec.\ref{sec-multimodal}: Xinda Wang, Keyan Ding, Qiang Zhang, Xiaotong Li, Yuhao Wang, Ming Qin, Huajun Chen  
    \item Sec.\ref{sec-conclusion}: Qiang Zhang, Keyan Ding, Huajun Chen 
\end{itemize}
	
	\clearpage
	\bibliographystyle{ACM-Reference-Format}
	\bibliography{acm-reference}
	

\end{document}